\documentclass[10pt,twocolumn,letterpaper]{article}

\usepackage{iccv}
\usepackage{times}
\usepackage{epsfig}
\usepackage[utf8]{inputenc}
\usepackage[T1]{fontenc}
\usepackage{url}
\newcommand{\mycomment}[1]{}
\usepackage{graphicx}
\usepackage{amsmath}
\usepackage{amssymb}
\usepackage[version=4]{mhchem}
\usepackage{algorithm}
\usepackage{algpseudocode}
\usepackage{subcaption}
\usepackage{mhchem}
\usepackage{stmaryrd}
\usepackage{bbold}
\usepackage{amsfonts}
\usepackage{array}
\usepackage{booktabs} 
\usepackage{enumitem}
\usepackage{textcomp}
\usepackage{tabu}

\usepackage{float}
\usepackage{xspace}
\usepackage{verbatim}
\usepackage{mathdots}
\usepackage{bm}
\usepackage{balance}

\newcolumntype{C}[1]{>{\centering\arraybackslash}p{#1}}
\usepackage{rotating} %table in landscape
\usepackage{color}
\usepackage{longtable}
\usepackage{calc}
\usepackage{hhline}
\usepackage{ifthen}
\usepackage{lscape}
\newcolumntype{L}{>{\centering\arraybackslash}m{2cm}}
\definecolor{cosmiclatte}{rgb}{1.0, 0.97, 0.91}
\newcolumntype{a}{>{\columncolor{cosmiclatte}}l}
\definecolor{kellygreen}{rgb}{0.3, 0.73, 0.09}
\usepackage{pifont}% http://ctan.org/pkg/pifont
\newcommand{\xmark}{\ding{55}}%
\newcommand{\greencheck}{{\color{kellygreen}\checkmark}}
\newcommand{\redcheck}{{\color{red}\xmark}}
\usepackage{caption} \usepackage[table,xcdraw]{xcolor}
\usepackage{multirow, booktabs, makecell}
\usepackage[table]{xcolor}
\usepackage[accsupp]{axessibility} 
\usepackage{colortbl}

\definecolor{saffron}{rgb}{0.96, 0.77, 0.19}
\definecolor{schoolbusyellow}{rgb}{1.0, 0.85, 0.0}
\definecolor{spirodiscoball}{rgb}{0.06, 0.75, 0.99}
\definecolor{paleaqua}{rgb}{0.74, 0.83, 0.9}
\definecolor{lightblue}{RGB}{221,235,247}
\definecolor{darkblue}{RGB}{31,78,120}
\definecolor{puce}{HTML}{513b41}
\definecolor{tifblue}{HTML}{c8f4f9}
\definecolor{lightbluegrey}{RGB}{230,240,255}
\definecolor{darkgrey}{RGB}{50,50,50}
\definecolor{darkgreen}{RGB}{27,64,74}
\definecolor{lightsalmon}{rgb}{1.0, 0.63, 0.48}
\definecolor{taupegray}{rgb}{0.55, 0.52, 0.54}
\definecolor{yaleblue}{rgb}{0.06, 0.3, 0.57}
\definecolor{darksalmon}{rgb}{0.91, 0.59, 0.48}
\definecolor{lightpastelpurple}{rgb}{0.69, 0.61, 0.85}
\usepackage[pagebackref=true,breaklinks=true,colorlinks,bookmarks=false]{hyperref}
\usepackage[capitalize]{cleveref}
\crefname{section}{Sec.}{Secs.}
\Crefname{section}{Section}{Sections}
\Crefname{table}{Table}{Tables}
\crefname{table}{Tab.}{Tabs.}
 \iccvfinalcopy % *** Uncomment this line for the final submission

\def\iccvPaperID{10950} % *** Enter the ICCV Paper ID here

\ificcvfinal\fi

\begin{document}

%%%%%%%%% TITLE
\title{Vision Transformer Adapters 
    for Generalizable Multitask Learning}

\author{Deblina Bhattacharjee, Sabine Süsstrunk and Mathieu Salzmann\\
School of Computer and Communication Sciences, EPFL, Switzerland\\
{\tt\small \{deblina.bhattacharjee, sabine.susstrunk, mathieu.salzmann\}@epfl.ch}
}

\maketitle
\ificcvfinal\thispagestyle{empty}\fi

%%%%%%%%% ABSTRACT
\begin{abstract}
  
  %We introduce the first multitasking vision transformer adapters that learn generalizable task affinities which can be applied to novel tasks and domains. Integrated into an off-the-shelf vision transformer backbone, our adapters can simultaneously solve multiple dense vision tasks in a parameter-efficient manner, unlike existing multitasking transformers that are parametrically expensive. Typically, multitask models require retraining or fine-tuning their shared parameters whenever a new task or domain is added. This means that the task affinities neither transfer to novel tasks nor generalize across different domains. Thanks to our multitask vision transformer adapters, we do not need to fully retrain a vision transformer model for a novel task or fine-tune it to a novel domain. To learn the generalizable task affinities, we introduce a task-adapted attention mechanism within our adapter framework that combines gradient-based task similarities with attention-based ones. The learned task affinities generalize to the following settings: zero-shot task transfer, unsupervised domain adaptation, and generalization without fine-tuning to novel domains. We demonstrate that our approach outperforms not only the existing convolutional neural network-based multitasking methods but also the vision transformer-based ones. We will make our code and models publicly available upon publication.

  We introduce the first multitasking vision transformer adapters that learn generalizable task affinities which can be applied to novel tasks and domains. Integrated into an off-the-shelf vision transformer backbone, our adapters can simultaneously solve multiple dense vision tasks in a parameter-efficient manner, unlike existing multitasking transformers that are parametrically expensive. In contrast to concurrent methods, we do not require retraining or fine-tuning whenever a new task or domain is added. We introduce a task-adapted attention mechanism within our adapter framework that combines gradient-based task similarities with attention-based ones. The learned task affinities generalize to the following settings: zero-shot task transfer, unsupervised domain adaptation, and generalization without fine-tuning to novel domains. We demonstrate that our approach outperforms not only the existing convolutional neural network-based multitasking methods but also the vision transformer-based ones. Our project page is at \url{https://ivrl.github.io/VTAGML}.
\end{abstract}

%%%%%%%%% BODY TEXT
\section{Introduction}
\label{sec:intro-AVTaR}

In the past few years, vision transformers~\cite{crossvit, cswin, dosovitskiy2021an, mvitv2-classification, swin, wang2021crossformer} have  grown in popularity at an incredible pace. They have now achieved state-of-the-art results, outperforming Convolutional Neural Network (CNN) based methods not only in image classification~\cite{foveaTer-classification, multi-label-image-classifcation-transformer, mvitv2-classification} but also in many dense prediction tasks such as semantic segmentation~\cite{maskformer,  segmenter2021, segmentation-txn-2022, topformer}, monocular depth estimation~\cite{DPT, yang2021transformers-depth}, and surface normal prediction~\cite{surfacenormal-txn-cvpr-20, depth-surfacenormal-txn}. 
Therefore, utilizing the power of vision transformers in a unified framework to simultaneously solve multiple tasks seems a natural way forward. %Therefore, a natural consequence lies in utilizing the power of vision transformers in a unified framework to simultaneously tackle multiple tasks. 
%With the exception of MulT~\cite{IPT, MulT, hu2021unit}, there are no existing works that leverage a transformer framework for multitask learning (MTL). 
Nevertheless, only a few works~\cite{ MulT, IPT, hu2021unit, spatiotemporalMTL, video-multitask-transformer} have attempted this so far, and all of them rely on handcrafted transformer architecture designs.
%such multitask vision transformers have handcrafted architectures. 
Specifically, IPT and ST-MTL~\cite{IPT, spatiotemporalMTL} exploit a multi-head multi-tail architecture 
%specifically 
tailored to solve specific tasks; MulT~\cite{MulT}
%adopts a transformer encoder-decoder structure for dense predictions, where the 
leverages a pairwise task attention strategy handcrafted to utilize surface normal prediction as reference task for dense predictions; and UniT~\cite{hu2021unit} as well as Vid-MTL~\cite{video-multitask-transformer} use a multimodal transformer architecture to achieve multiple pairwise task predictions across different modalities.
%Although vision transformers for individual tasks have achieved great success, real-world problems are not inherently isolated, which calls for vision transformers that can perform multiple tasks concurrently. For example, autonomous driving requires lane detection, semantic understanding, obstacle detection, and depth estimation simultaneously. %For example, reconfiguring and understanding comics scenes requires semantic understanding, depth estimation, and visual saliency prediction simultaneously. 
%However, such multitask vision transformers have handcrafted architectures. In particular, IPT~\cite{IPT} proposed a multi-head multi-tail method strictly catering to four low-level tasks. MulT~\cite{MulT}, on the other hand, adopts a transformer encoder-decoder structure for dense predictions, where the pairwise task attention is handcrafted to utilize the surface normal task as a reference task. Similarly, UniT~\cite{hu2021unit}, uses a multimodal transformer architecture to learn the most prominent tasks across different modalities. 
While these multitasking vision transformer-based methods~\cite{IPT, MulT, hu2021unit,  spatiotemporalMTL, video-multitask-transformer} outperform their multitasking CNN-based counterparts~\cite{endtoendMTL, xtam, cross-stitch, standley2019, taskswitchnet, Vandenhende_2021, mti-net, padnet, zamir2020consistency, zamir2018taskonomy}, none of the existing vision transformer-based or CNN-based MTL methods can adapt to new tasks as well as to novel domains. %(shown in Table~\ref{tb:Taxonomy-MTL}).
\begin{figure}[t]
\centering
{\includegraphics[width=1.0\linewidth ]{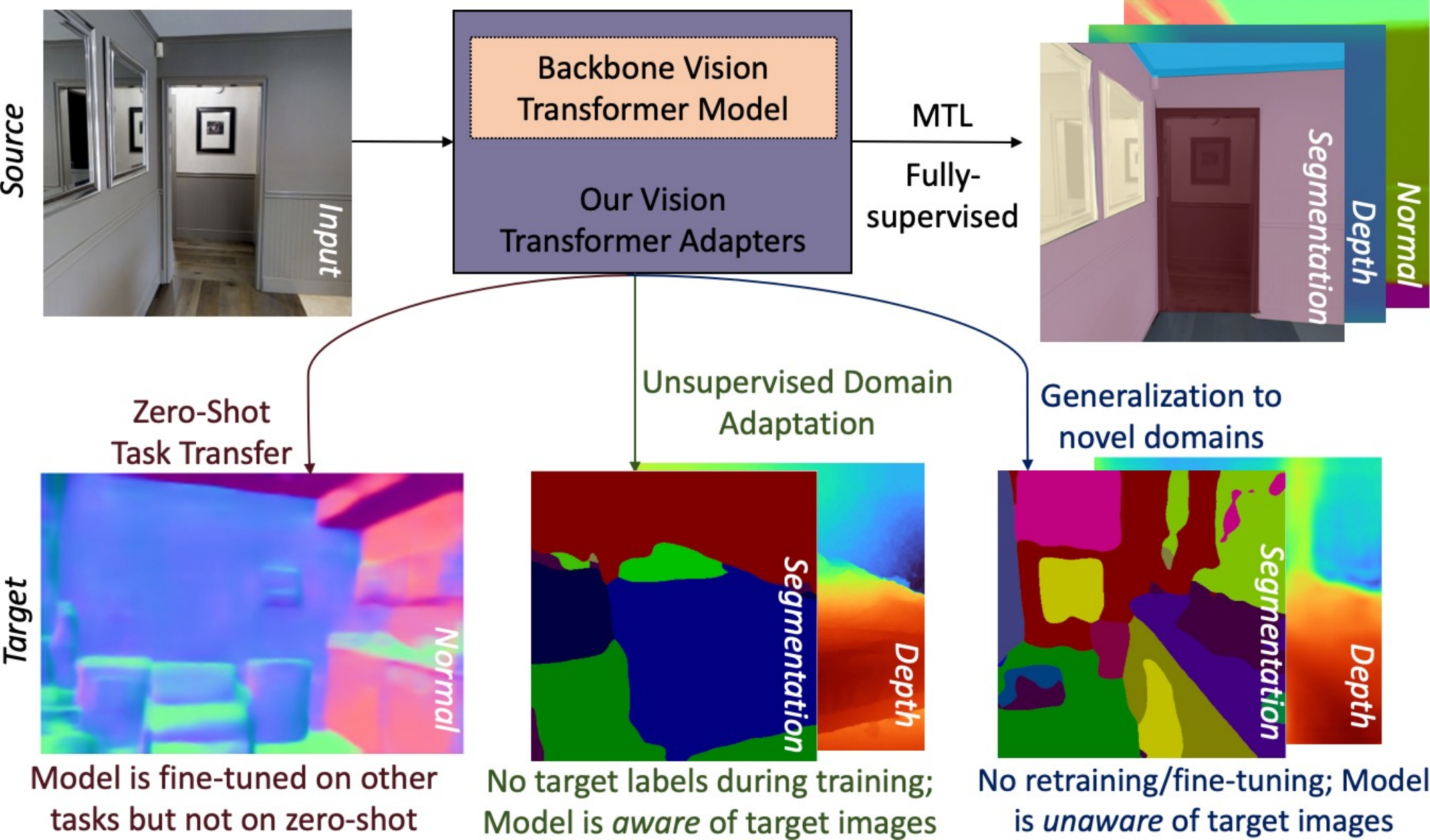}}
\setlength{\abovecaptionskip}{2mm}\caption{\textbf{Motivation of our work.} Unlike existing MTL methods, our vision transformer adapters generalize to novel tasks and domains.}
\vspace{-19pt}\label{fig:teaser}
\end{figure}
%Besides the three transformer-based MultiTask Learning (MTL) methods~\cite{ MulT, IPT, UNIT}, 
%
%By leveraging task relationship that, in turn, allows sharing task parameters, multitask vision transformers can reduce computational training costs and yield better performance over the single-task counterpart. However, training a multitasking framework is computationally costly, if not 
%Typically, multitasking frameworks learn task relations in a pairwise manner~\cite{xtam, mti-net, Vandenhende_2021, MulT}, with the exception of ~\cite{taskswitchnet} where a
%Typically, existing multitask learning (MTL) frameworks fail to generalize or transfer their learned task affinities across new tasks or domains
In fact, it was observed in the seminal work of~\cite{zamir2018taskonomy} and confirmed in subsequent MTL studies~\cite{MulT, standley2019,zamir2020consistency} that the multitask affinities learned by existing MTL frameworks are \emph{not} transferable or generalizable.  

This raises the following question: Is there a way we can learn transferable and generalizable task affinities such that multitask affinities transfer to novel tasks and generalize to novel domains, thereby allowing us to reuse an existing network? 
To answer this, we introduce vision transformer adapters for generalizable multitask learning and propose an \emph{automated} framework that can learn \emph{transferable and generalizable} task affinities which can adapt to new tasks or domain representations in a \emph{parameter-efficient} manner. Additionally, unlike existing transformer-based handcrafted MTL methods~\cite{IPT, MulT, hu2021unit} that learn task affinities in a pair-wise manner, our vision transformer adapters learn task affinities in an automated way and across \emph{all} the tasks.
To achieve this, we equip our vision adapters with three mechanisms: (1) An improved gradient-based task similarity approach (TROA) first introduced in~\cite{tawt}; (2) a novel task-adapted attention mechanism (TAA) that combines the gradient-based task similarities with attention-based ones, thereby learning transferable and generalizable task affinities; and (3) a task-scaled normalization to account for the different task scales. The resulting module can then be seamlessly integrated with a pre-trained, frozen encoder backbone architecture such as ViT~\cite{dosovitskiy2021an}, Swin~\cite{swin}, Pyramid Transformer~\cite{wang2021pvtv2}, or Focal Transformer~\cite{yang2021focal}. Our approach is independent of the choice of the vision transformer backbone, unlike existing transformer-based MTL methods.
Our contributions are summarized as follows:
\vspace{-3pt}
\begin{itemize}
    \item We introduce vision adapters for generalizable multitask learning that leverages a pre-trained vision transformer backbone to learn transferable and generalizable features at a low computational cost.
    \vspace{-2pt}
    \item At the heart of our vision adapter, we introduce a novel task-adapted attention mechanism (TAA) that automatically learns task dependencies from the shared representation, by combining gradient-based task similarities (TROA) 
    with attention-based ones. 
    \vspace{-2pt}
    \item Our task affinities transfer to different settings including multitask learning, zero-shot task transfer learning, and unsupervised domain adaptation. Moreover, our task affinities generalize to novel domains \emph{without} requiring any fine-tuning. 
    \vspace{-2pt}
    \item Our multitasking vision transformer adapters can be integrated with different transformer backbones such as ViT~\cite{dosovitskiy2021an}, Swin~\cite{swin}, Pyramid Transformer~\cite{wang2021pvtv2}, and Focal Transformer~\cite{yang2021focal}, achieving a significant increase in performance in a parameter-efficient way.
\end{itemize}
Our experiments evidence that our method outperforms both state-of-the-art CNN-based multitasking methods~\cite{tawt, endtoendMTL, xtam, cross-stitch, ttnet, taskswitchnet, Vandenhende_2021, zamir2020consistency, zamir2018taskonomy} as well as transformer-based ones~\cite{MulT, hu2021unit}. %as shown in Figure~\ref{fig:teaser-avtar-ch5}. 
%In short, AVTaR achieves higher prediction accuracy at a lower computational cost across novel tasks and different domains.
\begin{table*}[h!]
\setlength\tabcolsep{1.7pt}
\centering
\scalebox{0.58}{
\begin{tabular}{llLLLLLLLl}
%\toprule 
&& \multicolumn{4}{l}{~~~~~~~~~~~~~~~~~~~~~~~~~~~~~~~~~~~~~\textbf{Architecture}} & \multicolumn{4}{l}{~~~~~~~~~~~~~~~~~~~~~\textbf{Task-affinity generalization}} \\ \cmidrule(lr){3-6}\cmidrule(lr){7-10}
&\multicolumn{1}{l}{~~\textbf{Methods}~~~~~} & \multicolumn{1}{l}{\textbf{Encoder-focused}} & \multicolumn{1}{l}{~~~~\textbf{Decoder-focused}} & \multicolumn{1}{l}{~~~\textbf{Attention}} & \multicolumn{1}{l}{~~~\textbf{Task-loss}} &\multicolumn{1}{l}{~~~~\textbf{MTL}~~~~}&\multicolumn{1}{l}{\textbf{Task-transfer}}& \multicolumn{1}{l}{~~~~\textbf{UDA}}& \multicolumn{1}{l}{\textbf{Novel domain}}\\ 
\cmidrule(lr){1-2}\cmidrule(lr){3-6}\cmidrule(lr){7-10}
\multirow{9}{*}{CNN-based} & MTL-baseline~\cite{Vandenhende_2021}                              & \multicolumn{1}{L}{\greencheck}                          &\multicolumn{1}{L}{\redcheck}                                &\multicolumn{1}{L}{\redcheck}                                         &\multicolumn{1}{L}{\redcheck}     &\multicolumn{1}{L}{\greencheck}&\multicolumn{1}{L}{\redcheck} & \multicolumn{1}{L}{\redcheck}&   \multicolumn{1}{L}{\redcheck}              \\
&Consistency~\cite{zamir2020consistency}                       &\multicolumn{1}{L}{\redcheck}                         & \multicolumn{1}{L}{\greencheck}                         &   \multicolumn{1}{L}{\redcheck}                                    & \multicolumn{1}{L}{ \greencheck }    &\multicolumn{1}{L}{\greencheck}&\multicolumn{1}{L}{\greencheck}&\multicolumn{1}{L}{\redcheck} &\multicolumn{1}{L}{\redcheck}                   \\
&XTAM~\cite{xtam}                               &\multicolumn{1}{L}{\redcheck}                         &\multicolumn{1}{L}{\redcheck}                         &  \multicolumn{1}{L}{\greencheck}                                      &\multicolumn{1}{L}{\redcheck}            &\multicolumn{1}{L}{\greencheck}&\multicolumn{1}{L}{\redcheck} & \multicolumn{1}{L}{\greencheck}    &\multicolumn{1}{L}{\redcheck}        \\
&TAWT~\cite{tawt}                               & \multicolumn{1}{L}{\greencheck}                         &\multicolumn{1}{L}{\redcheck}                         &\multicolumn{1}{L}{\redcheck}                                          &\multicolumn{1}{L}{\redcheck}            &\multicolumn{1}{L}{\greencheck}&\multicolumn{1}{L}{\greencheck}&\multicolumn{1}{L}{\redcheck}      &\multicolumn{1}{L}{\redcheck}       \\
%&MTL-Uncertainty~\cite{MTL-uncertainty}                           &\multicolumn{1}{L}{\redcheck}                         &\multicolumn{1}{L}{\redcheck}                         &    \multicolumn{1}{L}{\redcheck}                                    & \multicolumn{1}{L}{\greencheck}     &\multicolumn{1}{L}{\greencheck}&\multicolumn{1}{L}{\redcheck} & \multicolumn{1}{L}{\redcheck}  &\multicolumn{1}{L}{\redcheck}                 \\
&Cross-stitch~\cite{cross-stitch}                           & \multicolumn{1}{L}{\greencheck}                         &\multicolumn{1}{L}{\redcheck}                         &  \multicolumn{1}{L}{\redcheck}                                      &\multicolumn{1}{L}{\redcheck}        &\multicolumn{1}{L}{\greencheck}&\multicolumn{1}{L}{\redcheck} &\multicolumn{1}{L}{\redcheck}      &\multicolumn{1}{L}{\redcheck}           \\ 
&MTAN~\cite{endtoendMTL}                           & \multicolumn{1}{L}{\greencheck}                         &\multicolumn{1}{L}{\redcheck}                         &\multicolumn{1}{L}{\redcheck}                                          &\multicolumn{1}{L}{\redcheck}             &\multicolumn{1}{L}{\greencheck}&\multicolumn{1}{L}{\redcheck} & \multicolumn{1}{L}{\redcheck}     &\multicolumn{1}{L}{\redcheck}      \\
%&MTI-Net~\cite{mti-net}                           &\multicolumn{1}{L}{\redcheck}                         & \multicolumn{1}{L}{\greencheck}                         &  \multicolumn{1}{L}{\greencheck}                                     &\multicolumn{1}{L}{\redcheck}         &\multicolumn{1}{L}{\greencheck}&\multicolumn{1}{L}{\redcheck} & \multicolumn{1}{L}{\redcheck}     &\multicolumn{1}{L}{\redcheck}          \\
&TSwitch~\cite{taskswitchnet}&\multicolumn{1}{L}{\redcheck}                         & \multicolumn{1}{L}{\greencheck}                         &\multicolumn{1}{L}{\redcheck}                                          & \multicolumn{1}{L}{\greencheck}      &\multicolumn{1}{L}{\greencheck}&\multicolumn{1}{L}{\redcheck} & \multicolumn{1}{L}{\redcheck} &\multicolumn{1}{L}{\redcheck}                 \\
%&Grad-norm~\cite{gradnorm}                           &\multicolumn{1}{L}{\redcheck}                         &\multicolumn{1}{L}{\redcheck}                         &\multicolumn{1}{L}{\redcheck}                                         & \multicolumn{1}{L}{\greencheck}      &\multicolumn{1}{L}{\greencheck}&\multicolumn{1}{L}{\redcheck} & \multicolumn{1}{L}{\redcheck}     &\multicolumn{1}{L}{\redcheck}             \\
%&PCGrad~\cite{pcgrad2020}                           &\multicolumn{1}{L}{\redcheck}                         &\multicolumn{1}{L}{\redcheck}                         & \multicolumn{1}{L}{\redcheck}                                        & \multicolumn{1}{L}{\greencheck}        &\multicolumn{1}{L}{\greencheck}&\multicolumn{1}{L}{\redcheck} &    \multicolumn{1}{L}{\redcheck}  &\multicolumn{1}{L}{\redcheck}           \\
&TTNet~\cite{ttnet}                           &\multicolumn{1}{L}{\redcheck}                         &\multicolumn{1}{L}{\redcheck}                         & \multicolumn{1}{L}{\redcheck}                                        & \multicolumn{1}{L}{\greencheck}        &\multicolumn{1}{L}{\greencheck}&\multicolumn{1}{L}{\greencheck}&    \multicolumn{1}{L}{\redcheck} &\multicolumn{1}{L}{\redcheck}            \\
%&PAD-Net~\cite{padnet}                           & \multicolumn{1}{L}{\redcheck}                         & \multicolumn{1}{L}{\greencheck}                         &  \multicolumn{1}{L}{\greencheck}                                     & \multicolumn{1}{L}{\redcheck}          &\multicolumn{1}{L}{\greencheck}&\multicolumn{1}{L}{\redcheck}& \multicolumn{1}{L}{\redcheck}      &\multicolumn{1}{L}{\redcheck}        \\
&Taskonomy~\cite{taskonomy2018}                           & \multicolumn{1}{L}{\redcheck}                         & \multicolumn{1}{L}{\greencheck}                         &    \multicolumn{1}{L}{\redcheck}                                     & \multicolumn{1}{L}{\greencheck}      &\multicolumn{1}{L}{\greencheck}&\multicolumn{1}{L}{\greencheck}& \multicolumn{1}{L}{\redcheck}     &\multicolumn{1}{L}{\redcheck}              \\
%&Taskgrouping~\cite{standley2019}                           & \multicolumn{1}{L}{\redcheck}                         & \multicolumn{1}{L}{\greencheck}                         &   \multicolumn{1}{L}{\redcheck}                                       & \multicolumn{1}{L}{\greencheck}      &\multicolumn{1}{L}{\greencheck}&\multicolumn{1}{L}{\redcheck}& \multicolumn{1}{L}{\redcheck}      &\multicolumn{1}{L}{\redcheck}             \\
\cmidrule(lr){1-2}\cmidrule(lr){3-6}\cmidrule(lr){7-10}
%&IPT~\cite{IPT}                               &\multicolumn{1}{L}{\redcheck}                         & \multicolumn{1}{L}{\greencheck}                         &  \multicolumn{1}{L}{\greencheck}                                       & \multicolumn{1}{L}{\greencheck}  &\multicolumn{1}{L}{\greencheck}&\multicolumn{1}{L}{\redcheck} &   \multicolumn{1}{L}{\redcheck} &\multicolumn{1}{L}{\redcheck}                    \\
\multirow{3}{*}{Vision Transformer-based}&ST-MTL~\cite{spatiotemporalMTL}                               &\multicolumn{1}{L}{\redcheck}                         & \multicolumn{1}{L}{\greencheck}                         &  \multicolumn{1}{L}{\greencheck}                                       & \multicolumn{1}{L}{\greencheck}  &\multicolumn{1}{L}{\greencheck}&\multicolumn{1}{L}{\redcheck} &   \multicolumn{1}{L}{\redcheck}   &\multicolumn{1}{L}{\redcheck}                  \\
%&Vid-MTL~\cite{video-multitask-transformer}                               &\multicolumn{1}{L}{\redcheck}                         & \multicolumn{1}{L}{\greencheck}                         &  \multicolumn{1}{L}{\greencheck}                                       & \multicolumn{1}{L}{\greencheck}  &\multicolumn{1}{L}{\greencheck}&\multicolumn{1}{L}{\redcheck} &   \multicolumn{1}{L}{\redcheck} &\multicolumn{1}{L}{\redcheck}                    \\
%&UniT~\cite{hu2021unit}                               &\multicolumn{1}{L}{\redcheck}                         & \multicolumn{1}{L}{\greencheck}                         &  \multicolumn{1}{L}{\greencheck}                                       & \multicolumn{1}{L}{\greencheck}  &\multicolumn{1}{L}{\greencheck}&\multicolumn{1}{L}{\redcheck} &   \multicolumn{1}{L}{\redcheck}   &\multicolumn{1}{L}{\redcheck}                  \\
&MulT~\cite{MulT}                               &\multicolumn{1}{L}{\redcheck}                         & \multicolumn{1}{L}{\greencheck}                         &  \multicolumn{1}{L}{\greencheck}                                       & \multicolumn{1}{L}{\greencheck}  &\multicolumn{1}{L}{\greencheck}&\multicolumn{1}{L}{\redcheck} &   \multicolumn{1}{L}{\greencheck}  &\multicolumn{1}{L}{\redcheck}                   \\
%\cmidrule(lr){2-2}\cmidrule(lr){3-6}\cmidrule(lr){7-10}
&\textbf{~~Our}      & \multicolumn{1}{L}{\greencheck}  & \multicolumn{1}{L}{\redcheck} &  \multicolumn{1}{L}{\greencheck}                                       & \multicolumn{1}{L}{\greencheck}  &\multicolumn{1}{L}{\greencheck}&\multicolumn{1}{L}{\greencheck}&\multicolumn{1}{L}{\greencheck}&\multicolumn{1}{L}{\greencheck} \\
\cmidrule(lr){1-2}\cmidrule(lr){3-6}\cmidrule(lr){7-10}
\end{tabular}}
\setlength{\abovecaptionskip}{0mm}
\caption[Taxonomy of MTL approaches]{\textbf{Taxonomy of MTL approaches.} Our vision transformer adapter method is an encoder-focused, task-balanced approach that uses task-adapted attention (TAA) to learn generalizable task affinities, unlike existing CNN-based and vision transformer-based MTL methods. Here, we list the methods that we evaluate in this work. A detailed taxonomy of other MTL methods is provided in supplementary.%This corroborates the findings of ~\cite{Vandenhende_2021}. The task relations learned by AVTaR generalize across MTL, task transfer, and the UDA settings, unlike the existing approaches.
}
   \label{tb:Taxonomy-MTL}%
\vspace{-10pt}
\end{table*}

\section{Related Work}
\label{sec:related work}
\paragraph{Multitask Learning.} Multitask learning has
been a fundamental problem for years; see Vandenhende et. al.~\cite{Vandenhende_2021} for a great survey.  %Conventionally, an MTL framework simultaneously predicts multiple outputs from a shared representation~\cite{zhang2021survey}. 
As noted by multiple works~\cite{fifty2021efficiently, standley2019, Vandenhende_2021}, MTL networks are unstable and require a strong \emph{balance} between tasks to perform well. Prior works~\cite{endtoendMTL, xtam, cross-stitch, standley2019, taskswitchnet, Vandenhende_2021, mti-net, padnet, zamir2020consistency, zamir2018taskonomy} aim to strike this balance 
%when jointly learning multiple vision tasks 
either using a gradient-based learning of task affinities in the encoded representations~\cite{tawt, endtoendMTL, cross-stitch, Vandenhende_2021, pcgrad2020}, or applying task conditioned gates to the decoder~\cite{taskswitchnet}, attention-based task similarities~\cite{xtam, mti-net, padnet} or weighted task losses~\cite{gradnorm, MTL-uncertainty, ttnet}. %(see Table~\ref{tb:Taxonomy-MTL}).
While these works, all based on the convolutional neural network (CNN) backbone, show promising results, they remain challenged by negative task transfer, i.e., the degraded performance of certain tasks when learned jointly. To overcome this, Standley et. al~\cite{standley2019} developed subsets of complementary tasks where each of these subsets, when trained, can overcome negative task transfer. Being a handcrafted approach, ~\cite{standley2019} resulted in a large number of subsets comprising different task combinations.  %Moreover, the multitask affinities of the above models do not generalize to task-transfer learning or different domains. 
%Recent approaches to overcoming negative task transfer consist of either carefully choosing which tasks to train as an intermediate step before leveraging the inductive tasks for prediction on a target task~\cite{taskonomy2018,zamir2020consistency}, or grouping the proximate tasks based on empirical findings~\cite{standley2019, fifty2021efficiently}. While, in the former case, the intermediate tasks 
%are not required to perform well 
%are just auxiliary to the target task and are typically not evaluated, in the latter, subsets of proximate tasks are trained together resulting in multiple MTL models rather than a single one.

Following this,  IPT~\cite{IPT} was the first transformer-based multitask network aiming to solve low-level vision tasks after fine-tuning a large pre-trained network. Subsequently,~\cite {spatiotemporalMTL}, jointly addressed the tasks of object detection and semantic segmentation, and \cite{video-multitask-transformer} used a similar architecture for scene and action understanding in videos. Recently, Hu et.al.~\cite{hu2021unit} proposed a framework that tackles several language tasks but a single vision one. MulT~\cite{MulT} showed the superiority of vision transformers over CNN-based networks in modeling the multitask affinities via its shared attention mechanism, thereby solving all the tasks in a single model. While these transformer-based frameworks~\cite{MulT, IPT, hu2021unit, spatiotemporalMTL, video-multitask-transformer} clearly outperform the existing CNN-based multitasking methods, they are handcrafted and cannot be integrated into a different transformer backbone. %Moreover, the shared attention mechanism is conditioned on a \emph{single} reference task, which is selected empirically. 
By contrast, our vision transformer adapters can be integrated into an off-the-shelf vision transformer backbone, while learning task affinities based on \emph{all} the tasks in an automated manner.
\vspace{-13pt}
\paragraph{Learning generalizable task affinities.}
Taskonomy~\cite{taskonomy2018} studied the relationships between multiple visual tasks for transfer learning. Following this, a number of recent works have studied tasks relationships for transfer learning~\cite{achille2019task2vec,  dwivedi2019representation, fifty2021efficiently, DBLP:journals/corr/abs-1903-01092,standley2019, NEURIPS2019_f490c742}. These works analyze a network that is trained on a source task and is applied to a different target task. None of these mentioned works demonstrate a correlation between the transfer task affinities and the multitask affinities. 
%The generalizability of task affinities was extensively studied in~\cite{standley2019, fifty2021efficiently}, wherein Standley et al.~\cite{standley2019} found notable differences in the transfer task affinities and multitask affinities. These results were affirmed by~\cite{fifty2021efficiently}. Following this, MulT~\cite{MulT} studied the task inter-dependencies by designing a multitask transformer model instead of a CNN one and reported the superior performance of vision transformers for multitask learning. However, none of these methods, CNN-based or transformer-based, can solve the generalizability of task affinities, i.e., leveraging the multitask affinities for task transfer learning or for generalization to novel domains. 
To address this, we introduce our multitask vision transformer adapters that can successfully transfer the multitask affinities to novel tasks \emph{and} novel domains. 
\vspace{-22pt}
\paragraph{Vision Transformer Adapters.}
First introduced for language tasks to leverage knowledge embedded in large pre-trained transformers, adapters~\cite{houlsby-adapter} are trainable modules that are attached to specific locations of a pre-trained transformer network, providing a way to limit the number of parameters needed when confronted with a large number of tasks. This approach is also effective with pre-trained vision transformers that have rich semantic information~\cite{chen2022vitadapter, vit-det2022, li2021}. Specifically, Li et al.~\cite{vit-det2022, li2021} proposed ViT-based adapters for object detection, whereas Chen et al.~\cite{chen2022vitadapter} added feed-forward bottlenecks in every transformer block for the separate downstream tasks of object detection and semantic segmentation. Such methods, however, adapt to a \emph{single} downstream task. By contrast, we propose vision transformer adapters that can infer on \emph{multiple} dense-vision tasks in a single run in a parameter-efficient manner. To the best of our knowledge, only the prior works of~\cite{pmlr-v97-stickland19a, tay2021hypergrid}---both in the field of NLP---mix multitask learning and adapters within large pre-trained \emph{language} transformers by creating local task modules that are controlled by a global task-agnostic module. This approach, however, has the drawback of adding new non-shared parameters whenever a new task is added, thereby failing to generalize on novel tasks. By contrast, our vision transformer adapters share all parameters across the tasks and can re-modulate the existing weights when a new task is introduced. Moreover, the task affinities learned by our vision transformer adapters generalize to novel domains, unlike any existing work.  
\vspace{-13pt}
%based on our task representation optimization algorithm (Algorithm~\ref{alg:troa}).
\paragraph{Transformer Attention Mechanisms.}
While many works exploit the long-range dependencies of transformers by computing a task-specific attention~\cite{chen2021regionvit, chu2021Twins, wang2021pvtv2, wang2021crossformer, xu2021coscale, yang2021focal} and pairwise task attention for MTL~\cite{MulT}, none of these attention mechanisms learn task-affinities based on \emph{all} the tasks in an automated manner. We, therefore, introduce a task-adapted attention (TAA) mechanism that learns the task affinities by combining gradient-based task similarities with the attention ones. In essence, our TAA conditions the self-attention of the transformer backbone on the gradient-based task similarities.                   %                           
\begin{figure*}[ht!]                                                    
  \centering                                                          
  {\includegraphics[width=0.95\textwidth]{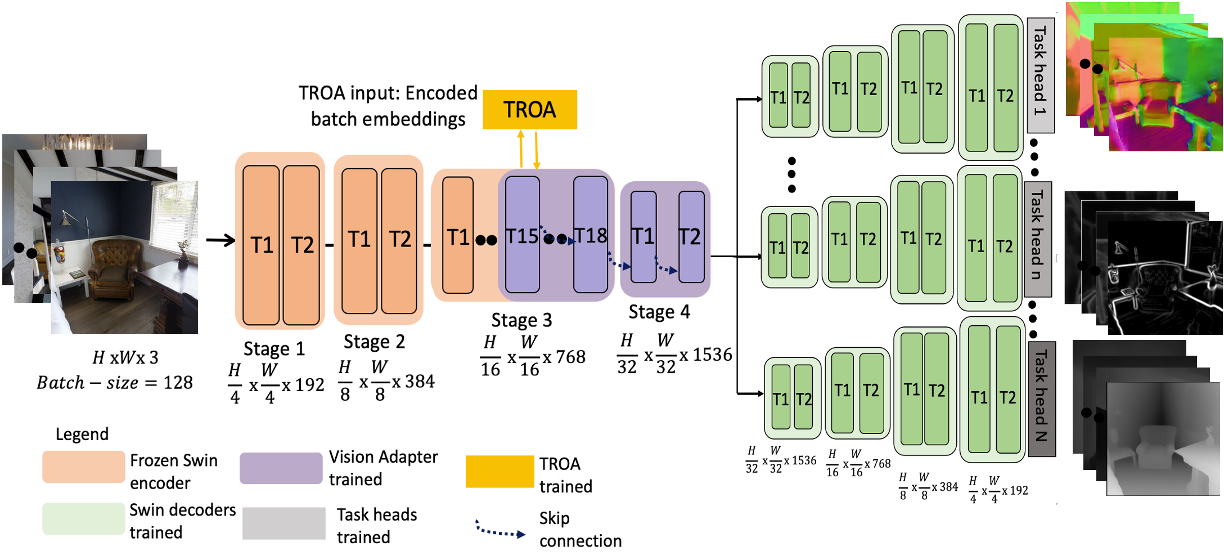}}
  \setlength{\abovecaptionskip}{2pt}\caption[Detailed overview of our method]{{\textbf{Detailed overview of our method.} %AVTaR learns the task inter-dependencies by leveraging the encoded task representations in our proposed vision adapter module.
  The frozen transformer encoder module (in \textcolor{lightsalmon}{orange}) extracts a shared representation of the input image, which is then utilized to learn the task affinities in our novel vision transformer adapters (in \textcolor{lightpastelpurple}{purple}). Each adapter layer uses gradient task similarity (TROA) (in \textcolor{schoolbusyellow}{yellow}) and Task-Adapted Attention (TAA) to learn the task affinities, which are communicated with skip connections (in \textcolor{yaleblue}{{blue}}) between consecutive adapter layers. The task embeddings are then decoded by the fully-supervised transformer decoders (in \textcolor{green}{{green}}) for the respective tasks. Note that the transformer decoders are shared but have different task heads (in \textcolor{taupegray}{grey}). For clarity, only three tasks are depicted here and TAA is explained in a separate figure. } }
  \vspace{-13pt}\label{fig:detailed-model-AVTaR}                    
  \end{figure*}                                                         

%\begin{figure}[h!]
%\centering{
%\begin{sideways}
%{\includegraphics[width=1.2\textwidth ]{images/ch5/avtar-updated-model.pdf}}
%\end{sideways}}
%\setlength{\abovecaptionskip}{2pt}
%\caption{{\textbf{Detailed overview of our AVTaR architecture.} AVTaR learns the task inter-dependencies by leveraging the encoded task representations in our proposed vision adapter module. The frozen Swin encoder module (in orange) embeds a shared representation of the input image, which is then learned along with the task representations in our novel vision adapter framework. This is then decoded by the fully-supervised transformer decoders (in green) for the respective tasks. Note that the transformer decoders have the same architecture but different task heads. For clarity, only three tasks are depicted here. } 
%}\label{fig:detailed-model-AVTaR}
%\end{figure}
%-------------------------------------------------------------------
%------------------------------------------

\vspace{-2pt}
\section{Method}
\label{sec:method}
%Let us now describe our approach for {\bf A}dapting {\bf V}isual {\bf Ta}sk {\bf R}epresentations (AVTaR) for multitask learning. At the heart of AVTaR is
Our novel vision transformer adapter method achieves predictions for a novel task or domain by learning transferable and generalizable task affinities. Our adapters leverage pre-trained vision transformer models that are readily and ubiquitously available. %, thanks to open-sourced models. 
While these easily available vision transformer models are pre-trained for classification on ImageNet, we aim to integrate them with multitasking. This calls for learning multitask affinities. To achieve this, within our vision adapters, we compute the gradient-based task similarity approach (TROA---Section~\ref{sec:troa}), that is, in turn, used by a novel task-adapted attention mechanism (TAA---Section~\ref{sec:TAA}).
%Concisely, the vision adapters learn task relationship by utilizing TAA that, in turn, utilizes TROA. 
This yields representations that are then normalized according to the task scales (Section~\ref{sec:TSN}),  
and finally decoded by the task-specific decoders and their respective task heads. Our overall framework is shown in Figure~\ref{fig:detailed-model-AVTaR}.  
%We present an analysis of AVTaR with different backbones, later in the chapter. 
Below, we discuss its different modules in detail.
Note that, although we present it using the Swin-B architecture, which is the most widely used backbone for dense prediction, our method can be integrated with any existing vision transformer backbone, such as ViT~\cite{dosovitskiy2021an}, Pyramid Transformer~\cite{wang2021pvtv2} or Focal Transformer~\cite{yang2021focal}, as will be shown by our experiments in the supplementary.
%using the Swin backbone, as it is the most widely used backbone for dense prediction. In the order of the workflow, we present 1) the encoder module, 2) the vision adapter, 3) the Task Representation Optimization Algorithm (TROA), 4) the Task-adapted Attention (TAA), 5) the Task-scaled Normalization (TSN), and 6) the decoder module with the task heads.
\subsection{Encoder Module}
\label{sec:encoder-module}
For the encoder, we adopt a pre-trained Swin-B V2~\cite{swin} model initialized with ImageNet-22K-trained weights. The encoder comprises four successive transformer stages employing a patch embedding that gradually decreases the resolution of the input image in a pyramidal manner while increasing the channel dimension. %along with a columnar sequence of transformer blocks. 
As shown in Figure~\ref{fig:detailed-model-AVTaR}, the first, second, and fourth stages have 2 transformer blocks while the third stage has 18 blocks. That is, following~\cite{resnet}, most of the computation is concentrated
in the third stage. Therefore, we propose to add trainable vision adapters on top of this stage --- specifically for transformer blocks 15 to 18 --- to leverage the rich embeddings it extracts. Nonetheless, to further reason about the high-level semantic information encoded in the final representation, we add two vision adapters for both transformer blocks in the fourth Swin stage. For any other vision transformer backbone~\cite{dosovitskiy2021an,wang2021pvtv2, yang2021focal}, our vision transformer adapters work best when integrated with layers comprising mid-level to high-level information.  %We now discuss our adapter module in detail. 
\subsection{Vision Transformer Adapter Module}
\begin{figure}[t]
\centering
{\includegraphics[width=0.76\linewidth]{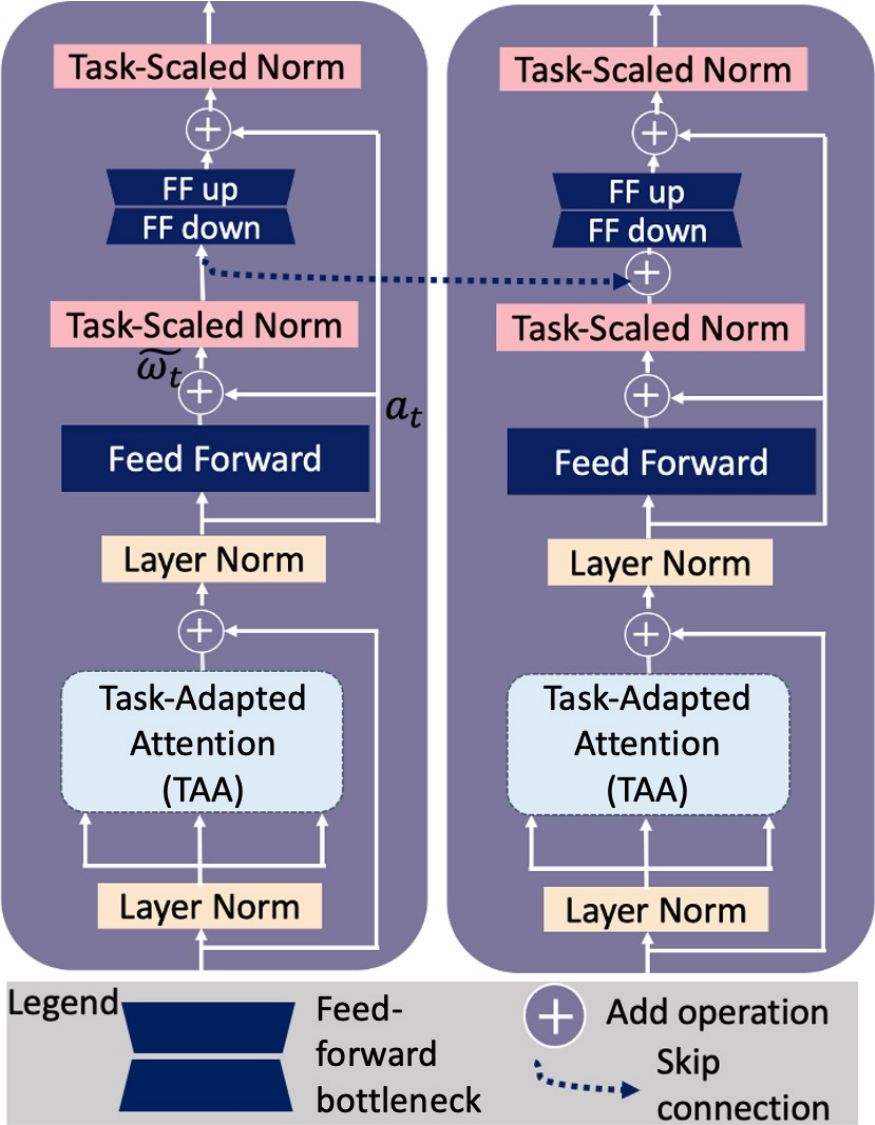}}
\caption[Detailed overview of our vision adapter]{{\textbf{Overview of our vision transformer adapter module.}  Our vision adapters learn transferable and generalizable task affinities in a parameter-efficient way. We show two blocks to depict the skip connectivity between them. }
}\label{fig:vision-adapter} %\vspace{-10pt}
\end{figure}

Our vision transformer adapters, depicted in Figure~\ref{fig:vision-adapter}, build on a sequence of transformer layers  of length consistent with the
Swin’s inter-window connectivity configurations. We connect the consequent adapter layers by using skip connections where the output of the previous layer is an input to the next layer.
This connectivity allows information to flow from preceding layers to later ones.
Within each vision adapter, different mechanisms are at play. %that we introduce in the following sections. 
In particular, these mechanisms are (i) TROA, which builds on~\cite{tawt}, and optimizes the task representations by computing their gradient
similarity; (ii) a novel task-adapted attention (TAA) module to combine gradient-based task affinities from TROA with attention ones;  and (iii) a novel task-scaled normalization (TSN) approach to balance the task scales. The adapter framework also relies on a bottleneck network consisting of a linear down-projection (FF down), a non-linearity, and a linear up-projection (FF up), used to decrease the number of parameters. In detail, for a batch of input images, where each image can be denoted as $X \subset \mathbb{R}^ {H \times W \times 3}$, the vision adapter encodes the representation of the batch of images as $\hat{\phi}$. These batch embeddings are normalized using a layer norm operation~\cite{layer-norm}. Once normalized, the embeddings are passed onto our novel TAA module which triggers the TROA mechanism within it to find the task similarities. We now explain the gradient-based task similarity computed by TROA.%, represented by $\hat{\omega_t}$ in Figure~\ref{fig:task-adapted-attention-detailed} and above.  %Once the task similarities are backpropagated by TROA, the similarities are modulated in a matrix to match the resolution of the corresponding self-attention matrix. 
%followed by a residual connection. It is positioned after the Task Adapted Multi-Head Attention (sec~\ref{sec:TAA}) layer and the feed-forward layer. In principle, the two-layer feed-forward bottleneck layer (as seen in Figure~\ref{fig:vision-adapter}) follows the task-scaled transformation (TSN as described in sec~\ref{sec:TSN}). The module is added to the topmost Transformer layers, in both stages 3 and 4, and uses the task-scaled norm (TSN). 
 %However, a recent study~\cite{tenney-NLP-adapter} in MTL states that 'different layers encode different task-specific information.' Therefore, to leverage the task-specific information from the different layers, we introduce a task-scaled normalization method, which we will detail in the later sections.
%In the remainder of this section, we describe the components of our vision adapter in more detail.
\label{sec:method-vision-adapter}
%\vspace{-15pt}
\mycomment{
%%%%%%%%%%%%%%%%% Goes in supplementary%%%%%%%%%%%%%%%%%%%
We visualize the task-adapted attention for each task in the 'S-D-N-E' setting and show that it differs from the existing self-attention mechanism in Figure~\ref{fig:TAA-visualization}. TAA is more task-specific compared with self-attention, thanks to its task conditioning from TROA. 
%We now explain the TROA mechanism in detail.
\begin{figure}[ht!]
    \centering
    \subcaptionbox{Test Image}
    {\includegraphics[width=0.4\linewidth]{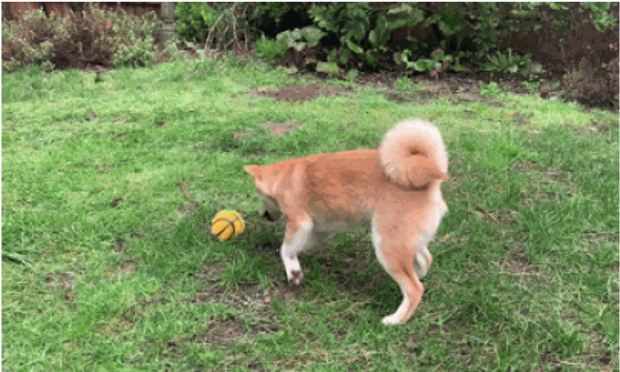}}
    \subcaptionbox{Self-attention}
    {\includegraphics[width=0.41\linewidth]{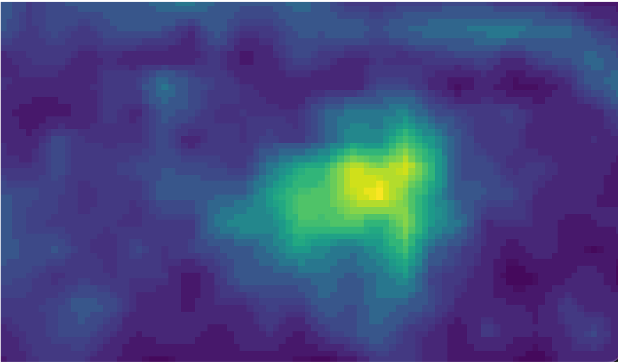}}\\
    \subcaptionbox{SemSeg TAA}
    {\includegraphics[width=0.4\linewidth]{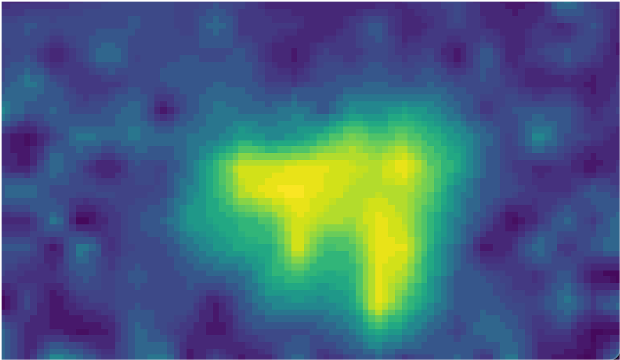}}
    \subcaptionbox{Depth TAA}
    {\includegraphics[width=0.4\linewidth]{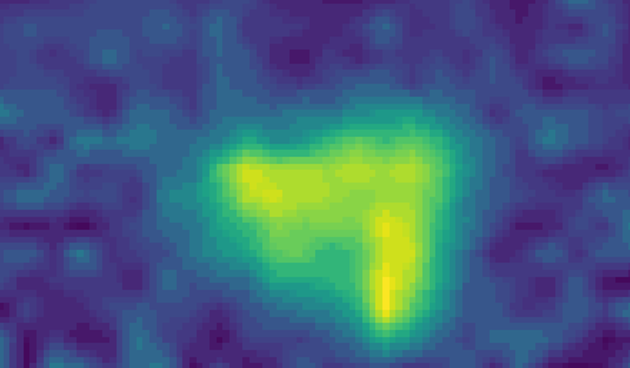}}
    \subcaptionbox{Normal TAA}
    {\includegraphics[width=0.4\linewidth]{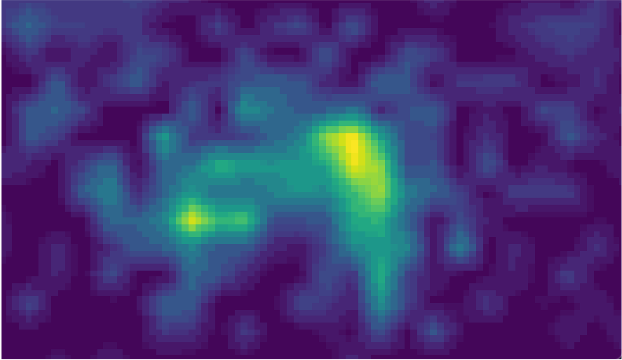}}
    \subcaptionbox{Edge TAA}
    {\includegraphics[width=0.4\linewidth]{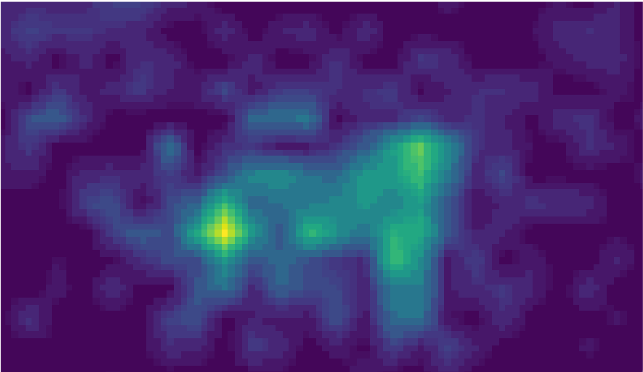}}
    \caption[Visualizing TAA versus the self-attention]{\textbf{Visualizing TAA versus the self-attention} of the encoder layer T18. We show that TAA has more task-specific attention compared to self-attention in the encoder. Here, our model is trained on MS-Coco~\cite{mscoco} with depth, surface normal, and edge labels from~\cite{midas, pseudo-surface-normal, rindnet}, respectively.}
    \label{fig:TAA-visualization}
\end{figure}
}
%\vspace{-22pt}
\subsubsection{Task Representation Optimization Algorithm (TROA)}
\label{sec:troa}
%Our TROA mechanism, which is implicitly leveraged by the task weight vector $\hat{\omega_t}$ in the TAA module discussed above, builds upon~\cite{Maml, tawt}. 
TROA computes a task representation $\hat{\theta}$ and a task affinity matrix $\hat{\omega}$ that depends on how correlated the tasks are. It is, therefore, named the Task Representation Optimization Algorithm, since it optimizes the task representations based on the task gradients. 
%the quality of the representation for each task.
%instead of assuming the existence of a single shared representation among all $N$ tasks. 
%This finer-grained notion of task affinities enables a better understanding of task-dependent and task-specific signals.
Specifically, this is computed by a gradient-based task affinity %, dubbed ${sim}_{t,n}$ in Algorithm~\ref{alg:troa}, 
which gives an interpretive measure of the influence of an inductive task $n \in [1, N]$ on a target task $t \in [1,N]$ based on the similarity between their learned representations.
%learnt through Adam~\cite{adam-w}. 
In TROA, we estimate these task affinities as the cosine similarity, dubbed ${sim}_{t,n}$ in Algorithm~\ref{alg:troa}, between the gradient of the inductive task and the gradient of the target task. This cosine similarity is computed for \emph{all} task combinations.  
Specifically, at iteration $i$,
%$'n^{th}$ inductive task and $'t^{th}$ target task, 
TROA starts with $\left(\theta^{i},\left\{m_{n}^{i}\right\}_{n=1}^{N}\right)$, i.e., the feature representation and the corresponding $N$ task-specific decoder functions. Upon making a forward pass, it learns the task weights by minimizing the overall multitask learning objective described by $\sum_{n=1}^{N} \omega_{t_n}^{i} \widehat{\mathcal{L}}_{n}\left(\theta, m_{n}\right)$ via Adam~\cite{adam-w}.
%in Algorithm~\ref{alg:troa}. 
We then calculate the cosine similarity between the task gradients to, ultimately, compute the task affinities. We employ a closed-form solution for the analytical weight update, $w_t^{i+1}$, given by the approximate mirror descent formula~\cite{mirror-descent} with a step-size $\kappa=1$. Note that the task weight vector $\hat{\omega}_t$ is updated via a combination of alternating minimization and mirror descent, where the minimization step prevents mode collapse if the task weights become equal. At the end of the $i^{th}$ iteration, we obtain a new task representation and a new weight vector $\hat{\omega_t}$ for the $t^{th}$ task, identifying its affinity with all the tasks. 
%\vspace{-8pt}
\setlength{\textfloatsep}{3pt}% Remove \textfloatsep
\begin{algorithm}[h!]
\small
\caption{TROA}\label{alg:troa}
\begin{algorithmic}
\Require  Batch embedding from vision adapter $\hat{\phi}$  %$\left\{P_{n}\right\}_{n=1}^{N}$
\Ensure Task representation $\widehat{\theta}$, task-specific decoder function $\widehat{m_{t}}$ and weight vector $\widehat{\omega_{t}}$ for the '$t$'th task . \\
\textbf{Initialize:} $\boldsymbol{\omega_t}^{1} \in \mathbb{R}^{N}$ uniformly,\\
$\qquad$ $\qquad$ $\theta^{1} \leftarrow \hat{\phi},\left\{m_{n}^{1}\right\}_{n=1}^{N} \subset \mathcal{M} ;$ \\
\textbf{for} $i=1, \ldots, I-1$ \textbf{do} \\ 
$\qquad$ \textbf{Starting With} $\left(\theta^{i},\left\{m_{n}^{i}\right\}_{n=1}^{N}\right)$; \textcolor{blue}{\% $i^{th}$ iteration.} \\
$\qquad$ \textbf{Run} a few steps of Adam to minimize 
$\qquad$ $\sum_{n=1}^{N} \omega_{t_n}^{i} \widehat{\mathcal{L}}_{n}\left(\theta, m_{n}\right)$
 and get $\left(\theta^{i+1},\left\{m_{n}^{i+1}\right\}_{n=1}^{N}\right)$;
\\
%$\qquad$ Use the approximate gradient $\nabla_{m} \widehat{\mathcal{L}_{1}}\left(\theta^{i+1}, m_{1}\right)$;
%\\
%$\qquad$ Run Adam from $m_{1}^{i}$ to get $m_{1}^{i+1} $; 
%\\
$\qquad$ \textbf{Run} ${sim}_{t,n}^i:=$ ${cossim}(\nabla_\theta \widehat{\mathcal{L}}_t(\theta^{i+1}, m_t^{i+1}), $ \\
$\qquad$ $\nabla_\theta \widehat{\mathcal{L}}_n\left(\theta^{i+1}, m_n^{i+1}\right))$; \textcolor{blue}{\% gradient similarity}.
\\
$\qquad$ \textbf{Update} $\omega_{t}^{i+1}:=\frac{\omega_{t}^{i} \exp \{-\kappa {sim}_{t,n}^{i}\}}{\sum_{n^{\prime}=1}^{N} \omega_{{n}^{\prime}}^{i} \exp\{-\kappa {sim}_{{t,n}^{\prime}}^{i}\}}$; \\
\textbf{end for} \\
\textbf{return} $\widehat{\theta}=\theta^{I}, \widehat{m_{t}}=m_{t}^{I},  \widehat{\omega_{t}}=\omega_{t}^{I}$ \\
\end{algorithmic}
\vspace{-12pt}
\end{algorithm}
\label{sec:TROA-visualization}
\begin{figure}[ht!]
\centering
{\includegraphics[ width=0.78\linewidth ]{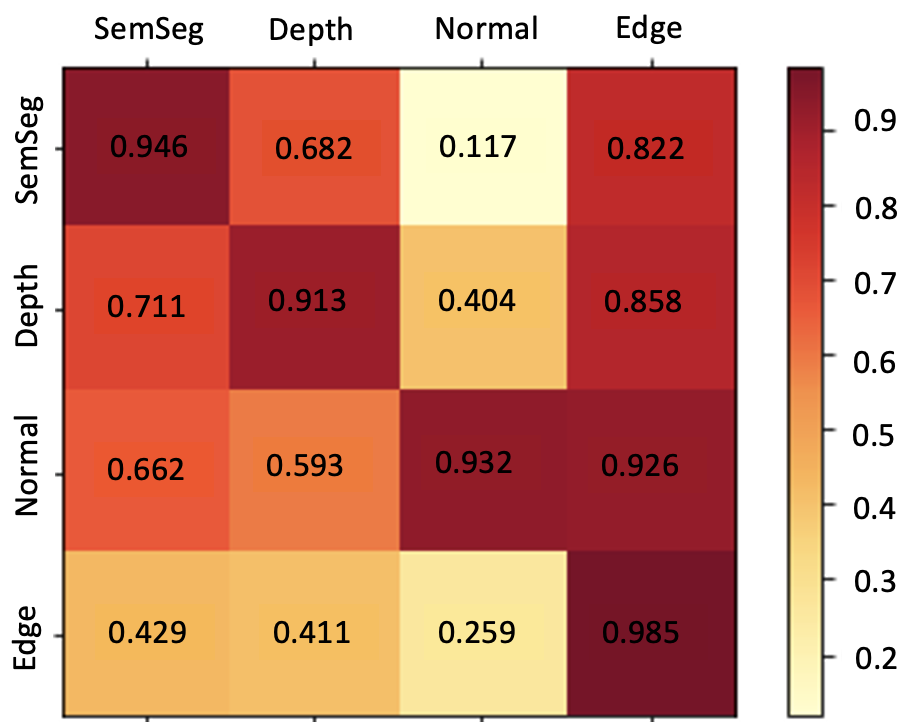}}
\setlength{\abovecaptionskip}{0pt}
\caption{{We show the \textbf{ gradient-based task affinities}, $\hat{\omega} \in \mathbb{R}^{N \times N}$ returned by TROA for N tasks.}
}\label{fig:weight-TROA}%\vspace{-10pt}
\end{figure}

In Figure~\ref{fig:weight-TROA} we show the task affinities from TROA when four tasks comprising semantic segmentation (SemSeg), depth, surface normal, and edges are jointly learned. We show that TROA learns a strong task affinity between the same task gradients, for example, segmentation with segmentation. This is a self-explanatory observation. Consequently, TROA also learns task affinities between proximate tasks such as segmentation and depth, and task affinities between non-proximate tasks such as segmentation and normal. Note that task dependence is assymetric, i.e. segmentation does not affect normal as normal effects segmentation. This is evidenced in Figure~\ref{fig:weight-TROA} and also by prior works~\cite{MulT, fifty2021efficiently, standley2019}. These task affinities are used by our novel task-adapted attention module as described in the following section. 

\subsubsection{Task-Adapted Attention (TAA)}
\label{sec:TAA}
\begin{figure}[ht]
\centering
{\includegraphics[width=1.0\linewidth]{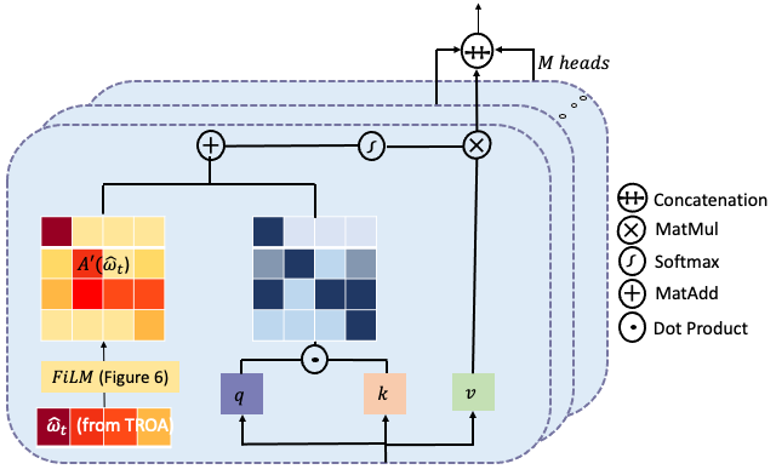}}
%\vspace{-3 pt}
\setlength{\abovecaptionskip}{0em}
\caption[Overview of our Task-Adapted Attention (TAA) mechanism]{Overview of our \textbf{Task-Adapted Attention (TAA)} mechanism that combines task affinities with image attention. Note that the process, in the foreground, is for a single attention head which is repeated for $M$ heads to give us the task-adapted multi-head attention.
}%\vspace{-11pt}
\label{fig:task-adapted-attention-detailed}
%\end{figure}
%\begin{figure}[ht!]
\centering
{\includegraphics[ width=1.0\linewidth ]{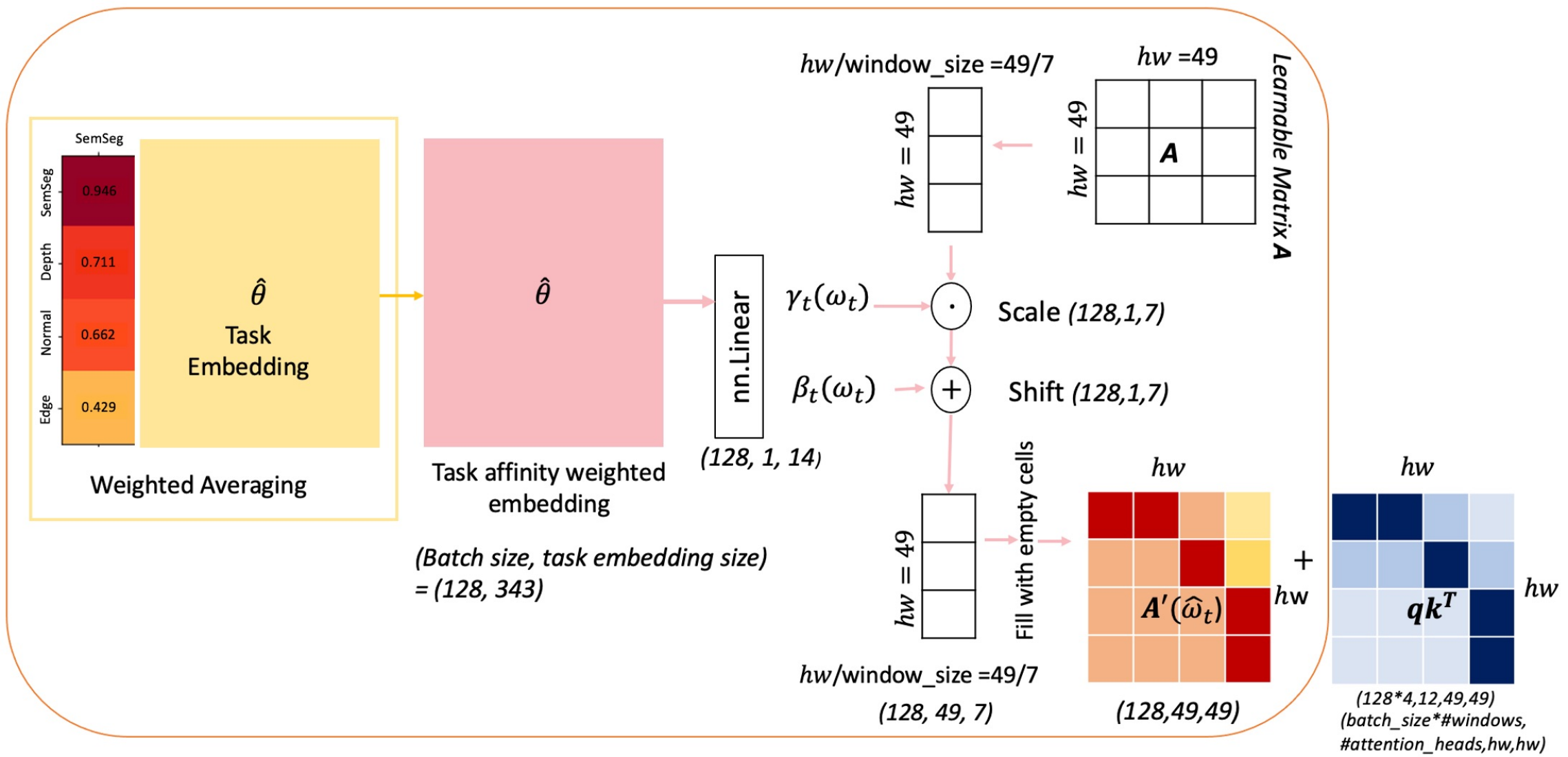}}
\setlength{\abovecaptionskip}{0pt}
\caption{\textbf{Detailed overview of Feature Wise Linear Modulation (FiLM)} which linearly shifts and scales tasks representations to match dimensions of the feature maps. The orange rectangular area is FiLM.  
}\label{fig:film}%\vspace{-10pt}
\end{figure}

Our task-adapted attention module, as shown in Figure~\ref{fig:task-adapted-attention-detailed}, combines gradient-based task affinities, represented by $\hat{\omega_t}$, with attention-based ones, represented by $q\cdot k^T$. 
The gradient-based task affinities, $\hat{\omega_t}$, are obtained from TROA as discussed in Section~\ref{sec:troa}. In a parallel branch, we extract a query $q$, key $k$, and value $v$ matrix from $\hat{\phi}$, following the standard approach in attention-based methods~\cite{attention-is-all-you-need}. The widely-known self-attention (SA)~\cite{attention-is-all-you-need} is computed as,
%\vspace{-10pt}
\begin{equation}
    \begin{aligned}
        SA(q, k, v) = softmax[ q\cdot k^T/\sqrt{c_{qkv}}]v\;,\\
    \end{aligned}
    \label{eq:self-attention}
\end{equation}
where ${c_{qkv}}$ is the channel dimension of the query, key, and value.
In contrast to this standard formulation, we condition the self-attention on the gradient-based task affinities, $\hat{\omega_t}$ from TROA (Section~\ref{sec:troa}). Formally, for a given task $t$, our task-adapted attention is
\vspace{-7pt}
\begin{eqnarray}
%\begin{aligned}
%\vspace{-5pt}
TAA(q, k, v, \hat{\omega_t}) =& \hspace{-0.3cm}softmax[A'(\hat{\omega_t}) + q\cdot k^T/\sqrt{c_{qkv}}]v\;,\label{eq:taa}\\
 &\hspace{-0.3cm}\text{where}\; A'(\hat{\omega_t}) = A\gamma_t(\hat\omega_t)+\beta_t(\hat\omega_t)\;.
%\end{aligned}
\label{eq:film}
\end{eqnarray}
Here, $\hat{\omega_t}$ is the $N$-dimensional vector of affinities for task $t$, i.e., the $t^{th}$ column of $\hat{\omega}$. As $\hat{\omega_t} \in \mathbb{R}^N$, we apply the widely-used Feature Wise
Linear Modulation~\cite{film} to match its dimension to the spatial dimensions of the feature maps and thus get $A'(\hat{\omega_t})$. %For a simple illustration of the Feature Wise Linear Modulation process, please refer to Figure~\ref{fig:film}. 
Specifically, the Feature Wise Linear Modulation (FiLM)~\cite{film} performs weighted averaging of the task representations w.r.t the task affinity weights, and then linearly shifts and scales the task representations as seen in Figure~\ref{fig:film}. It is more stable, unlike other dimension-matching techniques and we use this technique in our TAA module to match the dimensions of the affinity matrix $\hat{\omega_t}$ to the spatial dimensions of the feature maps and thus get $A'(\hat{\omega_t})$.

Formally, as indicated in Eq.~\ref{eq:film}, $A'(\hat{\omega_t})$ is computed by first mapping $\hat{\omega_t} \in \mathbb{R}^N$ to matrices of size $hw\times hw$ via the Feature Wise Linear Modulation~\cite{film} functions $\gamma_t, \beta_t$. The matrix output by $\gamma_t(\hat{\omega}^t)$ is then linearly transformed by a randomly-initialized learnable matrix $A$. 
%Once $\hat{\omega_t}\in \mathbb{R}^N$ has been mapped to $A'(\cdot)\in \mathbb{R}^{hw\times hw}$, 
Subsequently, we combine $A'(\cdot)$ with the $q\cdot k^T/\sqrt{c_{qkv}}$ matrix to obtain the TAA as in Eq.~\ref{eq:taa}.
%Ultimately, this process gives us task-specific features. 
In essence, for the $t^{th}$ task, the TAA module aids the query and key matrix to compute attention from the most similar tasks. Note that we generically use $h$ and $w$ to denote the spatial dimensions of the feature maps at different stages (i.e., $H/16, W/16$ for stage 3, and $H/32, W/32$ for stage 4). 
%we re-scale it using the widely-used Feature Wise
%Linear Modulation~\cite{film} functions. Specifically, we construct $A$ using 
%$H/16\times W/16$ 
%trainable and randomly initialized $H/16\times W/16$ dimensional matrices to match the spatial dimension of the query, key, and value. 
%While the original attention matrix depends on the hidden states of just the query, key, and value, $A'(\hat{\omega_t})$ is a learnable weight matrix that depends on the task weight vector $\hat{\omega_t}$ and  $\gamma_t, \beta_t$.
%Precisely, $\gamma_t, \beta_t$ maps $\mathbb{R}^N\longrightarrow\mathbb{R}^{HW\times HW}$.

The process described above corresponds to a single attention head.  In practice, as shown in Figure~\ref{fig:task-adapted-attention-detailed}, we perform this for $M$ heads, where $M=24$ and $48$ for the third and fourth stage, respectively, resulting in $M$ task-specific feature vectors. We then concatenate these vectors into a representation. 
Note that we apply the same procedure for the task-adapted attention in all the vision adapters. We defer qualitative comparisons of our TAA module w.r.t the typical self-attention (SA) to the supplementary.
 
 %We now combine this weight vector $\hat{\omega_t}$ (from TROA) with its corresponding task embedding, i.e., we compute the weighted average of the task embeddings. Thereafter, This process is illustrated below in Fig~\ref{fig:film}, where we use the widely-used Feature Wise Linear Modulation
 
Referring to Figure~\ref{fig:vision-adapter}, the output of the task-adapted multi-head attention is employed in a residual connection followed by a layer norm operation, a feed-forward network, and another residual addition resulting in an  $\tilde{{\omega_t}}$ matrix. This matrix is scaled w.r.t. the task $t$. Our vision adapters achieve task-scaling by employing the Task-Scaled Norm, which is described in the following section. %sec~\ref{sec:TSN}.
\vspace{-10pt}
\subsubsection{Task-Scaled Normalization (TSN)}
\label{sec:TSN}
TSN balances the different scales of the tasks. Balancing the task scales is necessary to avoid learning interference in a multitasking framework~\cite{Vandenhende_2021}. To this end, inspired by the Conditional Batch Normalization~\cite{conditional-BN} strategy, we formulate TSN as follows.
%we condition the task-specific activations $a_t \in \mathbb{R}^{hw\times 768}$, obtained after the residual connections (c.f. Figure~\ref{fig:vision-adapter}), on the task affinity weights. 
For task $t; t \in {1, \ldots,N}$, 
\vspace{-10pt}
\begin{equation}
\begin{aligned}
{TSN}_t=\frac{1}{\sigma} *\left({a}_t-\mu\right) * \hat{\gamma}_t\left(\tilde{{\omega}}_t\right)+\beta_t\left(\tilde{{\omega}}_t\right)\;,\\
  \text{where}\;  \hat{\gamma}_t(\tilde{{\omega}}_t)={\gamma'} \gamma_t\left(\tilde{{\omega}}_t\right)+{\beta'}\;,
\end{aligned}
\end{equation} 
$a_t$, as shown in Figure~\ref{fig:vision-adapter}, is the task-specific activation obtained from the residual connection, and $\tilde{{\omega_t}}$ is the summed  output of the feed-forward network with the residual connection. Furthermore, $\mu$ and $\sigma$ are the mean and the variance of all the inputs within each layer, as defined in~\cite{layer-norm}, and ${\gamma'}$ and ${\beta'}$ are the Swin's Layer Normalization weight and bias functions.
Our TSN mechanism contrasts with Layer Norm in the following two ways: 1) While the Layer Norm weight and bias functions are kept fixed, the TSN ones are trained; 2) while Layer Norm normalizes the input across features, TSN modulates the normalization output based on the task weights.
%the Therefore, we employ the methodology of Conditional Batch Normalization~\cite{conditional-BN} to modulate

\subsection{Decoder Module} Leveraging the idea in~\cite{swin, MulT}, our decoder architecture comprises four stages, each containing 2 sequential transformer blocks for a total of 8. In each stage, the two sequential transformer blocks alternate regular and shifted window attention mechanisms, as in~\cite{swin}. Between each stage, we employ an upsampling layer to double the spatial resolution and halve the channel dimension; we therefore adjust the number of attention heads accordingly to 48, 24, 12, and 6, in the first, second, third, and fourth stage, respectively.  Unlike in~\cite{MulT}, where the lower-resolution stages of the decoder are guided by the higher-level deeper encoded features and vice versa, our model employs trainable vision adapters to guide the stages of the decoder in a sequential manner. 
%Note that the first transformer block in each stage of the decoder uses a regular window partitioning while the second uses a shifted window partitioning; this can easily be extended to using a longer sequence of transformer blocks, as long as the length is a multiple of 2, which makes it possible to alternate between the two configurations. 
To perform predictions on multiple tasks, we share the vision adapters across all tasks and use task-specific decoders with the same architecture but different parameter values. We then simply append task-specific heads to the decoder. 
\\
\textbf{Task Heads and Training.} The decoded feature representations are passed into the linear task-specific heads, such that the task head outputs an $H\times W\times K$ map, where $H$, $W$, and $K$ are the input image dimensions and the task-specific channels, respectively. We jointly train the adapters and the decoders by employing a linear combination of the task losses, where the losses are calculated between the ground truth and predictions for each task. To maintain consistency with the baselines~\cite{MulT, xtam, zamir2020consistency}, we use the cross-entropy for segmentation, the rotated loss for depth, and the $l_1$ loss for surface normal and 2D edges, respectively. 

%The entire process depicted above was for a given task $t$ at a given iteration $i$. The process is repeated for all tasks and until the total linear combination of the task losses converges. 
\begin{table*}[t]
\setlength\tabcolsep{3pt}
\centering
\scalebox{0.68}{
\begin{tabular}{llllllllllllllll}
%\toprule
\multicolumn{11}{l}{~~~~~~~~~~~~~~~~~~~~~~~~~~~~~~~~~~~~~~~~~~~~~~~~~~~~~~~~ ~ ~ ~ ~ ~ ~ ~ ~ ~ ~ ~ \textbf{Quantitative results on Taskonomy}~\cite{taskonomy2018}} & \multicolumn{5}{l}{~~~~~~~~~\textbf{Quantitative results on Cityscapes}~\cite{Cordts2016Cityscapes}}\\ \cmidrule(lr){3-11} \cmidrule(lr){12-16}
\multicolumn{1}{l}{}&\multicolumn{1}{l}{}                                   & \multicolumn{2}{l}{~~~~~~~~\textbf{\textit{'S-D'}}}                                                                                                                             & \multicolumn{3}{l}{~~~~~~~~~~~~~~\textbf{\textit{'S-D-N'}}}                                                                                                                                                                                                             &       \multicolumn{4}{l}{~~~~~~~~~~~~~~~~~~\textbf{\textit{'S-D-N-E'}}}   &  \multicolumn{2}{l}{~~~~~~~~\textbf{\textit{'S-D'}}}                                                                                                                             & \multicolumn{3}{l}{~~~~~~~~~~~~~~\textbf{\textit{'S-D-N'}}}                                                                                                                                                          \\
\cmidrule(lr){1-2}\cmidrule(lr){3-11} \cmidrule(lr){12-16}
&\multicolumn{1}{c}{\multirow{-2}{*}{\textbf{Methods}}} & \cellcolor[HTML]{FFCCC9}\begin{tabular}[c]{@{}l@{}}SemSeg\\ mIoU\%$\uparrow$\end{tabular} & \cellcolor[HTML]{DAE8FC}\begin{tabular}[c]{@{}l@{}}Depth\\ RMSE$\downarrow$\end{tabular} & \cellcolor[HTML]{FFCCC9}\begin{tabular}[c]{@{}l@{}}SemSeg\\ mIoU\%$\uparrow$\end{tabular} & \cellcolor[HTML]{DAE8FC}\begin{tabular}[c]{@{}l@{}}Depth\\ RMSE$\downarrow$\end{tabular} & \cellcolor[HTML]{FFFFC7}\begin{tabular}[c]{@{}l@{}}Normal\\ mErr.$\downarrow$\end{tabular} & \cellcolor[HTML]{FFCCC9}\begin{tabular}[c]{@{}l@{}}SemSeg\\ mIoU\%$\uparrow$\end{tabular} & \cellcolor[HTML]{DAE8FC}\begin{tabular}[c]{@{}l@{}}Depth\\ RMSE$\downarrow$\end{tabular} & \cellcolor[HTML]{FFFFC7}\begin{tabular}[c]{@{}l@{}}Normal\\ mErr. $\downarrow$\end{tabular} & \cellcolor[HTML]{C8E685}\begin{tabular}[c]{@{}l@{}}Edges\\ F1\%$\uparrow$\end{tabular} & \cellcolor[HTML]{FFCCC9}\begin{tabular}[c]{@{}l@{}}SemSeg\\ mIoU\%$\uparrow$\end{tabular} & \cellcolor[HTML]{DAE8FC}\begin{tabular}[c]{@{}l@{}}Depth\\ RMSE$\downarrow$\end{tabular} & \cellcolor[HTML]{FFCCC9}\begin{tabular}[c]{@{}l@{}}SemSeg\\ mIoU\%$\uparrow$\end{tabular} & \cellcolor[HTML]{DAE8FC}\begin{tabular}[c]{@{}l@{}}Depth\\ RMSE$\downarrow$\end{tabular} & \cellcolor[HTML]{FFFFC7}\begin{tabular}[c]{@{}l@{}}Normal\\ mErr.$\downarrow$\end{tabular}\\
\cmidrule(lr){1-2}\cmidrule(lr){3-11} \cmidrule(lr){12-16}
\multirow{9}{*}{CNN}&MTL-baseline~\cite{Vandenhende_2021}        &41.22 &0.5640 &45.16 &0.5398 &29.30 &47.64 &0.5091 &25.11 &53.96  & 70.66 & 6.726  & 70.93  & 6.721  &43.60
                                                                    \\
&Cross-stitch~\cite{cross-stitch}  & 30.83    &  0.7780   & 32.84 & 0.7530  & 34.11  & 34.91 & 0.6980  & 32.04   &   37.58 & 50.33 &7.683  &54.99 &7.311 & 44.10                                                                         \\    
&MTAN~\cite{endtoendMTL} &  32.94   & 0.6800    & 35.79 &  0.6440  & 32.25  & 38.88  & 0.6030   & 30.55   &  40.19  &  53.86 &  7.318 & 57.23 & 7.050  & 42.09 \\  
&TTNet~\cite{ttnet}&  42.75   &  0.5270   & 45.85 & 0.5011   & 24.88  & 52.79  & 0.4872   & 24.11   &  56.10  & 71.00  &  6.655 & 71.40 & 6.511  & 41.23 \\  
&Taskonomy~\cite{taskonomy2018}&  42.00   & 0.5633    & 43.11 &  0.5391  & 29.67  & 48.40  &  0.5086  & 26.62   &  53.99  & 60.02  & 7.204  & 63.41 & 7.044  & 41.91 \\  
&TSwitch~\cite{taskswitchnet}& 42.79    & 0.5222    &  45.91& 0.5007   & 24.82  & 52.81  & 0.4873   & 24.00   & 56.72   & 71.13  & 6.634  & 71.45 & 6.509  & 41.00 \\  
&Consistency~\cite{zamir2020consistency}&   42.46 &  0.5293 &  45.69 & 0.5013 & 27.22  & 52.55  &  0.4899  &24.02 & 55.50    & 70.23 & 6.671 & 71.67 & 6.575  &  41.46                                                                       \\
&  XTAM~\cite{xtam}  & 43.24 &0.4966 &45.77 &0.4888 &25.05 &52.71 &0.4701 &22.19 &58.10 & 75.02   & 6.653   &  75.92 & 6.419   &40.39
                                                                           \\
& TAWT~\cite{tawt}   & 44.07 & \underline{0.4935} &48.92 &0.4833 &24.86 &53.15 &0.4658 &22.02 &61.77& 74.95 & 6.649 &76.08   & 6.407  &40.05
                                        \\
                                        \cmidrule(lr){1-2}\cmidrule(lr){3-11} \cmidrule(lr){12-16}
\multirow{3}{*}{Transformer}&ST-MTL~\cite{spatiotemporalMTL}& 45.12    &  0.4990  & 49.34 & 0.4750   & 23.11  & 53.17  & 0.4600   & 21.80   & 62.85   & 75.01  & 6.655  & 76.13 & 6.429  & 39.95 \\  
&MulT~\cite{MulT}   & \underline{49.73} &0.4981 &\underline{52.13} & \underline{0.4501} & \underline{21.86} &\underline{54.04} &\underline{0.4429} &\underline{20.10} &\underline{65.62}& 
\underline{76.05} & 6.650 &\underline{77.50} & \underline{6.391} & \underline{39.84} 
                                                                     \\

&\cellcolor[HTML]{EFEFEF}\textbf{Our}   & \cellcolor[HTML]{EFEFEF}\textbf{52.46} &\cellcolor[HTML]{EFEFEF}\textbf{0.4524} &\cellcolor[HTML]{EFEFEF}\textbf{57.03} &\cellcolor[HTML]{EFEFEF}\textbf{0.4291} &\cellcolor[HTML]{EFEFEF}\textbf{19.46} &\cellcolor[HTML]{EFEFEF}\textbf{60.80} &\cellcolor[HTML]{EFEFEF}\textbf{0.3903} &\cellcolor[HTML]{EFEFEF}\textbf{17.13} &\cellcolor[HTML]{EFEFEF}\textbf{71.09} & \cellcolor[HTML]{EFEFEF}
\textbf{78.00} &\cellcolor[HTML]{EFEFEF}\textbf{6.503} &\cellcolor[HTML]{EFEFEF}\textbf{80.55} &\cellcolor[HTML]{EFEFEF}\textbf{6.307} &\cellcolor[HTML]{EFEFEF}\textbf{39.05}
 \\          
 %\cmidrule(lr){1-1}\cmidrule(lr){2-10} \cmidrule(lr){11-15} 
% \multicolumn{10}{l}{~~~~~~~~~~~~~~~~~~~~~~~~~~~~~~~~~~~~~~~~~~~~~~~~~~~~~~~~ ~ ~ ~ ~ ~ ~ ~ ~ ~ ~ ~ ~ \textbf{Quantitative results on NYUDv2}~\cite{NYU}} & \multicolumn{5}{l}{~~~~~~~~~\textbf{Quantitative results on Cityscapes}~\cite{Cordts2016Cityscapes}}\\
%\cmidrule(lr){1-1}\cmidrule(lr){2-3}\cmidrule(lr){4-6}\cmidrule(lr){7-10}\cmidrule(lr){11-12}\cmidrule(lr){13-15}
% STL~\cite{unet}  &  38.70&0.635&38.70&0.635&36.90&38.70&0.635&36.90&54.90  &  67.93&6.683&61.93&6.683&44.10
                                                                         \\
%MTL-baseline~\cite{Vandenhende_2021}        &39.44&0.638&39.90&0.642&36.07&39.70&0.636&36.10&55.11 &70.66&6.726&70.93&6.721&43.60

%                                                                    \\
%Consistency~\cite{zamir2020consistency} &35.38 &0.642&36.14&0.630&35.80&39.79&0.613&35.58&57.19& 70.23&6.671&71.67&6.575&43.46
 %                                                   \\
 %XTAM~\cite{xtam}   & 38.93&0.616&40.11&0.611&34.78&40.24&0.606&34.44&60.55& 75.02&6.653&75.92&6.419&40.39
  %                                                                         \\
% TAWT$+$MTL-baseline~\cite{tawt,Vandenhende_2021}   & 38.99&0.604&40.28&0.598&33.72&40.84&0.593&33.38&61.12 & 74.95& \underline{6.649} &76.08&6.407&40.05
%                                        \\

%MulT~\cite{MulT}   & \underline{39.50} &\underline{0.592}
% &\underline{41.48} & \underline{0.569} & \underline{31.55} &\underline{43.16} &\underline{0.554} &\underline{31.25} &\underline{64.66}& \underline{76.05} &6.650 &\underline{77.50} & \underline{6.391} & \underline{39.84} 
 %                                                                    \\
%\rowcolor[HTML]{EFEFEF}
%\textbf{Our}   & \textbf{49.93} &\textbf{0.519} &\textbf{50.60} &\textbf{0.504} &\textbf{30.54} &\textbf{52.95} &\textbf{0.489} &\textbf{28.68} &\textbf{70.02}& \textbf{78.00} &\textbf{6.503} &\textbf{80.55} &\textbf{6.307} &\textbf{39.05}
% \\                   
\end{tabular}}
\setlength{\abovecaptionskip}{0pt}
\caption{\textbf{Quantitative comparison} on the Taskonomy~\cite{taskonomy2018} and Cityscapes~\cite{Cordts2016Cityscapes} benchmarks for different multitask settings of \textit{'S-D'}, \textit{'S-D-N'} and \textit{'S-D-N-E'}. Our model consistently outperforms both CNN-based and vision transformer-based MTL baselines. We show that adding more tasks improves their respective performances based on their task affinities. Bold and underlined values show the best and
second-best results, respectively. }
   \label{tb:AVTaR-taxonomy-nyudv2-results}%
\vspace{-10pt}
\end{table*}

\section{Experiments and Results}
%We compare our method against multiple baselines on five multitask datasets. 
Considering the number of experiments and results we report, we
highlight in the main paper one consistent set of results and defer additional qualitative and quantitative results to the supplementary material. For easier comparison, we only report here the results of our vision adapters with the SWIN-B transformer backbone. Results with
other transformer backbones like ViT~\cite{dosovitskiy2021an}, Pyramid Transformer~\cite{wang2021pvtv2}, and Focal Transformer~\cite{yang2021focal} are also in the supplementary along with descriptions of the datasets, baselines, and evaluation metrics that we use. 
\vspace{-10pt}
\paragraph{Experimental Setup.} The experiments were performed using the following 4 dense prediction tasks: semantic segmentation (\textit{S}), depth (zbuffer) (\textit{D}), surface normal (\textit{N}), and 2D (Sobel) texture edges (\textit{E}). We report results in the following settings: 1) The \emph{MultiTask Learning (MTL) setting}%, where we compare our method against baselines that encompass single-task vision transformers, multitask CNN- and vision transformer-based methods
; 2) the \emph{ Zero-shot Task Transfer setting}%, where we compare our method against task transfer learning baselines (CNN- and vision transformer-based), all of which use either a reference task or an intermediate task signal to predict the target task output
; 3) the \emph{Unsupervised Domain Adaptation (UDA) setting}%, where we compare our method against CNN- and vision transformer-based UDA baselines
; and 4) \emph{Generalization to Novel Domains}.%, where we compare our method's generalizability to novel domains without any fine-tuning.  %by either training on Synthia~\cite{synthia} and adapting to Cityscapes~\cite{Cordts2016Cityscapes}, or training on Vkitti2~\cite{vkitti2} and adapting to Cityscapes~\cite{Cordts2016Cityscapes} (Table~\ref{tb:AVTaR-uda-results-syn2cityscapes}).

For the MTL setting, the methods are jointly trained in a fully-supervised manner on task combinations 
such as \textit{‘S-D’} (segmentation $+$ depth), \textit{‘S-D-N’} (segmentation $+$ depth $+$ normal) and \textit{‘S-D-N-E’} (segmentation $+$ depth $+$ normal $+$ edges) on the Taskonomy benchmark~\cite{taskonomy2018} and the NYUDv2 benchmark~\cite{NYU}.  
We also evaluate all models on Synthia~\cite{synthia}, Cityscapes~\cite{Cordts2016Cityscapes}, and Vkitti2~\cite{vkitti2} for the task combinations \textit{‘S-D’} and  \textit{‘S-D-N’}.

For the Zero-shot Task Transfer setting, all the methods are first trained on Vkitti2~\cite{vkitti2} and then fine-tuned on Cityscapes or Synthia using the ground-truth segmentation labels in the \textit{'S-D'} case, and the ground-truth segmentation and depth labels in the \textit{'S-D-N'} one.  %(Table~\ref{tb:zero-shot-task-transfer-results}). 
%In our implementation, we train with a batch size of 128 on 4 A100 40GB GPUs in a distributed fashion, using PyTorch. We use the Adam optimizer~\cite{adam-w} with a learning rate of $5.0e-5$ and a warm-up cosine learning rate schedule. 
%The optimizer updates the model parameters based on gradients from the task losses (also mentioned in Algorithm~\ref{alg:troa}). 

For the UDA setting, we deal with distribution shifts between a source domain, with labeled data, and a target domain, in which only unlabeled data is available for training. All models are trained with the source domain labels of Vkitti2~\cite{vkitti2} with the models \emph{aware} of the images in both the source~\cite{vkitti2} and target~\cite{Cordts2016Cityscapes} domain. %We also report experiments on Synthia~\cite{synthia} as source domain and Cityscapes~\cite{Cordts2016Cityscapes} as target domain in the supplementary material.

To evaluate the generalizability of our learned task affinities to novel domains, wherein the model is \emph{unaware} of the images in the target domain, 
%we conduct additional experiments. W
we train the models on Taskonomy~\cite{taskonomy2018} and apply them to NYUDv2~\cite{NYU} without any fine-tuning. Furthermore, we train our model on MS-Coco~\cite{mscoco} and apply them to a highly disparate comics domain that differs in styles and contents from real-world imagery. 
%We observe that our task affinities generalize well to novel domains even without fine-tuning. Results on generalization on comics are in the supplementary.
 
\subsection{Qualitative Results}
\begin{figure}[t]
\centering
%{\includegraphics[ height= 5cm, width=8.3cm]{main/images/taskonomy-qr-new.pdf}}
%{\includegraphics[width=0.48\textwidth]{main/images/taskonomy-qr-new.pdf}}
{\includegraphics[width=0.48\textwidth]{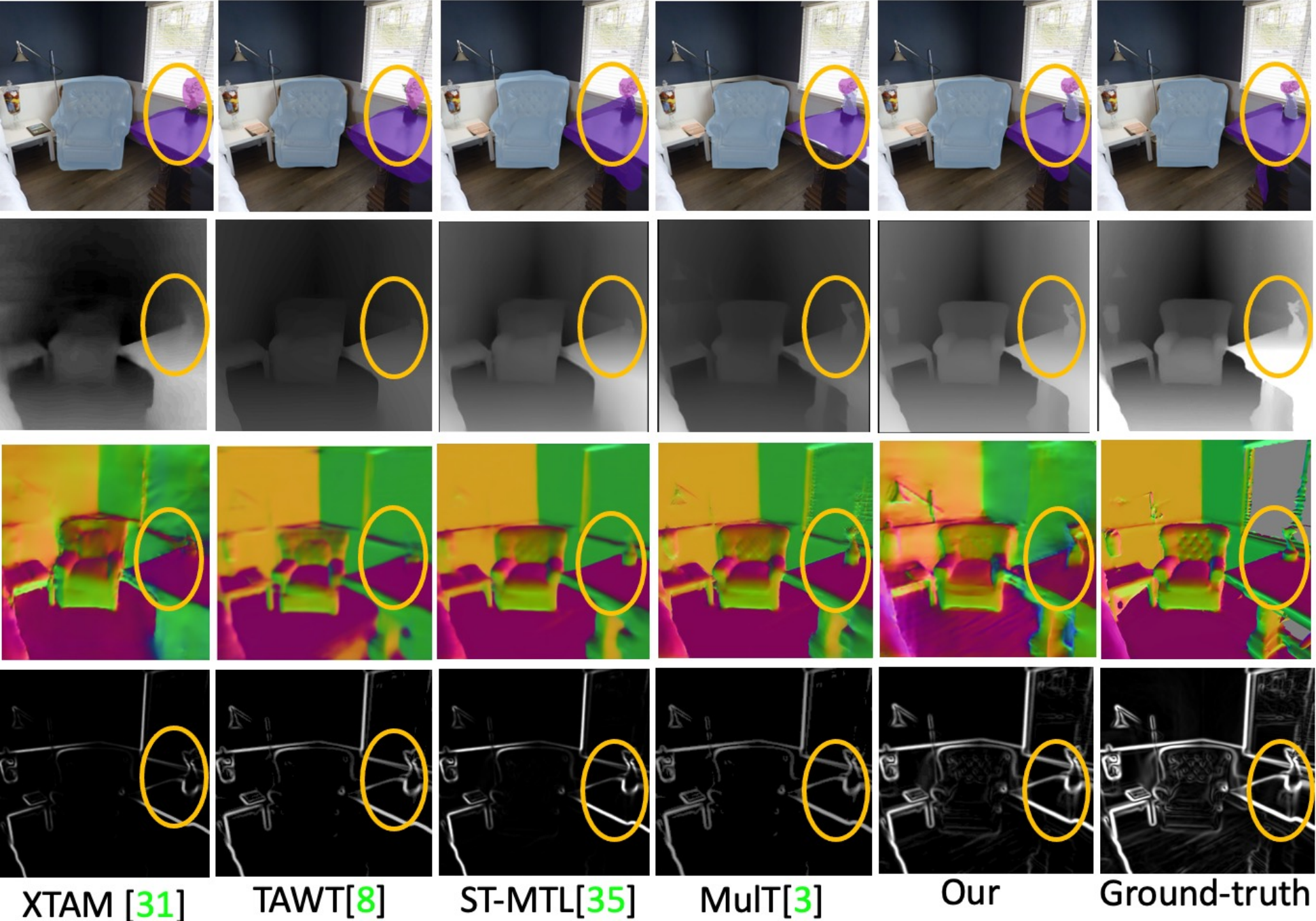}}
%\vspace{-11.3 pt}
\caption{\textbf{Qualitative comparison} on Taskonomy benchmark~\cite{taskonomy2018} for \textit{'S-D-N-E'}. From top to bottom, we show results on segmentation, depth, surface normal, and edges. %We show, from left to right,  results using STL~\cite{unet}, MTL-baseline~\cite{Vandenhende_2021}, Consistency~\cite{zamir2020consistency}, XTAM~\cite{xtam}, TAWT with MTL-baseline~\cite{tawt, Vandenhende_2021}, MulT~\cite{MulT}, \textbf{Our} and ground-truth in the multitask setting. 
Our model outperforms all the multitask baselines. We report the best-performing methods from Table~\ref{tb:AVTaR-taxonomy-nyudv2-results}. Best seen on screen and zoomed within the yellow circled regions.
}\label{fig:qualitative-result-taskonomybenchmark}%\vspace{-5pt}
\end{figure}
We qualitatively compare the results of our model with different baselines~\cite{MulT, tawt, xtam, spatiotemporalMTL} in Figure~\ref{fig:qualitative-result-taskonomybenchmark} for the tasks of segmentation, depth, normal and edge prediction on the Taskonomy~\cite{taskonomy2018} benchmark for the MTL setting. 
%The results show the performance of the different multitasking networks on all four vision tasks except for the single task baseline which is trained on the respective tasks. 
%Furthermore, in Figure~\ref{fig:uda-results-syntocs} we show the qualitative results for unsupervised domain adaptation for Synthia$\rightarrow$Cityscapes for the tasks of segmentation and depth. In both cases,
Our method yields higher-quality predictions than all the baselines. This is noticeable when looking at thin elements (e.g., flower vases, and table lamps) and object contours. The visuals correspond to the quantitative analysis. More qualitative results are provided in the supplementary material.
%\vspace{-5pt}
\subsection{Quantitative Results}
\vspace{-2pt}
\textbf{Multitask Setting:} Table~\ref{tb:AVTaR-taxonomy-nyudv2-results} reports our main experimental results on two datasets where all models are initialized with the pre-trained
ImageNet 22k model weights for a fair comparison. The baselines are selected based on their encoder-focused architectures that compare with our encoder-focused framework, as well as their task-affinity generalization, shown in Table~\ref{tb:Taxonomy-MTL}. %, across three multitask setups: \textit{‘S-D’}, \textit{‘S-D-N’} and \textit{‘S-D-N-E’} for the indoor scene benchmarks~\cite{taskonomy2018, NYU} and two multitask setups: \textit{‘S-D’} and \textit{‘S-D-N’} for the urban scene ones~\cite{synthia, Cordts2016Cityscapes}. 
On Taskonomy~\cite{taskonomy2018} with the \textit{'S-D-N-E'} labels, our method outperforms both the CNN-based~\cite{tawt, endtoendMTL, xtam, cross-stitch, ttnet, Vandenhende_2021, zamir2020consistency, taskonomy2018} and vision transformer-based MTL~\cite{spatiotemporalMTL, MulT} baselines by a considerable margin, showing the benefit of leveraging task-adapted attention. %instead of using attention from a single task, as done in MulT~\cite{MulT}. 
%The same trend can be seen across NYUDv2~\cite{NYU}, Synthia~\cite{synthia}, and 
The same trend can be seen on Cityscapes~\cite{Cordts2016Cityscapes}. Furthermore, we observe an increase in performance across all tasks with the addition of more tasks. This evidences the benefit of injecting additional geometrical cues in the form of surface normal or edges, respectively, to help the other tasks. We do not evaluate on IPT~\cite{IPT} because it was built to specifically solve deraining, denoising, and super-resolution. We also do not compare Vid-MTL~\cite{video-multitask-transformer} or UniT~\cite{hu2021unit} as they cater to different modalities of learning such as video and text, respectively. The NYUDv2~\cite{NYU}, Synthia~\cite{synthia}, Vkitti2~\cite{vkitti2} MTL results are provided in the supplementary material. 
\vspace{-15pt}
\paragraph{Zero-Shot Task Transfer:} In Table~\ref{tb:zero-shot-task-transfer-results}, we apply the models trained on Vkitti2~\cite{vkitti2} to Cityscapes~\cite{Cordts2016Cityscapes}. As the name suggests, a model that infers a zero-shot task is not trained with \emph{any} labels corresponding to that task. However, it should have a notion of the zero-shot task, which it leverages from the trained Vkitti2 labels. As shown in Table~\ref{tb:zero-shot-task-transfer-results}, our method outperforms all the baselines on zero-shot depth prediction and zero-shot normal prediction on Cityscapes by at %least 0.268, and 3.25 points, respectively, on the Synthia benchmark~\cite{synthia}. Furthermore,
least 0.196, and 1.59 points, respectively. 
%The performance gain in Synthia over that of Cityscapes is justified as Synthia, being a synthetic dataset, has perfect segmentation and depth labels. Nonetheless, our model outperforms all the baselines across both the benchmarks and tasks. 
For the zero-shot task transfer experiments, we compare with the baselines that have investigated task transfer learning~\cite{tawt, ttnet, zamir2020consistency, taskonomy2018}, shown in Table~\ref{tb:Taxonomy-MTL}. We also compare with methods that use the same transformer backbone such as Vanilla MTL Swin~\cite{swin} and MulT~\cite{MulT}. 
%reference task attention for MulT~\cite{MulT} is from segmentation in the \textit{'S-D'} setting and from depth in the \textit{'S-D-N'} setting, i.e., the best-performing tasks in their setup. 
See the supplementary for the results on Synthia.

Although we have shown experiments on dense tasks throughout our paper, note that our model is not restricted to just dense tasks. In the supplementary, we further report our model’s performance for the zero-shot image captioning task (IC) on the ’noCaps out-of-domain' benchmark.
\vspace{-5pt}
\begin{table}[ht]
\setlength\tabcolsep{3pt}
\centering
\scalebox{0.67}{
\begin{tabular}{lllllll}
%\toprule
%\multicolumn{6}{l}{~~~~~~~~~~~~~~~~~~~~~~~~~~~~~~~~~~~\textbf{Zero-shot task transfer on Cityscapes}~\cite{Cordts2016Cityscapes}}\\ \hline
&\multicolumn{1}{l}{}                                   & \multicolumn{2}{l}{~~~~~~~~\textbf{\textit{'S-D'}}}                                                                                                                             & \multicolumn{3}{l}{~~~~~~~~~~~~~~\textbf{\textit{'S-D-N'}}}                                                \\
\cmidrule(lr){1-2}\cmidrule(lr){3-4}\cmidrule(lr){5-7}
&\multicolumn{1}{c}{\multirow{-2}{*}{\textbf{Methods}}} & \cellcolor[HTML]{FFCCC9}\begin{tabular}[c]{@{}l@{}}SemSeg\\ mIoU\%$\uparrow$\end{tabular} & \cellcolor[HTML]{DAE8FC}\begin{tabular}[c]{@{}l@{}}Depth\\ RMSE$\downarrow$\end{tabular} & \cellcolor[HTML]{FFCCC9}\begin{tabular}[c]{@{}l@{}}SemSeg\\ mIoU\%$\uparrow$\end{tabular} & \cellcolor[HTML]{DAE8FC}\begin{tabular}[c]{@{}l@{}}Depth\\ RMSE$\downarrow$\end{tabular} & \cellcolor[HTML]{FFFFC7}\begin{tabular}[c]{@{}l@{}}Normal\\ mErr.$\downarrow$\end{tabular} \\
\cmidrule(lr){1-2}\cmidrule(lr){3-4}\cmidrule(lr){5-7}
%MTL-baseline~\cite{Vandenhende_2021}        &69.83&\cellcolor[HTML]{DAE8FC}6.707&72.27&4.949&\cellcolor[HTML]{FFFFC7}25.40
 %                            \\
%Consistency~\cite{zamir2020consistency}&   77.95&\cellcolor[HTML]{DAE8FC}5.666&78.37&4.209&\cellcolor[HTML]{FFFFC7}25.09
%                                 \\
%TAWT$+$MTL-baseline~\cite{tawt, Vandenhende_2021}    & 80.87&\cellcolor[HTML]{DAE8FC}5.400&83.03&4.088&\cellcolor[HTML]{FFFFC7}22.51
% \\
%MulT~\cite{MulT}   & 
%\underline{83.04} &\cellcolor[HTML]{DAE8FC}\underline{5.051} &\underline{85.93} & \underline{3.718} & \cellcolor[HTML]{FFFFC7}\underline{19.33} 
%                                  \\
%\rowcolor[HTML]{EFEFEF}
%\textbf{Our}   & 
%\textbf{85.13} &\cellcolor[HTML]{DAE8FC}\textbf{4.783} &\textbf{88.50} &\textbf{3.476} &\cellcolor[HTML]{FFFFC7}\textbf{16.08} 
% \\                       \hline   
% \multicolumn{6}{l}{~~~~~~~~~~~~~~~~~~~~~~~~~~~~~~~~~~~\textbf{Zero-shot task transfer on Cityscapes}~\cite{Cordts2016Cityscapes}}\\
%\cmidrule(lr){1-1}\cmidrule(lr){2-3}\cmidrule(lr){4-6}
\multirow{4}{*}{CNN}&TTNet~\cite{ttnet}  &  71.00&\cellcolor[HTML]{DAE8FC}8.101&71.40&6.511&\cellcolor[HTML]{FFFFC7}49.22\\
&Taskonomy~\cite{taskonomy2018}  &  60.02&\cellcolor[HTML]{DAE8FC}8.694&63.41&7.044&\cellcolor[HTML]{FFFFC7}53.57
                                    \\
&Consistency~\cite{zamir2020consistency}        &70.23&\cellcolor[HTML]{DAE8FC}7.773&71.67&6.575&\cellcolor[HTML]{FFFFC7}48.51
                             \\
&TAWT~\cite{tawt}    & 75.02&\cellcolor[HTML]{DAE8FC}\underline{7.596}&75.92&6.419&\cellcolor[HTML]{FFFFC7}45.28
                                       \\
                                       \cmidrule(lr){1-2}\cmidrule(lr){3-4}\cmidrule(lr){5-7}
\multirow{3}{*}{Transformer} &Vanilla MTL Swin~\cite{swin}   &  75.10 &\cellcolor[HTML]{DAE8FC}8.003& 75.97 & 8.000 &\cellcolor[HTML]{FFFFC7} 49.05
                                  \\                                      
&MulT~\cite{MulT}   & \underline{76.05} &\cellcolor[HTML]{DAE8FC}7.115 &\underline{77.50} & \underline{6.391} &\cellcolor[HTML]{FFFFC7} \underline{42.69} 
                                  \\
&\cellcolor[HTML]{EFEFEF}
\textbf{Our}   & \cellcolor[HTML]{EFEFEF}\textbf{78.00} &\cellcolor[HTML]{DAE8FC}\textbf{6.919} &\cellcolor[HTML]{EFEFEF}\textbf{80.55} &\cellcolor[HTML]{EFEFEF}\textbf{6.307} &\cellcolor[HTML]{FFFFC7}\textbf{41.10} 
 \\                      
\end{tabular}}
\setlength{\abovecaptionskip}{1mm}
\caption{\textbf{Results on zero-shot task transfer}. Our method outperforms all the MTL baselines. All the methods are first trained on the Vkitti2 benchmark and then fine-tuned to Cityscapes~\cite{Cordts2016Cityscapes}. %Note that in the \textit{'S-D'} setting, the models only have access to the segmentation ground-truth and in the \textit{'S-D-N'} setting, the models  have access to the segmentation and depth ground-truth. 
Zero-shot task predictions are highlighted in blue and yellow, respectively. Bold and underlined values show the best and second-best results. }\vspace{-10pt}
   \label{tb:zero-shot-task-transfer-results}%
%\vspace{-8pt}
%\end{table}
%-----------------------
%\begin{table}[ht]
\vspace{10pt}
\setlength\tabcolsep{3pt}
\centering
\scalebox{0.67}{
\begin{tabular}{lllll}
%\toprule
%\multicolumn{2}{l}{~~~~~~~~\textbf{Vkitti2~\cite{vkitti2}$\longrightarrow$Cityscapes}~\cite{Cordts2016Cityscapes}} 
%& \multicolumn{2}{l}{~~~~~~~~\textbf{\textit{'S-D'}}}                                                     \\\cmidrule(lr){1-2}\cmidrule(lr){3-4}
&\multicolumn{1}{c}{\multirow{-2}{*}{\textbf{Methods}}} &
\multicolumn{1}{c}{\multirow{-2}{*}{\textbf{MTL}}}& \cellcolor[HTML]{FFCCC9}\begin{tabular}[c]{@{}l@{}}SemSeg\\ mIoU\%$\uparrow$\end{tabular} & \cellcolor[HTML]{DAE8FC}\begin{tabular}[c]{@{}l@{}}Depth\\ RMSE$\downarrow$\end{tabular}  \\
\cmidrule(lr){1-2}\cmidrule(lr){3-3}\cmidrule(lr){4-5}
%\rowcolor[HTML]{C0C0C0}STL-source~\cite{unet}&  & 30.58 &15.84
%\\
%STL-UDA~\cite{unet}  &  & 37.12
%&14.39                        \\
%MTL-baseline-UDA~\cite{Vandenhende_2021}        &\checkmark &17.26 &14.85                       \\
%Consistency-UDA~\cite{zamir2020consistency}& \checkmark  &  34.19 &  12.84     
%                              \\
%XTAM-UDA~\cite{xtam}    & \checkmark & 37.93 &11.66
%                             \\
%
%MulT-UDA~\cite{MulT}   &\checkmark &\underline{42.12} &\underline{09.55}
 %                             \\
%\rowcolor[HTML]{EFEFEF}
%\textbf{Our-UDA}   & \checkmark &\textbf{50.03} &\textbf{06.99} 
% \\                       \hline   
%\multicolumn{4}{l}{~~~~~~~~\textbf{Vkitti2~\cite{vkitti2}$\longrightarrow$Cityscapes}~\cite{Cordts2016Cityscapes}}                                                     \\
%\cmidrule(lr){1-1}\cmidrule(lr){2-2}\cmidrule(lr){3-4}
%\rowcolor[HTML]{C0C0C0}1-task Swin-source~\cite{unet}&  & 58.72  &  12.04\\

\multirow{3}{*}{CNN} & MTL-baseline-UDA~\cite{Vandenhende_2021}        &\greencheck &57.26 &11.85                       \\
& Consistency-UDA~\cite{zamir2020consistency}& \greencheck  & 62.19 &  11.33               \\
%TAWT$+$MTL-baseline-UDA~\cite{tawt, Vandenhende_2021}    &\checkmark  & &
  %                      \\
&XTAM-UDA~\cite{xtam}    & \greencheck & 63.76 &11.15 
                                \\\cmidrule(lr){1-2}\cmidrule(lr){3-3}\cmidrule(lr){4-5}
\multirow{4}{*}{Transformer}& 1-task Swin-UDA~\cite{unet}  & \redcheck & 63.88   &   11.09                                                                             \\
 &MulT-UDA~\cite{MulT}   &\greencheck &\underline{66.12} &\underline{10.35} 
                              \\
&\cellcolor[HTML]{EFEFEF}\textbf{Our-UDA}   & \cellcolor[HTML]{EFEFEF}\greencheck &\cellcolor[HTML]{EFEFEF}\textbf{70.93} &\cellcolor[HTML]{EFEFEF}\textbf{08.66} \\ 
& \cellcolor[HTML]{C0C0C0}1-task Swin-target (Oracle)~\cite{swin}  & \cellcolor[HTML]{C0C0C0}\redcheck & \cellcolor[HTML]{C0C0C0}75.97 &\cellcolor[HTML]{C0C0C0}06.65  
\end{tabular}}
\setlength{\abovecaptionskip}{1mm}
\caption{\textbf{Unsupervised Domain Adaptation (UDA)} results for %Synthia~\cite{synthia}$\rightarrow$Cityscapes~\cite{Cordts2016Cityscapes} (top) and 
Vkitti2~\cite{vkitti2}$\rightarrow$Cityscapes~\cite{Cordts2016Cityscapes}. Our model outperforms all the baselines. Bold and underlined values show the best and
second-best results, respectively.}
   \label{tb:AVTaR-uda-results-syn2cityscapes}%
%\vspace{-10pt}
%\end{table}
%-------------------------------------------------------------------------
%\begin{table}[ht]
\vspace{5pt}
\setlength\tabcolsep{3pt}
\centering
\scalebox{0.65}{
\begin{tabular}{lllll}
%\toprule
%\multicolumn{2}{l}{~~~~~~~~\textbf{Taskonomy~\cite{taskonomy2018}$\longrightarrow$NYUDv2}~\cite{NYU}} 
%& \multicolumn{2}{l}{~~~~~~~~\textbf{\textit{'S-D'}}}                                                     \\\cmidrule(lr){1-2}\cmidrule(lr){3-4}
&\multicolumn{1}{c}{\multirow{-2}{*}{\textbf{Methods}}} &
\multicolumn{1}{c}{\multirow{-2}{*}{\textbf{MTL}}}& \cellcolor[HTML]{FFCCC9}\begin{tabular}[c]{@{}l@{}}SemSeg\\ mIoU\%$\uparrow$\end{tabular} & \cellcolor[HTML]{DAE8FC}\begin{tabular}[c]{@{}l@{}}Depth\\ RMSE$\downarrow$\end{tabular}  \\
\cmidrule(lr){1-2}\cmidrule(lr){3-3}\cmidrule(lr){4-5}
%\rowcolor[HTML]{C0C0C0}STL-source~\cite{unet}&  & 58.72  &  12.04\\
%STL-UDA~\cite{unet}  &  & 61.60   &   11.45                                                  \\
\multirow{2}{*}{CNN}& Consistency~\cite{zamir2020consistency}        &\greencheck & 26.24 &  0.771                   \\
&XTAM~\cite{xtam}& \greencheck  & 29.13 & 0.750               \\
\cmidrule(lr){1-2}\cmidrule(lr){3-3}\cmidrule(lr){4-5}
%TAWT$+$MTL-baseline-UDA~\cite{tawt, Vandenhende_2021}    &\checkmark  & &
  %                      \\
\multirow{4}{*}{Transformer}& 1-task Swin~\cite{swin}    & \redcheck & 32.09 & 0.722
                                \\
& ST-MTL~\cite{spatiotemporalMTL}   &\greencheck &{32.51} &{0.720} 
                              \\
& MulT~\cite{spatiotemporalMTL}   &\greencheck &\underline{33.68} &\underline{0.701} 
                              \\
& \cellcolor[HTML]{EFEFEF}
\textbf{Our}   &\cellcolor[HTML]{EFEFEF} \greencheck &\cellcolor[HTML]{EFEFEF}\textbf{40.77} &\cellcolor[HTML]{EFEFEF}\textbf{0.652}    
\end{tabular}}
\setlength{\abovecaptionskip}{1mm}
\caption{\textbf{Generalization to novel domains} results for %Synthia~\cite{synthia}$\rightarrow$Cityscapes~\cite{Cordts2016Cityscapes} (top) and 
Taskonomy~\cite{taskonomy2018}$\rightarrow$NYUDv2~\cite{NYU}. Our model outperforms all the baselines. Bold and underlined values show the best and
second-best results, respectively.}
   \label{tb:generalization-taskonomy-nyu}%
%\vspace{-10pt}
\end{table}
\vspace{-10pt}
\paragraph{Unsupervised Domain Adaptation:}  In this setting, the goal is to perform well on average on all tasks in the target domain, when the model is trained only on source domain labels but is \emph{aware} of the target domain images. We argue that task adaptation is beneficial for multitasking UDA as semantic and geometrical tasks exhibit complementary behaviors. We report results for the typical synthetic to real scenario, namely Vkitti2$\rightarrow$Cityscapes, in Table~\ref{tb:AVTaR-uda-results-syn2cityscapes} for the \textit{‘S-D’} multitask setup. We adopt a simple multitask Domain Adaptation (DA) solution based on output-level DA adversarial training~\cite{Saha_2021_CVPR} for all the models. We also report the 1-task Swin-target (Oracle), trained on the labeled target data. The use of our vision transformer adapters' task-adaptation mechanism significantly improves performance on all metrics. %The 1-task Swin-UDA performs better than CNN-based multitask learning in UDA~\cite{xtam, Vandenhende_2021, zamir2020consistency}. 
The selected baselines for UDA evaluation are those that have investigated UDA in their respective works~\cite{MulT, xtam,  Vandenhende_2021, zamir2020consistency}. Further details are provided in the supplementary material.
\vspace{-13pt}
\paragraph{Generalization to Novel Domains:}
 The TROA and TAA modules, in our vision transformer adapters, achieve generalization. In this section, we demonstrate how well our method generalizes to new domains without any fine-tuning. We compare our model with the two CNN-based MTL baselines of Consistency~\cite{zamir2020consistency}, XTAM~\cite{xtam}, as well as  the 1-task Swin baseline~\cite{swin}, and vision transformer-based MTL baselines such as ST-MTL~\cite{spatiotemporalMTL}, and MulT~\cite{MulT}, reported in Table~\ref{tb:generalization-taskonomy-nyu}. We use the models trained on Taskonomy dataset~\cite{taskonomy2018} and apply them to the NYUDv2~\cite{NYU} dataset without fine-tuning, as we find the
task affinities are similar across these domains. For example, TROA finds how similar segmentation and depth tasks (c.f. Figure~\ref{fig:weight-TROA}) are for Taskonomy comprising indoor scenes. This affinity when used together with TAA, ultimately, generalizes to NYU-v2 comprising indoor scenes. %We see that our method generalizes well to novel domains even without fine-tuning, unlike the baseline methods. 
%The XTAM model~\cite{xtam} shows better performance in comparison to the other CNN-based  MTL baselines because of its pairwise task attention, whereas the 1-task Swin model~\cite{swin} suffers due to the lack of joint task attention. ST-MTL~\cite{spatiotemporalMTL} lacks in learning task pairings, while MulT~\cite{MulT} shows an improved performance due to its shared attention mechanism across the evaluated tasks. In contrast, our vision transformer adapter model outperforms all the baselines and shows better generalization.
 An intuitive observation is that none of these models generalize to extremely disparate domains i.e. the networks trained on indoor scenes from Taskonomy cannot generalize to datasets with ‘faces’ or ‘animals’, simply because the networks have no notion of such categories of data. Nonetheless, we study the generalizability of our method to a disparate comics domain when the network is trained on MS-Coco~\cite{mscoco} which contains 'faces' or 'animals'. We provide these results in the supplementary.

\vspace{-10pt}
\begin{table}[ht]
\setlength\tabcolsep{3pt}
\centering
\scalebox{0.68}{
\arrayrulecolor{black}
\begin{tabular}{llll}

%\multicolumn{3}{l}{~ ~ ~ ~ ~ ~ ~ ~ ~ ~ ~ ~ ~ ~ ~ ~ ~\textbf{ ~ Parameter Comparison}}   \\
                                           &   Model      &\cellcolor[HTML]{C0C0C0}\begin{tabular}[c]{@{}l@{}}\# Parameters\\ (Millions)$\downarrow$\end{tabular} 
                                             & \cellcolor[HTML]{C0C0C0}\begin{tabular}[c]{@{}l@{}}Training time\\ (mins per epoch)$\downarrow$\end{tabular} \\ 
\cmidrule(lr){1-2}\cmidrule(lr){3-3}\cmidrule(lr){4-4}
\multirow{2}{*}{CNN}&XTAM~\cite{xtam}                                             & 304                                               & 16                                      \\
&Consistency~\cite{zamir2020consistency}                                              & \underline{228}                                               & \underline{14}                                       \\ \cmidrule(lr){1-2}\cmidrule(lr){3-3}\cmidrule(lr){4-4}
\multirow{3}{*}{Transformer}&Vanilla MTL Swin~\cite{swin}                                      & {348}                                               & {18}                                      \\ 
&MulT~\cite{MulT}                                                & 447                                               & 22                                       \\ 

&\cellcolor[HTML]{EFEFEF}\textbf{Our}                                              & \cellcolor[HTML]{EFEFEF}\cellcolor[HTML]{EFEFEF}\textbf{105.7}                                             &\cellcolor[HTML]{EFEFEF} \textbf{8}                                      \\ 
%\arrayrulecolor{black}\hline
\end{tabular}}%
\setlength{\abovecaptionskip}{0mm}
\caption{\textbf{Parameter and training time comparison} of our model on the Taskonomy~\cite{taskonomy2018} benchmark. Our method is more parameter efficient than all the MTL baselines. 
%Bold and underlined values show the best and second-best results, respectively.
}
\label{tb:parameter-comparison}%
%\vspace{-8pt}
\end{table}%
\vspace{-20pt}
\paragraph{Parameter Comparison}
In Table~\ref{tb:parameter-comparison}, we compare the time taken to train the models, in minutes per epoch. We show that our method is more parameter efficient than the CNN-based MTL baselines ~\cite{xtam, zamir2020consistency}, vanilla Swin model~\cite{swin}, and the transformer-based MTL approach~\cite{MulT} on \textit{'S-D-N-E'}, thanks to the vision adapters' bottleneck network that decreases the computational requirement by over an order of magnitude. We defer ablations for different modules of our network, different network sizes, freezing of encoder layers, and placement of the adapters in the supplementary.
%\paragraph{Supplementary Material.} We defer additional discussions and experiments, particularly analyzing the effect of the different transformer adapter backbones in our AVTaR model, the effect of the network size for different task combinations, as well as additional qualitative results to the supplementary material. We also discuss the environmental impact of training such models in the supplementary material.
%-------------------------------------------------------------------------
\vspace{-13pt}
\section{Conclusion and Limitations}
Our method demonstrates the benefit of task-adaptive learning for generalizable multitasking. Across the four settings, %of MTL, zero-shot task transfer, UDA, and generalization to novel domains,
our method outperforms not only CNN-based MTL methods but also vision transformer-based ones. %Precisely, for the multitask setting, our model outperforms the multitask models by at least 6.76, 0.0526, 2.97, 5.47 points and the single-task dedicated models by 13.4, 0.1183, 8.44, 33.51 points on Taskonomy's semantic segmentation, depth estimation, surface normal estimation and edge detection tasks, respectively.
%requiring ’N’ separately tuned single task models, or ’N’ times the number of parameters, where N is the number of tasks being solved. 
%Our method continues to outperform due to it’s learnt task dependencies. 
%This is shown by the reduction in the performance variances between the tasks as the task covariances become more aligned (please see supplementary).
This shows that our method  mitigates task interference and negative task transfer while promoting more efficient parameter sharing. %Our method can dynamically adapt the task affinities in different settings by using knowledge embedded in large pre-trained vision transformer models. 
%In summary, we present a multitasking model that can be applied to novel tasks \emph{and} novel domains without fully retraining or fine-tuning the model. 
Driven by the generalizability of our model, we hope that our method can help to solve dense task predictions on domains with limited data labels such as comics.
However, our framework has some limitations:\\
\textbf{Data Dependency.} Our model is data-intensive in the MTL setting. When trained on a limited amount of data, it may not achieve the same performance as reported in this work which is also the case for all the baselines. However, we generalize to other tasks and domains, unlike the baselines.\\
\textbf{Unpaired Data.}  Our current MTL model is trained in a supervised manner, thereby needing paired data. Extending our methodology to an unsupervised paradigm for MTL is feasible, as in~\cite{unsupervised-MTL}. 
Besides addressing these limitations, employing different pre-training modalities, such as text or video as in~\cite{chen2022vitadapter}, is also feasible.
\\

\textbf{{Acknowledgement.}} This work was supported in part by the Swiss National Science Foundation via the Sinergia grant CRSII5$-$180359.
%\textbf{Broader Impact.}
%Inter-task dependencies can incorrectly create an association and/or causation relationship among tasks that can be problematic for those datasets involving sensitive prediction quantities related to race, gender, religion etc. That said, we believe acknowledging these risks mitigates their potential for abuse, and the benefit from this work-most notably decreasing computational resources by over an order of magnitude compared with a state-of-the-art method in MTL while performing competitively— merits its dissemination.

%%%%%%%%% REFERENCES
%{\small
%\bibliographystyle{ieee_fullname}
%\bibliography{egbib}
%}

%\clearpage
%\newpage

%%%%%%%%% TITLE
\paragraph{\Large{Supplementary~ ~ ~ ~ ~ ~ ~ ~ ~ ~ ~ ~ ~ ~ ~ ~ ~ ~ ~ ~ }}~ ~ ~ ~ ~ ~ ~ ~ ~ ~ ~ ~ ~ ~ ~ ~ ~ ~ ~ ~ 
~ ~ ~ ~ ~ ~ ~ ~ ~ ~ ~ ~ ~ ~ ~ ~ ~ ~ ~ ~ 
The supplementary is organized as follows:
\begin{itemize}
    \item Section~\ref{sec:additional-quantitative}: Additional Quantitative Results 
    \item Section~\ref{sec:additional-exp-details}: Additional Experimental Details 
    \item Subsection~\ref{sec:datasets}: Datasets 
    \item Subsection~\ref{sec:baselines}: Baselines
    \item Subsection~\ref{sec:metrics}: Metrics
     \item Subsection~\ref{sec:training-details}: Training details
    %\item Section~\ref{sec:TROA-visualization}: Visualizing task affinity from TROA
    \item Section~\ref{sec:TAA-visualizing}: Visualizing task-adapted attention (TAA)
   % \item Section~\ref{sec:FiLM}: Visualizing Feature Wise Linear Modulation
    \item Section~\ref{sec:ablation}: Ablation study
      \item Section~\ref{sec:addtional-qualitative}: Additional qualitative results
\end{itemize}
\paragraph{Detailed Taxonomy}
We discuss a detailed taxonomy of multi-task learning approaches in Table~\ref{tb:Taxonomy-MTL-extended} extending those provided in Table~\ref{tb:Taxonomy-MTL} of the main paper. %(c.f. L281- 282 main paper). 
\begin{table*}[ht!]
\setlength\tabcolsep{1.7pt}
\centering
\scalebox{0.6}{
\begin{tabular}{llLLLLLLLl}
%\toprule 
&& \multicolumn{4}{l}{~~~~~~~~~~~~~~~~~~~~~~~~~~~~~~~~~~~~~\textbf{Architecture}} & \multicolumn{4}{l}{~~~~~~~~~~~~~~~~~~~~~\textbf{Task-affinity generalization}} \\ \cmidrule(lr){3-6}\cmidrule(lr){7-10}
&\multicolumn{1}{l}{~~\textbf{Methods}~~~~~} & \multicolumn{1}{l}{\textbf{Encoder-focused}} & \multicolumn{1}{l}{~~~~\textbf{Decoder-focused}} & \multicolumn{1}{l}{~~~\textbf{Attention}} & \multicolumn{1}{l}{~~~\textbf{Task-loss}} &\multicolumn{1}{l}{~~~~\textbf{MTL}~~~~}&\multicolumn{1}{l}{\textbf{Task-transfer}}& \multicolumn{1}{l}{~~~~\textbf{UDA}}& \multicolumn{1}{l}{\textbf{Novel domain}}\\ 
\cmidrule(lr){1-2}\cmidrule(lr){3-6}\cmidrule(lr){7-10}
\multirow{9}{*}{CNN-based} & MTL-baseline~\cite{Vandenhende_2021}                              & \multicolumn{1}{L}{\greencheck}                          &\multicolumn{1}{L}{\redcheck}                                &\multicolumn{1}{L}{\redcheck}                                         &\multicolumn{1}{L}{\redcheck}     &\multicolumn{1}{L}{\greencheck}&\multicolumn{1}{L}{\redcheck} & \multicolumn{1}{L}{\redcheck}&   \multicolumn{1}{L}{\redcheck}              \\
&Consistency~\cite{zamir2020consistency}                       &\multicolumn{1}{L}{\redcheck}                         & \multicolumn{1}{L}{\greencheck}                         &   \multicolumn{1}{L}{\redcheck}                                    & \multicolumn{1}{L}{ \greencheck }    &\multicolumn{1}{L}{\greencheck}&\multicolumn{1}{L}{\greencheck}&\multicolumn{1}{L}{\redcheck} &\multicolumn{1}{L}{\redcheck}                   \\
&XTAM~\cite{xtam}                               &\multicolumn{1}{L}{\redcheck}                         &\multicolumn{1}{L}{\redcheck}                         &  \multicolumn{1}{L}{\greencheck}                                      &\multicolumn{1}{L}{\redcheck}            &\multicolumn{1}{L}{\greencheck}&\multicolumn{1}{L}{\redcheck} & \multicolumn{1}{L}{\greencheck}    &\multicolumn{1}{L}{\redcheck}        \\
&TAWT~\cite{tawt}                               & \multicolumn{1}{L}{\greencheck}                         &\multicolumn{1}{L}{\redcheck}                         &\multicolumn{1}{L}{\redcheck}                                          &\multicolumn{1}{L}{\redcheck}            &\multicolumn{1}{L}{\greencheck}&\multicolumn{1}{L}{\greencheck}&\multicolumn{1}{L}{\redcheck}      &\multicolumn{1}{L}{\redcheck}       \\
%&MTL-Uncertainty~\cite{MTL-uncertainty}                           &\multicolumn{1}{L}{\redcheck}                         &\multicolumn{1}{L}{\redcheck}                         &    \multicolumn{1}{L}{\redcheck}                                    & \multicolumn{1}{L}{\greencheck}     &\multicolumn{1}{L}{\greencheck}&\multicolumn{1}{L}{\redcheck} & \multicolumn{1}{L}{\redcheck}  &\multicolumn{1}{L}{\redcheck}                 \\
&Cross-stitch~\cite{cross-stitch}                           & \multicolumn{1}{L}{\greencheck}                         &\multicolumn{1}{L}{\redcheck}                         &  \multicolumn{1}{L}{\redcheck}                                      &\multicolumn{1}{L}{\redcheck}        &\multicolumn{1}{L}{\greencheck}&\multicolumn{1}{L}{\redcheck} &\multicolumn{1}{L}{\redcheck}      &\multicolumn{1}{L}{\redcheck}           \\ 
&MTAN~\cite{endtoendMTL}                           & \multicolumn{1}{L}{\greencheck}                         &\multicolumn{1}{L}{\redcheck}                         &\multicolumn{1}{L}{\redcheck}                                          &\multicolumn{1}{L}{\redcheck}             &\multicolumn{1}{L}{\greencheck}&\multicolumn{1}{L}{\redcheck} & \multicolumn{1}{L}{\redcheck}     &\multicolumn{1}{L}{\redcheck}      \\
&MTI-Net~\cite{mti-net}                           &\multicolumn{1}{L}{\redcheck}                         & \multicolumn{1}{L}{\greencheck}                         &  \multicolumn{1}{L}{\greencheck}                                     &\multicolumn{1}{L}{\redcheck}         &\multicolumn{1}{L}{\greencheck}&\multicolumn{1}{L}{\redcheck} & \multicolumn{1}{L}{\redcheck}     &\multicolumn{1}{L}{\redcheck}          \\
&TSwitch~\cite{taskswitchnet}&\multicolumn{1}{L}{\redcheck}                         & \multicolumn{1}{L}{\greencheck}                         &\multicolumn{1}{L}{\redcheck}                                          & \multicolumn{1}{L}{\greencheck}      &\multicolumn{1}{L}{\greencheck}&\multicolumn{1}{L}{\redcheck} & \multicolumn{1}{L}{\redcheck} &\multicolumn{1}{L}{\redcheck}                 \\
&Grad-norm~\cite{gradnorm}                           &\multicolumn{1}{L}{\redcheck}                         &\multicolumn{1}{L}{\redcheck}                         &\multicolumn{1}{L}{\redcheck}                                         & \multicolumn{1}{L}{\greencheck}      &\multicolumn{1}{L}{\greencheck}&\multicolumn{1}{L}{\redcheck} & \multicolumn{1}{L}{\redcheck}     &\multicolumn{1}{L}{\redcheck}             \\
&PCGrad~\cite{pcgrad2020}                           &\multicolumn{1}{L}{\redcheck}                         &\multicolumn{1}{L}{\redcheck}                         & \multicolumn{1}{L}{\redcheck}                                        & \multicolumn{1}{L}{\greencheck}        &\multicolumn{1}{L}{\greencheck}&\multicolumn{1}{L}{\redcheck} &    \multicolumn{1}{L}{\redcheck}  &\multicolumn{1}{L}{\redcheck}           \\
&TTNet~\cite{ttnet}                           &\multicolumn{1}{L}{\redcheck}                         &\multicolumn{1}{L}{\redcheck}                         & \multicolumn{1}{L}{\redcheck}                                        & \multicolumn{1}{L}{\greencheck}        &\multicolumn{1}{L}{\greencheck}&\multicolumn{1}{L}{\greencheck}&    \multicolumn{1}{L}{\redcheck} &\multicolumn{1}{L}{\redcheck}            \\
&PAD-Net~\cite{padnet}                           & \multicolumn{1}{L}{\redcheck}                         & \multicolumn{1}{L}{\greencheck}                         &  \multicolumn{1}{L}{\greencheck}                                     & \multicolumn{1}{L}{\redcheck}          &\multicolumn{1}{L}{\greencheck}&\multicolumn{1}{L}{\redcheck}& \multicolumn{1}{L}{\redcheck}      &\multicolumn{1}{L}{\redcheck}        \\
&Taskonomy~\cite{taskonomy2018}                           & \multicolumn{1}{L}{\redcheck}                         & \multicolumn{1}{L}{\greencheck}                         &    \multicolumn{1}{L}{\redcheck}                                     & \multicolumn{1}{L}{\greencheck}      &\multicolumn{1}{L}{\greencheck}&\multicolumn{1}{L}{\greencheck}& \multicolumn{1}{L}{\redcheck}     &\multicolumn{1}{L}{\redcheck}              \\
&Taskgrouping~\cite{standley2019}                           & \multicolumn{1}{L}{\redcheck}                         & \multicolumn{1}{L}{\greencheck}                         &   \multicolumn{1}{L}{\redcheck}                                       & \multicolumn{1}{L}{\greencheck}      &\multicolumn{1}{L}{\greencheck}&\multicolumn{1}{L}{\redcheck}& \multicolumn{1}{L}{\redcheck}      &\multicolumn{1}{L}{\redcheck}             \\
&ATRC~\cite{atrc}                           & \multicolumn{1}{L}{\redcheck}                         & \multicolumn{1}{L}{\greencheck}                         &   \multicolumn{1}{L}{\redcheck}                                       & \multicolumn{1}{L}{\greencheck}      &\multicolumn{1}{L}{\greencheck}&\multicolumn{1}{L}{\greencheck}& \multicolumn{1}{L}{\redcheck}      &\multicolumn{1}{L}{\redcheck}             \\
\cmidrule(lr){1-2}\cmidrule(lr){3-6}\cmidrule(lr){7-10}
\multirow{8}{*}{Vision Transformer-based}&IPT~\cite{IPT}                               &\multicolumn{1}{L}{\redcheck}                         & \multicolumn{1}{L}{\greencheck}                         &  \multicolumn{1}{L}{\greencheck}                                       & \multicolumn{1}{L}{\greencheck}  &\multicolumn{1}{L}{\greencheck}&\multicolumn{1}{L}{\redcheck} &   \multicolumn{1}{L}{\redcheck} &\multicolumn{1}{L}{\redcheck}                    \\
&ST-MTL~\cite{spatiotemporalMTL}                               &\multicolumn{1}{L}{\redcheck}                         & \multicolumn{1}{L}{\greencheck}                         &  \multicolumn{1}{L}{\greencheck}                                       & \multicolumn{1}{L}{\greencheck}  &\multicolumn{1}{L}{\greencheck}&\multicolumn{1}{L}{\redcheck} &   \multicolumn{1}{L}{\redcheck}   &\multicolumn{1}{L}{\redcheck}                  \\
&Vid-MTL~\cite{video-multitask-transformer}                               &\multicolumn{1}{L}{\redcheck}                         & \multicolumn{1}{L}{\greencheck}                         &  \multicolumn{1}{L}{\greencheck}                                       & \multicolumn{1}{L}{\greencheck}  &\multicolumn{1}{L}{\greencheck}&\multicolumn{1}{L}{\redcheck} &   \multicolumn{1}{L}{\redcheck} &\multicolumn{1}{L}{\redcheck}                    \\
&UniT~\cite{hu2021unit}                               &\multicolumn{1}{L}{\redcheck}                         & \multicolumn{1}{L}{\greencheck}                         &  \multicolumn{1}{L}{\greencheck}                                       & \multicolumn{1}{L}{\greencheck}  &\multicolumn{1}{L}{\greencheck}&\multicolumn{1}{L}{\redcheck} &   \multicolumn{1}{L}{\redcheck}   &\multicolumn{1}{L}{\redcheck}                  \\
&InvPT~\cite{invpt2022}                               &\multicolumn{1}{L}{\redcheck}                         & \multicolumn{1}{L}{\greencheck}                         &  \multicolumn{1}{L}{\greencheck}                                       & \multicolumn{1}{L}{\greencheck}  &\multicolumn{1}{L}{\greencheck}&\multicolumn{1}{L}{\redcheck} &   \multicolumn{1}{L}{\redcheck}  &\multicolumn{1}{L}{\redcheck}                   \\
&Taskprompter~\cite{taskprompter2023}                               &\multicolumn{1}{L}{\greencheck}                         & \multicolumn{1}{L}{\greencheck}                         &  \multicolumn{1}{L}{\greencheck}                                       & \multicolumn{1}{L}{\redcheck}  &\multicolumn{1}{L}{\greencheck}&\multicolumn{1}{L}{\greencheck} &   \multicolumn{1}{L}{\redcheck}  &\multicolumn{1}{L}{\redcheck}                   \\
&MulT~\cite{MulT}                               &\multicolumn{1}{L}{\redcheck}                         & \multicolumn{1}{L}{\greencheck}                         &  \multicolumn{1}{L}{\greencheck}                                       & \multicolumn{1}{L}{\greencheck}  &\multicolumn{1}{L}{\greencheck}&\multicolumn{1}{L}{\redcheck} &   \multicolumn{1}{L}{\greencheck}  &\multicolumn{1}{L}{\redcheck}                   \\
%\cmidrule(lr){2-2}\cmidrule(lr){3-6}\cmidrule(lr){7-10}
&\textbf{~~Our}      & \multicolumn{1}{L}{\greencheck}  & \multicolumn{1}{L}{\redcheck} &  \multicolumn{1}{L}{\greencheck}                                       & \multicolumn{1}{L}{\greencheck}  &\multicolumn{1}{L}{\greencheck}&\multicolumn{1}{L}{\greencheck}&\multicolumn{1}{L}{\greencheck}&\multicolumn{1}{L}{\greencheck} \\
\cmidrule(lr){1-2}\cmidrule(lr){3-6}\cmidrule(lr){7-10}
\end{tabular}}
\setlength{\abovecaptionskip}{0mm}
\caption[Taxonomy of MTL approaches]{ %Our vision transformer adapter method is an encoder-focused, task-balanced approach that uses task-adapted attention (TAA) to learn generalizable task affinities, unlike existing CNN-based and vision transformer-based MTL methods. Here, we list the methods that we evaluate in this work. 
A detailed taxonomy of MTL methods (c.f. Table~\ref{tb:Taxonomy-MTL} main paper).%This corroborates the findings of ~\cite{Vandenhende_2021}. The task relations learned by our method generalize across MTL, task transfer, and the UDA settings, unlike the existing approaches.
}
   \label{tb:Taxonomy-MTL-extended}%
\vspace{-10pt}
\end{table*}
%\vspace{-10pt}
\section{Additional Quantitative Results}
\label{sec:additional-quantitative}
\subsection{Multitask Learning: Effect of Backbones } We report our MTL experiments on the NYUDv2~\cite{NYU} dataset in Table~\ref{tb:AVTaR-nyudv2-results}, where we also integrate the best-performing models with the same vision transformer backbones such as Swin~\cite{swin}, ViT~\cite{dosovitskiy2021an}, Pyramid Transformer (PVTv2-B5)~\cite{wang2021pvtv2}, and Focal Transformer (Focal-B)~\cite{yang2021focal} for a fair comparison. %(c.f. L597- 599 main paper).
Further, in Table~\ref{tb:AVTaR-synthia-vkitti2-results} we evaluate the methods on additional datasets such as Synthia~\cite{synthia} and Vkitti2~\cite{vkitti2}. %(c.f. L727- 729 main paper).
Our method outperforms all the baselines, showing the benefit of leveraging task-adapted attention instead of using attention from a single task, as done in the second-best performing model of MulT~\cite{MulT}. This corroborates the trend seen across Taskonomy~\cite{taskonomy2018} and Cityscapes~\cite{Cordts2016Cityscapes} in Table~\ref{tb:AVTaR-taxonomy-nyudv2-results} of the main paper. Furthermore, we show that the task performances consistently improve with the addition of more tasks, signifying the benefit of injecting additional geometrical cues to help the other tasks. Note that all the methods are initialized with pre-trained ImageNet-22K weights.

Our method with the Swin backbone~\cite{swin} shows the best performance. We, thus, choose to report the results with the Swin backbone in the main paper. Note that Swin is the most widely used model for dense prediction and its architecture compares with the hierarchical architecture of CNN-based baselines in Tables~\ref{tb:AVTaR-taxonomy-nyudv2-results}, ~\ref{tb:AVTaR-nyudv2-results}, and ~\ref{tb:AVTaR-synthia-vkitti2-results}.
\begin{table*} [ht!]
\centering
\setlength\tabcolsep{3pt}
\scalebox{0.68}{
\begin{tabular}{lllllllllll}
%\toprule
\multicolumn{11}{l}{~~~~~~~~~~~~~~~~~~~~~~~~~~~~~~~~~~~~~~~~~~~~~~~~~~~~~~~~~~~~~~~~~~~~~\textbf{Quantitative results on NYUDv2~\cite{NYU}}} \\
&\multicolumn{1}{l}{}                                   & \multicolumn{2}{l}{~~~~~~~~\textbf{$\textit{'S-D'}$}}                                                                                                                             & \multicolumn{3}{l}{~~~~~~~~~~~~~~\textbf{$\textit{'S-D-N'}$}}                                                                                                                                                                                                             &       \multicolumn{4}{l}{~~~~~~~~~~~~~~~~~~\textbf{$\textit{'S-D-N-E'}$}}                                         \\
\cmidrule(lr){1-2}\cmidrule(lr){3-4}\cmidrule(lr){5-7}\cmidrule(lr){8-11}
&\multicolumn{1}{c}{\multirow{-2}{*}{\textbf{Methods}}} & \cellcolor[HTML]{FFCCC9}\begin{tabular}[c]{@{}l@{}}SemSeg\\ mIoU\%$\uparrow$\end{tabular} & \cellcolor[HTML]{DAE8FC}\begin{tabular}[c]{@{}l@{}}Depth\\ RMSE$\downarrow$\end{tabular} & \cellcolor[HTML]{FFCCC9}\begin{tabular}[c]{@{}l@{}}SemSeg\\ mIoU\%$\uparrow$\end{tabular} & \cellcolor[HTML]{DAE8FC}\begin{tabular}[c]{@{}l@{}}Depth\\ RMSE$\downarrow$\end{tabular} & \cellcolor[HTML]{FFFFC7}\begin{tabular}[c]{@{}l@{}}Normal\\ mErr.$\downarrow$\end{tabular} & \cellcolor[HTML]{FFCCC9}\begin{tabular}[c]{@{}l@{}}SemSeg\\ mIoU\%$\uparrow$\end{tabular} & \cellcolor[HTML]{DAE8FC}\begin{tabular}[c]{@{}l@{}}Depth\\ RMSE$\downarrow$\end{tabular} & \cellcolor[HTML]{FFFFC7}\begin{tabular}[c]{@{}l@{}}Normal\\ mErr. $\downarrow$\end{tabular} & \cellcolor[HTML]{C8E685}\begin{tabular}[c]{@{}l@{}}Edges\\ F1\%$\uparrow$\end{tabular}\\
\cmidrule(lr){1-2}\cmidrule(lr){3-4}\cmidrule(lr){5-7}\cmidrule(lr){8-11}
%\multirow{10}{*}{CNN}&STL~\cite{unet}  &  38.70&0.635&38.70&0.635&36.90&38.70&0.635&36.90&54.90  
 %       \\
\multirow{10}{*}{ResNet-50 (CNN) backbone}&MTL-baseline~\cite{Vandenhende_2021}        &44.40&0.5870&44.61&0.5790&28.34&45.26&0.5407&25.79&76.07 

                                                                    \\
&Cross-stitch~\cite{cross-stitch} &44.20 &0.5900 & 44.40&0.5850 & 28.57&45.04 &0.5500 &26.11 & 76.00
                                                    \\
&MTAN~\cite{endtoendMTL} &45.00 & 0.5840 & 45.04 & 0.5490 &27.85 &45.50 & 0.5263 & 25.62 &76.18  \\
&TTNet~\cite{ttnet} & 45.12&0.5730  &45.16  &0.5400  &25.80 &46.00  &0.5217  &24.58  & 76.20\\
&Taskonomy~\cite{taskonomy2018} &44.33 & 0.5890&44.52 &0.5820 &28.47 & 45.10&0.5433 &25.94 &76.12 \\
&TSwitch~\cite{taskswitchnet} &47.04 &0.5581 &47.55 &0.5270 &24.02 &47.99 &0.5151 &25.40 &76.34 \\
&Consistency~\cite{zamir2020consistency} &45.38 &0.5627&46.14&0.5360&24.73&46.79&0.5204&25.58&76.30
                                                    \\
&ATRC~\cite{atrc} & 46.77&0.5436 &47.18 &0.5195 &23.30 &47.56 &0.5167 &22.11 &76.58 \\
& XTAM~\cite{xtam}   & 46.90&0.5372&47.24&0.5177&22.73&48.80&0.5150&22.39&76.88

                                                                           \\
 &TAWT~\cite{tawt}   & 47.02&0.5330&47.29&0.5152&22.69&48.87&0.5146&22.30&76.90 
                                        \\
                                        \cmidrule(lr){1-2}\cmidrule(lr){3-4}\cmidrule(lr){5-7}\cmidrule(lr){8-11}
\multirow{1}{*}{HRNet-48 (CNN) backbone}&MTI-Net~\cite{mti-net} &49.00 & 0.5290& 49.52& 0.5050& 20.24 &49.88 & 0.4940&20.13 & 76.95 \\
\cmidrule(lr){1-2}\cmidrule(lr){3-4}\cmidrule(lr){5-7}\cmidrule(lr){8-11}
\multirow{9}{*}{Swin-B V2~\cite{swin} transformer backbone}&ATRC~\cite{atrc} &49.11&0.5273 &49.55 &0.5034 &20.36 &50.40 & 0.4978&20.06 &77.11 \\
& XTAM~\cite{xtam}   &49.25 &0.5251 &50.13 &0.5008 &20.19 &50.55 &0.4977 &19.50  &77.30

                                                                           \\
 &TAWT~\cite{tawt}   & 49.33&0.5242 &50.20 &0.5001 &20.10 &50.71 &0.4955 &19.28 &77.35 
                                        \\
                                       
&MTI-Net~\cite{mti-net} & 49.33& 0.5180 &49.81 &0.4990 & 20.15&50.38 &0.4933 & 19.08& 77.95 \\
&ST-MTL~\cite{spatiotemporalMTL}   &49.84  &0.5178
 &52.68 &0.4975 & 19.82 & 53.72&0.4924 & 18.19&78.10
                                                                     \\
&InvPT~\cite{invpt2022}   & 51.59 &0.5166
 &52.94 & 0.4960&19.15  &54.37 &0.4906 & 18.08&78.61
                                                                     \\
&Taskprompter~\cite{taskprompter2023}   & 53.27 &0.5150
 & 54.04&0.4951 & 18.88 & 55.34&0.4888 &18.00 &\underline{78.71}
                                                                     \\
&MulT~\cite{MulT}   & \underline{53.48} &\underline{0.5130}
 &\underline{54.17} & \underline{0.4937} & \underline{18.72} &\underline{55.89} &\underline{0.4885} &\underline{17.97} &{78.65}
                                                                     \\

&\cellcolor[HTML]{EFEFEF}\textbf{Our}   & \cellcolor[HTML]{EFEFEF}\textbf{53.61} &\cellcolor[HTML]{EFEFEF}\textbf{0.5111} &\cellcolor[HTML]{EFEFEF}\textbf{54.80} &\cellcolor[HTML]{EFEFEF}\textbf{0.4922} &\cellcolor[HTML]{EFEFEF}\textbf{18.63} &\cellcolor[HTML]{EFEFEF}\textbf{56.13} &\cellcolor[HTML]{EFEFEF}\textbf{0.4861} &\cellcolor[HTML]{EFEFEF}\textbf{17.50} &\cellcolor[HTML]{EFEFEF}\textbf{80.03}
 \\ \cmidrule(lr){1-2}\cmidrule(lr){3-4}\cmidrule(lr){5-7}\cmidrule(lr){8-11}
\multirow{9}{*}{ViT-B~\cite{dosovitskiy2021an} transformer backbone}&ATRC~\cite{atrc} &48.49 & 0.5285& 49.38 & 0.5050&20.52 &50.25 & 0.4991& 20.20& 76.91\\
& XTAM~\cite{xtam}   &49.11 & 0.5265&49.94 & 0.5024&20.29 &50.39 &0.4987 & 19.61 &77.17

                                                                           \\
 &TAWT~\cite{tawt}   &49.15 & 0.5255& 50.00& 0.5022& 20.19& 50.54& 0.4968& 19.40& 77.22
                                        \\
                                       
&MTI-Net~\cite{mti-net} &49.24 &0.5192 & 49.70& 0.4997& 20.21&50.25 &0.4938 & 19.21& 77.80 \\
&ST-MTL~\cite{spatiotemporalMTL}   & 49.72 &0.5187
 &52.54 &0.4988 &19.96  &53.57 &0.4936 &18.25 &77.97
                                                                     \\
&InvPT~\cite{invpt2022}   &51.49  &0.5177
 &52.83 &0.4974 &19.33  &54.23 &0.4920 &18.21 &78.44
                                                                     \\
&Taskprompter~\cite{taskprompter2023}   &53.22  &0.5164
 &53.92 & 0.4966& 19.00 & 55.30& 0.4910&18.19 &\underline{78.56}
                                                                     \\
&MulT~\cite{MulT}   & \underline{53.33} &\underline{0.5148}
 &\underline{54.00} & \underline{0.4955} & \underline{18.91} &\underline{55.81} &\underline{0.4901} &\underline{18.13} &{78.50}
                                                                     \\

&\cellcolor[HTML]{EFEFEF}\textbf{Our}   & \cellcolor[HTML]{EFEFEF}\textbf{53.47} &\cellcolor[HTML]{EFEFEF}\textbf{0.5123 } &\cellcolor[HTML]{EFEFEF}\textbf{54.68 } &\cellcolor[HTML]{EFEFEF}\textbf{0.4937 } &\cellcolor[HTML]{EFEFEF}\textbf{18.75 } &\cellcolor[HTML]{EFEFEF}\textbf{56.00} &\cellcolor[HTML]{EFEFEF}\textbf{0.4872 } &\cellcolor[HTML]{EFEFEF}\textbf{17.57} &\cellcolor[HTML]{EFEFEF}\textbf{79.83 }
 \\ 
 \cmidrule(lr){1-2}\cmidrule(lr){3-4}\cmidrule(lr){5-7}\cmidrule(lr){8-11}
\multirow{9}{*}{PVTv2-B5~\cite{wang2021pvtv2} transfomer backbone}&ATRC~\cite{atrc} & 49.00&0.5280 & 49.47&0.5043&20.46 & 50.33&0.4986 &20.14 & 77.00\\
& XTAM~\cite{xtam}   &49.17 &0.5258 &50.03 &0.5019 &20.25 &50.45 &0.4983 &19.58  &77.20

                                                                           \\
 &TAWT~\cite{tawt}   &49.22 & 0.5250&50.11 &0.5017 &20.13 & 50.60& 0.4963 & 19.37 & 77.27 
                                        \\
                                       
&MTI-Net~\cite{mti-net} & 49.26& 0.5189&49.72 &0.4995 & 20.20 & 50.29 & 0.4938 & 19.18 & 77.83  \\
&ST-MTL~\cite{spatiotemporalMTL}   & 49.77 & 0.5185
 & 52.60& 0.4983& 19.91 & 53.62& 0.4934 & 18.22 & 78.02
                                                                     \\
&InvPT~\cite{invpt2022}   & 51.53 & 0.5172
 & 52.88 & 0.4969 & 19.27 & 54.29& 0.4915&18.18 &78.50
                                                                     \\
&Taskprompter~\cite{taskprompter2023}   & 53.25 & 0.5161
 & 54.00& 0.4960 & 18.96 & 55.33 & 0.4904& 18.14&\underline{78.60}
                                                                     \\
&MulT~\cite{MulT}   & \underline{53.37} &\underline{0.5144}
 &\underline{54.04} & \underline{0.4950} & \underline{18.87} &\underline{55.85} &\underline{0.4898} &\underline{18.11} &{78.55}
                                                                     \\

&\cellcolor[HTML]{EFEFEF}\textbf{Our}   & \cellcolor[HTML]{EFEFEF}\textbf{53.53 } &\cellcolor[HTML]{EFEFEF}\textbf{0.5117} &\cellcolor[HTML]{EFEFEF}\textbf{54.72 } &\cellcolor[HTML]{EFEFEF}\textbf{0.4934} &\cellcolor[HTML]{EFEFEF}\textbf{18.71 } &\cellcolor[HTML]{EFEFEF}\textbf{56.05} &\cellcolor[HTML]{EFEFEF}\textbf{0.4869 } &\cellcolor[HTML]{EFEFEF}\textbf{17.55} &\cellcolor[HTML]{EFEFEF}\textbf{79.88 }
 \\ 
 \cmidrule(lr){1-2}\cmidrule(lr){3-4}\cmidrule(lr){5-7}\cmidrule(lr){8-11}
\multirow{9}{*}{Focal-B~\cite{yang2021focal} transformer backbone}&ATRC~\cite{atrc} &49.09 &0.5277 &49.50 & 0.5038&20.42 & 50.39&0.4982 & 20.10& 77.07\\
& XTAM~\cite{xtam}   &49.20 & 0.5253&50.07 & 0.5010&20.22 & 50.51& 0.4981& 19.53 &77.26

                                                                           \\
 &TAWT~\cite{tawt}   &49.28 &0.5247 &50.15 &0.5011 &20.11 & 50.66&0.4960 &19.32 & 77.31
                                        \\
                                       
&MTI-Net~\cite{mti-net} & 49.29& 0.5186&49.77 & 0.4992&20.19 & 50.33&0.4938 &19.15 & 77.88 \\
&ST-MTL~\cite{spatiotemporalMTL}   &49.80  &0.5181
 &52.64 &0.4980 & 19.87 &53.66 &0.4930 &18.26 &78.06
                                                                     \\
&InvPT~\cite{invpt2022}   &51.56  &0.5168
 &52.90 &0.4965 &19.22  &54.33 &0.4913 &18.15 &78.56
                                                                     \\
&Taskprompter~\cite{taskprompter2023}   & 53.26  &0.5158
 &54.02 &0.4955 & 18.93 & 55.36&0.4892 &18.07 &\underline{78.73}
                                                                     \\
&MulT~\cite{MulT}   & \underline{53.40} &\underline{0.5137}
 &\underline{54.11} & \underline{0.4944} & \underline{18.80} &\underline{55.90} &\underline{0.4890} &\underline{18.03} &{78.68}
                                                                     \\

&\cellcolor[HTML]{EFEFEF}\textbf{Our}   & \cellcolor[HTML]{EFEFEF}\textbf{53.57} &\cellcolor[HTML]{EFEFEF}\textbf{0.5115} &\cellcolor[HTML]{EFEFEF}\textbf{54.75 } &\cellcolor[HTML]{EFEFEF}\textbf{0.4933 } &\cellcolor[HTML]{EFEFEF}\textbf{18.68 } &\cellcolor[HTML]{EFEFEF}\textbf{56.07 } &\cellcolor[HTML]{EFEFEF}\textbf{0.4867} &\cellcolor[HTML]{EFEFEF}\textbf{17.52} &\cellcolor[HTML]{EFEFEF}\textbf{79.91 }
 \\ 
\end{tabular}}
\setlength{\abovecaptionskip}{1mm}
\caption[Multitask learning results on NYUDv2]{\textbf{Multitask learning results} on the  NYUDv2~\cite{NYU} benchmark for different multitask settings of $\textit{'S-D'}$, $\textit{'S-D-N'}$, and $\textit{'S-D-N-E'}$. Our model consistently outperforms both the CNN-based and vision transformer-based baselines. Adding more tasks improves their respective performances based on their task affinities. Bold and underlined values show the best and second-best results, respectively. }
\label{tb:AVTaR-nyudv2-results}%
\end{table*}
%%%%%%%%%%%%%%%%%%%
\vspace{-6pt}
\subsection{Zero-Shot Task Transfer}
Although we have shown experiments on dense tasks throughout our paper, note that our model is not restricted to just dense tasks. In Table~\ref{tb:text-zero-shot}, we report our model's  performance for the zero-shot image captioning task (IC) on 'noCaps out-of-domain' benchmark. Following the typical zero-shot task transfer setting, our model is trained with segmentation and captions from Coco and applied to noCaps for zero-shot IC. For training, we follow the GIT~\cite{wang2022git} text decoder configuration.
%Our model uses only ImageNet-22K pretraining, unlike GIT.
During training on Coco, we enforce the highest similarity between the TAA token and the text decoder output token. On noCaps, we achieve comparable IC performance to GIT using a quarter \#params. 
\begin{table}[ht!]
\setlength\tabcolsep{3pt}
\centering
\scalebox{0.79}{
\arrayrulecolor{black}
\begin{tabular}{!{\color{white}\vrule}l!{\color{white}\vrule}c!{\color{white}\vrule}c!{\color{white}\vrule}c!{\color{white}\vrule}c!{\color{white}\vrule}c}
\hline
{Method}&Pretraining& \#Params & METEOR$\uparrow$& CIDEr$\uparrow$& SPICE$\uparrow$ \\
\cmidrule(lr){1-1}\cmidrule(lr){2-2}\cmidrule(lr){3-3}\cmidrule(lr){4-6}
GIT                                              &   800M image \& text                                        &  700M                                     & \textbf{30.45}                                           & \textbf{122.04}                                             &  \textbf{15.70}                                                                              \\ 
\cellcolor[HTML]{EFEFEF}\textbf{Ours}                                               & \cellcolor[HTML]{EFEFEF}ImageNet-22K only                                         & \cellcolor[HTML]{EFEFEF}\textbf{163M}                                                &\cellcolor[HTML]{EFEFEF} 29.82                                          & \cellcolor[HTML]{EFEFEF}119.93                                           & \cellcolor[HTML]{EFEFEF}13.13                                            \\ 
%\arrayrulecolor{black}\hline
\end{tabular}}%
\setlength{\abovecaptionskip}{1mm}
\caption[Zero-Shot Task Transfer results for Image Captioning]{\textbf{Zero-Shot Task Transfer results for Image Captioning} on the  noCaps out-of-domain benchmark. Our model is comparable in performance to GIT~\cite{wang2022git} while using a quarter number of parameters. }
   \label{tb:text-zero-shot}%
\end{table}%
%\vspace{-6pt}
\subsection{Unsupervised Domain Adaptation}
%\paragraph{Architecture of our UDA method}
\begin{figure*}[ht!]
\centering
{\includegraphics[ width=0.85\linewidth ]{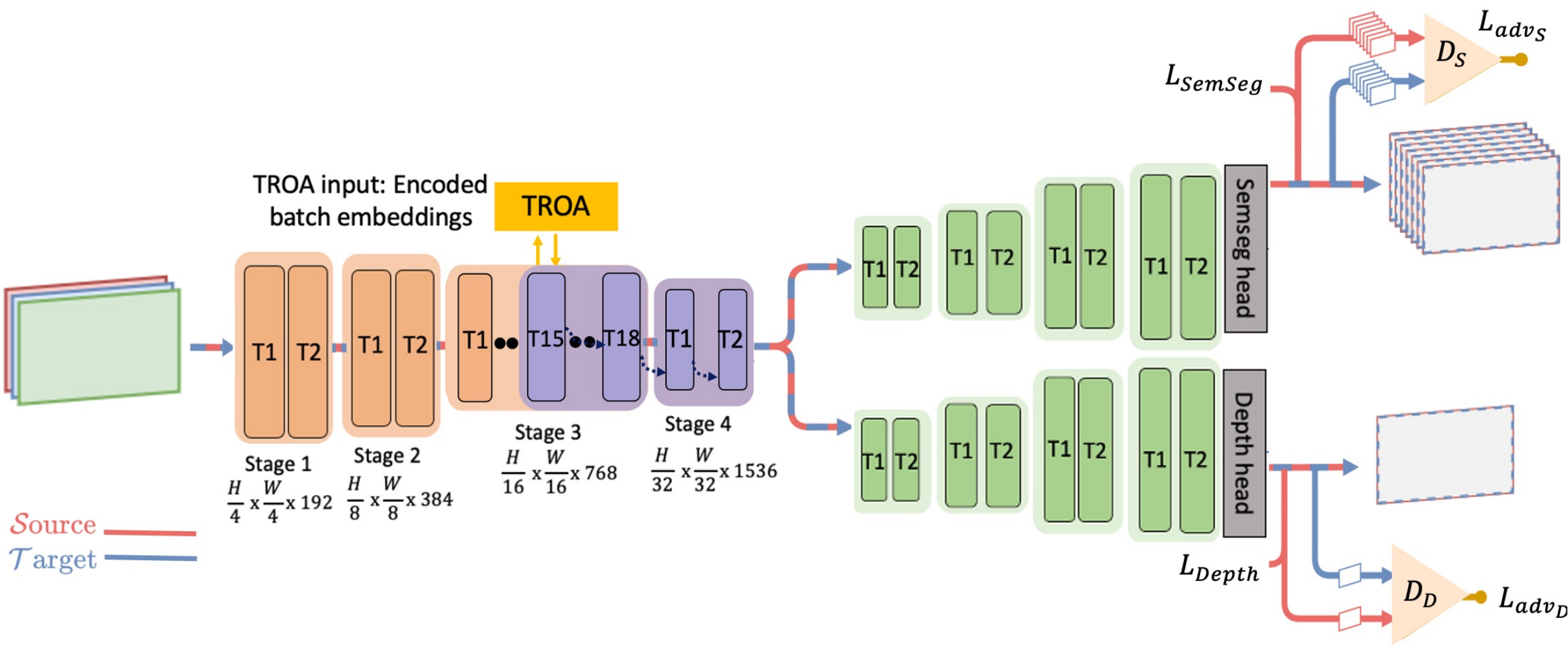}}
\setlength{\abovecaptionskip}{0pt}
\caption{\textbf{UDA architecture for our method} with output-level adversarial learning. Arrows indicating data flows are drawn in either red (source), blue (target), or a mix (both). Domain Discriminators (shown as yellow triangles) are jointly trained with our multitask model. }\label{fig:uda-architecture}\vspace{-10pt}
\end{figure*}
In Figure~\ref{fig:uda-architecture}, we illustrate the architecture employed for our UDA setting that makes use of the adversarial learning scheme in~\cite{Saha_2021_CVPR}. We align the source and the target domains by applying a task-head discriminator. The alignment is done at the final output-levels for both segmentation and depth in order to preserve the architecture of our original model. In Table~\ref{tb:uda-syn2cityscapes}, we report additional UDA results %(c.f. L798- 799 main paper)
with the source domain as Synthia~\cite{synthia} and the target domain as Cityscapes~\cite{Cordts2016Cityscapes}. Following the same trend as seen in Table~\ref{tb:AVTaR-uda-results-syn2cityscapes} of the main paper, we outperform both the ResNet-50 (CNN) baselines as well as the Swin-B V2 transformer baselines. 

\subsection{Generalization}
%The TROA and TAA in our vision transformer adapters achieve generalization. For example, TROA finds how similar segmentation and depth tasks are for MS-Coco~\cite{mscoco} comprising images with ’faces’ or ’animals’ and this affinity is leveraged when generalizing to the comics domain comprising ’faces’ or ’animals, as we find the task affinities are similar across these domains. To use the task affinities for each image query token, we combine the FilM~\cite{film} transformed affinity matrix with attention, getting TAA. 
We study the generalizability of our method to the comics domain~\cite{dcm} when the network is trained on MS-Coco~\cite{mscoco} dataset and is no fine-tuned to the DCM comics dataset. Shown in Table~\ref{tb:uda-dcm}, our model outperforms both the ResNet-50 (CNN) baselines as well as the Swin-B V2 transformer baselines. 
%(c.f. L818- 819 main paper).
%For segmentation alignment, we use “weighted self-information” map computed from the softmax segmentation output as in~\cite{vu2019advent}. For depth alignment, we normalize the depth-map outputs using the source’s min and max depth values, and directly align the continuous normalized maps as in~\cite{wang2022semi}. We follow the same strategy for all the baseline models by appending the domain discriminators to the baseline output-levels.
%%%%%%

\section{Additional Experimental Details }
\label{sec:additional-exp-details}
\subsection{Datasets}
\label{sec:datasets}
We evaluate our method on the following datasets.\\
\textbf{Taskonomy~\cite{taskonomy2018}} comprises 4 million real images of indoor scenes with multi-task annotations for each image. The experiments were performed using the following 4 tasks from the dataset: semantic segmentation, depth (zbuffer), surface normals, and 2D (Sobel) texture edges. The tasks were selected to cover both geometric and semantic-based cues and have sensor-based/semantic ground truth. We report results on the official test set.\\ 
\textbf{NYUDv2~\cite{NYU}} consists of sequences of RGB images, depth recorded by a Kinect camera, and dense labeling for semantic segmentation covering 894 classes. Officially, 249 scenes are reserved for training (for a total amount of 240k frames) and 215 scenes are reserved for testing. We use the official train-test split for our evaluations.\\
\textbf{Cityscapes~\cite{Cordts2016Cityscapes}} consists of 5000 images with semantic annotations for 30 classes, grouped into 8 categories. For depth, we use depth from semi-global matching as depth labels. We estimate surface normal labels from the depth maps following~\cite{pseudo-surface-normal}.\\
\textbf{Vkitti2~\cite{vkitti2}} contains 50 high-resolution monocular videos (21,260 frames) generated from five different virtual worlds. These photo-realistic synthetic videos are densely, exactly, and fully annotated with semantic segmentation (14 classes) and depth labels. \\
\textbf{Synthia~\cite{synthia}} contains synthetic images of 9400 multi-viewpoint photo-realistic frames rendered from a virtual city. We use pixel-level semantic annotations for 16 classes and depth labels from Synthia's RAND-Cityscapes-CVPR-2016 benchmark as used in~\cite{xtam}. \\
\textbf{MS-Coco ~\cite{mscoco}} comprises 164k training images that span over 80 categories with semantic segmentation annotations. For depth, normal, and edge we use pseudo-labels from ~\cite{midas, pseudo-surface-normal, rindnet}, respectively. We use the MS-Coco dataset to evaluate if the models generalize to novel domains like comics which comprise data categories like 'faces' and 'animals'. \\
%\vspace{5pt}
\textbf{DCM~\cite{dcm}} is a comics dataset comprising 772 full-page images with multiple comics panel images within. We use these images as test images from a novel domain for our additional experiments in the \textit{generalization to novel domains} setting. 
\subsection{Baselines}
\label{sec:baselines}
The baselines for our evaluation are described below. To prevent confounding factors, all baselines in the main paper (Tables 2-6) were implemented using the training procedure and the best model configurations as outlined in their respective works. Additionally, as shown in Table~\ref{tb:AVTaR-nyudv2-results}, we report the best-performing CNN-based baselines on the same transformer backbone architectures.  
\vspace{-15pt}
\subsubsection{{CNN-based Methods}}
\textbf{MTL-baseline~\cite{Vandenhende_2021}}: This is a naive multi-task learning network with one shared encoder and multiple task-specific decoders based on a ResNet-50 backbone. \\
\textbf{Cross-stitch~\cite{cross-stitch}}: introduced soft-parameter sharing in deep MTL architectures. %The model uses a linear combination of the activations in every layer of the task-specific networks as a means for soft feature fusion. 
Being a ResNet-50 encoder-focused method that can achieve task transfer learning, we use this baseline for comparison.\\
\textbf{MTAN~\cite{endtoendMTL}}: used an attention mechanism
to share a general feature pool amongst the task-specific networks. Being an encoder-focused method, we use this baseline for comparison.
%A concern with soft parameter sharing approaches
%is scalability, as the size of the multi-task network tends to
%grow linearly with the number of tasks 
\\
\textbf{TSwitch~\cite{taskswitchnet}}: We use this as a baseline for comparison, as it uses a task embedding network to learn task-specific conditioning parameters that encourages constructive interaction between tasks in a pairwise manner. \\
\textbf{TTNet~\cite{ttnet}}: presents a meta-learning algorithm that regresses model parameters for novel tasks for which no ground truth is available (zero-shot tasks). \\
\textbf{Taskonomy~\cite{taskonomy2018}}: We use this as a baseline as Taskonomy studies the relationships between multiple visual tasks for task transfer learning. \\
\textbf{MTI-Net~\cite{mti-net}}: is a multiscale distillation procedure to explicitly model the unique task interactions that happen at each individual scale. We use MTI-Net with HRNet-48 as a baseline. Additionally, we compare MTI-Net with the the different transformer backbones. \\
\textbf{Consistency~\cite{zamir2020consistency}}: This work presents a data-driven framework for augmenting standard multi-task learning with a cross-task consistency constraint, which is learned over a graph of arbitrary tasks. \\
\textbf{TAWT~\cite{tawt}}: This method uses gradient-loss to find optimal task representations to perform multi-task learning. TAWT shows that learning task representations in the encoder benefits multi-task learning. Being an encoder-focused method that can achieve task transfer learning, we use this baseline for comparison.\\
\textbf{XTAM~\cite{xtam}}: exploits correlation-guided attention between task pairs to enhance the average representation learning for all tasks. We use this baseline for comparison as it investigates the problem of MTL and UDA.\\
\textbf{Adaptive Task-Relational Context (ATRC)~\cite{atrc}}: leverages pairwise task similarities to create attention gates for global cross-task message passing.
\balance%It samples each task pair using neural architecture search and outputs the optimal configuration. 
%Being a task transfer learning network, we use this baseline for comparison. 
%\vspace{-10pt}
\begin{table}[h]
\setlength\tabcolsep{3pt}
\centering
\scalebox{0.63}{
\begin{tabular}{lllllll}
%\toprule
\multicolumn{7}{l}{~~~~~~~~~~~~~~~~~~~~~~~~~~~~~~~~~~~\textbf{Quantitative results on Synthia}~\cite{synthia}}\\
&\multicolumn{1}{l}{}                                   & \multicolumn{2}{l}{~~~~~~~~\textbf{$\textit{'S-D'}$}}                                                                                                                           & \multicolumn{3}{l}{~~~~~~~~~~~~~~\textbf{$\textit{'S-D-N'}$}}                                                \\
\cmidrule(lr){1-2}\cmidrule(lr){3-4}\cmidrule(lr){5-7}
&\multicolumn{1}{c}{\multirow{-2}{*}{\textbf{Methods}}} & \cellcolor[HTML]{FFCCC9}\begin{tabular}[c]{@{}l@{}}SemSeg\\ mIoU\%$\uparrow$\end{tabular} & \cellcolor[HTML]{DAE8FC}\begin{tabular}[c]{@{}l@{}}Depth\\ RMSE$\downarrow$\end{tabular}  &\cellcolor[HTML]{FFCCC9}\begin{tabular}[c]{@{}l@{}}SemSeg\\ mIoU\%$\uparrow$\end{tabular} & \cellcolor[HTML]{DAE8FC}\begin{tabular}[c]{@{}l@{}}Depth\\ RMSE$\downarrow$\end{tabular} & \cellcolor[HTML]{FFFFC7}\begin{tabular}[c]{@{}l@{}}Normal\\ mErr.$\downarrow$\end{tabular} \\
\cmidrule(lr){1-2}\cmidrule(lr){3-4}\cmidrule(lr){5-7}
%STL~\cite{unet}  &  67.43 & 5.379 &67.43 &5.379 &19.61
 %                                    \\
\multirow{9}{*}{ResNet-50 backbone}&MTL-baseline~\cite{Vandenhende_2021}        &69.83&5.166&72.27&4.949&19.28
                                                                    \\
 &Cross-stitch~\cite{cross-stitch} &69.00 & 5.228& 71.80&5.085 &21.05 \\  
&MTAN~\cite{endtoendMTL} &77.42 & 4.285& 77.90& 4.298 & 17.48\\  
&TTNet~\cite{ttnet} &77.51 & 4.270& 78.00&4.266 &17.54 \\  
&Taskonomy~\cite{taskonomy2018} & 69.40& 5.209& 72.16& 4.974& 20.09\\  
&TSwitch~\cite{taskswitchnet} & 78.01&4.255 &78.42 &4.200 &17.05 \\                                                                      
&Consistency~\cite{zamir2020consistency}&   77.95&4.263&78.37&4.209&17.28
                                 \\
&XTAM~\cite{xtam}     & 80.53&4.222&82.99&4.088&14.46
                                                                           \\
&TAWT~\cite{tawt}   & 80.87 & 4.161 &83.03 &4.056 &14.30
                                        \\\cmidrule(lr){1-2}\cmidrule(lr){3-4}\cmidrule(lr){5-7}
\multirow{5}{*}{Swin-B V2 backbone}&XTAM~\cite{xtam}     &81.70 & 4.199 & 83.40 & 4.040 & 14.00
                                                                           \\
&TAWT~\cite{tawt}   & 81.91 & 4.118 & 83.75& 4.000& 13.66
                                        \\
&ST-MTL~\cite{spatiotemporalMTL} & 82.48 & 4.001 & 85.02& 3.808&  13.49\\  

&MulT~\cite{MulT}   & 
\underline{83.04} &\underline{3.883} &\underline{86.90} & \underline{3.662} & \underline{13.27} 
                                                                     \\
&\cellcolor[HTML]{EFEFEF}
\textbf{Our}   & 
\cellcolor[HTML]{EFEFEF}\textbf{85.13} &\cellcolor[HTML]{EFEFEF}\textbf{3.695} &\cellcolor[HTML]{EFEFEF}\textbf{88.50} &\cellcolor[HTML]{EFEFEF}\textbf{3.476} &\cellcolor[HTML]{EFEFEF}\textbf{13.10} 
 \\                      \cmidrule(lr){1-2}\cmidrule(lr){3-4}\cmidrule(lr){5-7}
\multirow{5}{*}{ViT-B backbone}&XTAM~\cite{xtam}     & 81.50& 4.207 & 83.32  & 4.049 &14.08
                                                                           \\
&TAWT~\cite{tawt}   & 81.82 & 4.133 & 83.66 &4.012 & 13.80
                                        \\
&ST-MTL~\cite{spatiotemporalMTL} & 82.37 & 4.009&84.00 & 3.851& 13.61\\  

&MulT~\cite{MulT}   & \underline{82.90}
 & \underline{3.892} & \underline{86.82}& \underline{3.689} & \underline{13.35}
                                                                     \\
&\cellcolor[HTML]{EFEFEF}
\textbf{Our}   & 
\cellcolor[HTML]{EFEFEF}\textbf{85.00} &\cellcolor[HTML]{EFEFEF}\textbf{3.707} &\cellcolor[HTML]{EFEFEF}\textbf{88.35} &\cellcolor[HTML]{EFEFEF}\textbf{3.487} &\cellcolor[HTML]{EFEFEF}\textbf{13.22} 
 \\                       \hline     
 \multicolumn{7}{l}{~~~~~~~~~~~~~~~~~~~~~~~~~~~~~~~~~~~\textbf{Quantitative results on VKITTI2}~\cite{vkitti2}}\\
 \hline 
 %STL~\cite{unet}  &  84.50   & 5.715    & 84.50 & 5.717  &   23.10                \\ 
\multirow{9}{*}{ResNet-50 backbone}&MTL-baseline~\cite{Vandenhende_2021}        &87.75 & 5.511&88.86 &5.312 &22.27
                                                                    \\
 &Cross-stitch~\cite{cross-stitch} & 86.11&5.719 &87.50 &5.505 & 23.03\\  
&MTAN~\cite{endtoendMTL} & 89.00& 4.425& 90.00&4.197 & 20.66 \\  
&TTNet~\cite{ttnet} & 89.13& 4.440 & 90.11 & 4.188 & 20.52 \\  
&Taskonomy~\cite{taskonomy2018} & 87.52& 5.517 &88.61 & 5.400 & 22.70 \\  
&TSwitch~\cite{taskswitchnet} &89.63 & 4.399 & 92.13 & 4.155 & 19.00 \\  
&Consistency~\cite{zamir2020consistency}&  89.25  & 4.461  & 90.75 & 4.180 & 19.37                                                                             \\

&XTAM~\cite{xtam}    &93.20 & 4.274 &95.44 &4.020 &17.00
                                        \\
&TAWT~\cite{tawt}    &93.41  & 4.202& 95.96& 3.991&16.76\\  \cmidrule(lr){1-2}\cmidrule(lr){3-4}\cmidrule(lr){5-7}
\multirow{5}{*}{Swin-B V2 backbone}&XTAM~\cite{xtam}    &96.93 & 3.425 &97.58 &3.092 & 14.49
                                        \\
&TAWT~\cite{tawt}    &97.52  & 3.385& 97.92 & 3.061& 14.41\\
&ST-MTL~\cite{spatiotemporalMTL} &97.91 &3.365 & 98.22& 3.040& 14.08 \\  
%InvPT~\cite{invpt2022} & & & & & \\                                        
%Taskprompter~\cite{taskprompter2023} & & & & & \\
&MulT~\cite{MulT}   & \underline{98.03} &\underline{3.341} &\underline{98.75} & \underline{3.015} & \underline{13.95} 
                                                                     \\

&\cellcolor[HTML]{EFEFEF}\textbf{Our}   & \cellcolor[HTML]{EFEFEF}\textbf{98.51} &\cellcolor[HTML]{EFEFEF}\textbf{3.297} &\cellcolor[HTML]{EFEFEF}\textbf{99.00} &\cellcolor[HTML]{EFEFEF}\textbf{2.881} &\cellcolor[HTML]{EFEFEF}\textbf{13.02} 
 \\     \cmidrule(lr){1-2}\cmidrule(lr){3-4}\cmidrule(lr){5-7}
\multirow{5}{*}{ViT-B backbone}&XTAM~\cite{xtam}    &96.80 & 3.433  &97.41 & 3.099 & 14.57
                                        \\
&TAWT~\cite{tawt}    & 97.40  & 3.391 & 97.81 & 3.065 & 14.55\\
&ST-MTL~\cite{spatiotemporalMTL} & 97.80& 3.372 & 98.13 & 3.049 & 14.16 \\  
%InvPT~\cite{invpt2022} & & & & & \\                                        
%Taskprompter~\cite{taskprompter2023} & & & & & \\
&MulT~\cite{MulT}   & \underline{98.00} &\underline{3.349} &\underline{98.66} & \underline{3.024} & \underline{14.05} 
                                                                     \\

&\cellcolor[HTML]{EFEFEF}\textbf{Our}   & \cellcolor[HTML]{EFEFEF}\textbf{98.43} &\cellcolor[HTML]{EFEFEF}\textbf{3.303}  &\cellcolor[HTML]{EFEFEF}\textbf{98.89} &\cellcolor[HTML]{EFEFEF}\textbf{2.890} &\cellcolor[HTML]{EFEFEF}\textbf{13.13}
 \\      
\end{tabular}}
\setlength{\abovecaptionskip}{1mm}
\caption[Multitask learning results on Synthia and Vkitti2]{\textbf{Multitask learning results} on the Synthia~\cite{synthia} (Top) and Vkitti2~\cite{vkitti2} (Bottom) benchmark, respectively. Our method consistently outperforms all the ResNet-50 backbone-based MTL methods and the Swin-B V2 backbone-based methods. We also alternate the ResNet-50 backbone and the Swin-B V2 backbone with the ViT-B backbone for the best-performing methods. Bold and underlined values show the best and
second-best results, respectively.}
\label{tb:AVTaR-synthia-vkitti2-results}%
\end{table}
\subsubsection{Transformer-based Methods}
\textbf{ST-MTL~\cite{spatiotemporalMTL}}: Leveraging vision transformers, this method achieves dense predictions in an encoder-decoder setup.  \\
\textbf{InvPT~\cite{invpt2022}}: performs simultaneous modeling
of spatial positions and multiple dense prediction tasks in a unified transformer framework. \\
\textbf{Taskprompter~\cite{taskprompter2023}}: focuses on the representation learning capability of the multitask networks by using a set of task prompts. %Being a task transfer learning method, we compare against this baseline. 
\\
\textbf{MulT~\cite{MulT}}: Based on the Swin backbone, MulT uses a shared attention mechanism from a reference task that models the dependencies across the tasks in an end-to-end transformer framework. \\
\textbf{Vanilla MTL Swin~\cite{swin}}: The Vanilla MTL Swin is based on the vanilla Swin-B V2 network with a single encoder and four shared decoders and task-specific heads. \\
\textbf{1-task Swin~\cite{swin}}: We compare our performance against single-task learning networks using the baseline Swin-B V2 backbone, where each task is predicted separately by a dedicated Swin-B V2 network. This baseline is used as an Oracle in our \textit{Unsupervised Domain Adaptation (UDA)} setting.
%All the multitask baselines were trained using their best model configurations as in~\cite{Vandenhende_2021, zamir2020consistency, xtam, MulT}, respectively. 
\subsection{Metrics}
\label{sec:metrics}
We report the performances of all the models by using four task-specific metrics as follows:\\
\textbf{Semantic segmentation} uses \textit{mIoU} as the average of the per-class Intersection over Union (\%) between the ground-truth segmentation and predicted map.\\
\textbf{Depth} uses the Root Mean Square Error \textit{(RMSE)} computed between the depth label and the predicted depth map, where the RMSE metric is reported in meters over the evaluated set of images.\\
\textbf{Normal estimation} uses the absolute angle error in degrees \textit{(mErr)} between the
normal ground-truth label and normal estimation map. \\ %Except for Taskonomy~\cite{taskonomy2018}, which provides ground-truth surface normal labels, we use~\cite{pseudo-surface-normal} to get the surface normal labels from the ground-truth depth for the remaining datasets. %In particular, we unproject the pixels using the camera parameters and depth map, followed by a cross-product calculation using 2D neighbouring points. We then average the cross-product over four pairs of neighbours. 
%Note that Cityscapes~\cite{Cordts2016Cityscapes} provides disparity maps which we use to compute noisy surface normals labels.\\
\textbf{Edge estimation} uses the \textit{F1-score} between the ground-truth edges and the predicted edge maps. %Note that NYUDv2~\cite{NYU} provides ground-truth semantic edge maps.
\subsection{Training Details}
\label{sec:training-details}
 We train all the multi-tasking models with the Adam optimizer~\cite{adam-w} with $\beta_1=0.9$ and $\beta_2=0.98$; learning rate of $5.0e-5$ and a warm-up cosine learning rate schedule. The number of warmup epoch is 5 out of the total 30 training epochs. We report the average over 3 runs. We use 4 A100 40 GB GPUs for training our MTL model. 
 %For the architecture in the main paper, that is based on the Swin-B V2~\cite{swin} backbone, we use the same configurations for the pre-trained window size $=[12,12,12,6]$. We experiment with various hidden feed-forward network dimensions with 48 or 96 hidden dimensions, matching the Swin v2 configurations. Further, we experiment with different activations like Tanh, ReLU and GeLU which are applied after the feed-forward network. We settle on GeLU activation as in~\cite{ houlsby-adapter, swin} as the performances are comparable with Tanh, ReLU or GeLU. We experiment with different bottleneck sizes for the FF down and FF up in our vision adapters, including 8, 12, and 24 hidden dimensions for the bottleneck. As we will show in Table~\ref{tb:ablation-network}, the best-performing bottleneck dimension is of 12, while being the most parameter efficient.
\begin{table}[ht]
\setlength\tabcolsep{3pt}
\centering
\scalebox{0.78}{
\begin{tabular}{lllll}
%\toprule
\multicolumn{3}{l}{~~~~~~~~\textbf{Synthia~\cite{synthia}$\longrightarrow$Cityscapes}~\cite{Cordts2016Cityscapes}}  
& \multicolumn{2}{l}{~~~~~~~~\textbf{\textit{'S-D'}}}                                                     \\\cmidrule(lr){1-2}\cmidrule(lr){3-3}\cmidrule(lr){4-5}
&\multicolumn{1}{c}{\multirow{-2}{*}{\textbf{Methods}}} &
\multicolumn{1}{c}{\multirow{-2}{*}{\textbf{MTL}}}& \cellcolor[HTML]{FFCCC9}\begin{tabular}[c]{@{}l@{}}SemSeg\\ mIoU\%$\uparrow$\end{tabular} & \cellcolor[HTML]{DAE8FC}\begin{tabular}[c]{@{}l@{}}Depth\\ RMSE$\downarrow$\end{tabular}  \\
\cmidrule(lr){1-2}\cmidrule(lr){3-3}\cmidrule(lr){4-5}
\multirow{3}{*}{CNN} & MTL-baseline-UDA~\cite{Vandenhende_2021}        &\greencheck &17.26 &14.85                       \\
& Consistency-UDA~\cite{zamir2020consistency}& \greencheck  & 34.19 &  12.84               \\
&XTAM-UDA~\cite{xtam}    & \greencheck & 37.93 &11.66 
                                \\\cmidrule(lr){1-2}\cmidrule(lr){3-3}\cmidrule(lr){4-5}
\multirow{4}{*}{Transformer}& 1-task Swin-UDA~\cite{unet}  & \redcheck & 39.00   &  11.03                                                                             \\
 &MulT-UDA~\cite{MulT}   &\greencheck &\underline{42.12} &\underline{09.55} 
                              \\
&\cellcolor[HTML]{EFEFEF}\textbf{Our-UDA}   & \cellcolor[HTML]{EFEFEF}\greencheck &\cellcolor[HTML]{EFEFEF}\textbf{50.03} &\cellcolor[HTML]{EFEFEF}\textbf{06.99} \\ 
& \cellcolor[HTML]{C0C0C0}1-task Swin-target (Oracle)~\cite{swin}  & \cellcolor[HTML]{C0C0C0}\redcheck & \cellcolor[HTML]{C0C0C0}75.97 &\cellcolor[HTML]{C0C0C0}06.65  
\end{tabular}}
\setlength{\abovecaptionskip}{1mm}
\caption{\textbf{Unsupervised Domain Adaptation (UDA)} results for Synthia~\cite{synthia}$\rightarrow$Cityscapes~\cite{Cordts2016Cityscapes}. Our model outperforms all the baselines. Bold and underlined values show the best and
second-best results, respectively.}
   \label{tb:uda-syn2cityscapes}%
   \vspace{-10pt}
\end{table}
%%%%%%%%%%%%%%%%%%%%% MS-Coco to DCM %%%%%%%%%
\begin{table} [ht]
\setlength\tabcolsep{3pt}
\centering
\scalebox{0.8}{
\begin{tabular}{lllll}
%\toprule
%&\multicolumn{2}{l}{~~~~~~~~\textbf{MS-Coco~\cite{mscoco}$\longrightarrow$DCM}~\cite{dcm}} 
%& \multicolumn{2}{l}{~~~~~~~~\textbf{$\textit{'S-D'}$}}                                                   \\
%\cmidrule(lr){1-2}\cmidrule(lr){3-3}\cmidrule (lr){4-5}
&\multicolumn{1}{c}{\multirow{-2}{*}{\textbf{Methods}}} &
\multicolumn{1}{c}{\multirow{-2}{*}{\textbf{MTL}}}& \cellcolor[HTML]{FFCCC9}\begin{tabular}[c]{@{}l@{}}SemSeg\\ mIoU\%$\uparrow$\end{tabular} & \cellcolor[HTML]{DAE8FC}\begin{tabular}[c]{@{}l@{}}Depth\\ RMSE$\downarrow$\end{tabular}  \\
\cmidrule(lr){1-2}\cmidrule(lr){3-3}\cmidrule (lr){4-5}
%\rowcolor[HTML]{C0C0C0}STL-target (Oracle)~\cite{unet}  &  & & \\
%\rowcolor[HTML]{C0C0C0}STL-source~\cite{unet}&  &  &\\
%\hline
\multirow{2}{*}{CNN}&Consistency~\cite{zamir2020consistency}  & \greencheck & 18.22
&    1.884                 \\
&XTAM~\cite{xtam}       &\greencheck & 18.63 &  1.795                   \\\cmidrule(lr){1-2}\cmidrule(lr){3-3}\cmidrule (lr){4-5}
\multirow{4}{*}{Transformer}&1-task Swin~\cite{swin}& \redcheck  & 20.76  & 1.508     
                              \\
&ST-MTL   & \greencheck & 22.49 & 1.446
                             \\

&MulT~\cite{MulT}   &\greencheck &\underline{24.02} &\underline{1.300}
                              \\

&\cellcolor[HTML]{EFEFEF}\textbf{Our}   &\cellcolor[HTML]{EFEFEF} \greencheck &\cellcolor[HTML]{EFEFEF}\textbf{27.11} &\cellcolor[HTML]{EFEFEF}\textbf{1.182} 
 \\                     
\end{tabular}}
\setlength{\abovecaptionskip}{1mm}
\caption {\textbf{Generalization results} of our model trained on MS-Coco~\cite{mscoco} and applied to DCM Comics~\cite{dcm}. Our method outperforms all the baselines. Bold and underlined values show the best and
second-best results, respectively.}
   \label{tb:uda-dcm}%
\vspace{-5pt}
\end{table}

%%%%%%%%%%%%%%%%%%%%%%%%%%%%%% Additional Ablation on Adapters %%%%%%%%%%%%%%%%%%%%%%%%%%%%%%%%%%%%%%%%%%%%%%%
\section{Visualizing Task-adapted Attention (TAA)}
\label{sec:TAA-visualizing}
\begin{figure}[ht!]
\centering
{\includegraphics[ width=1.0\linewidth ]{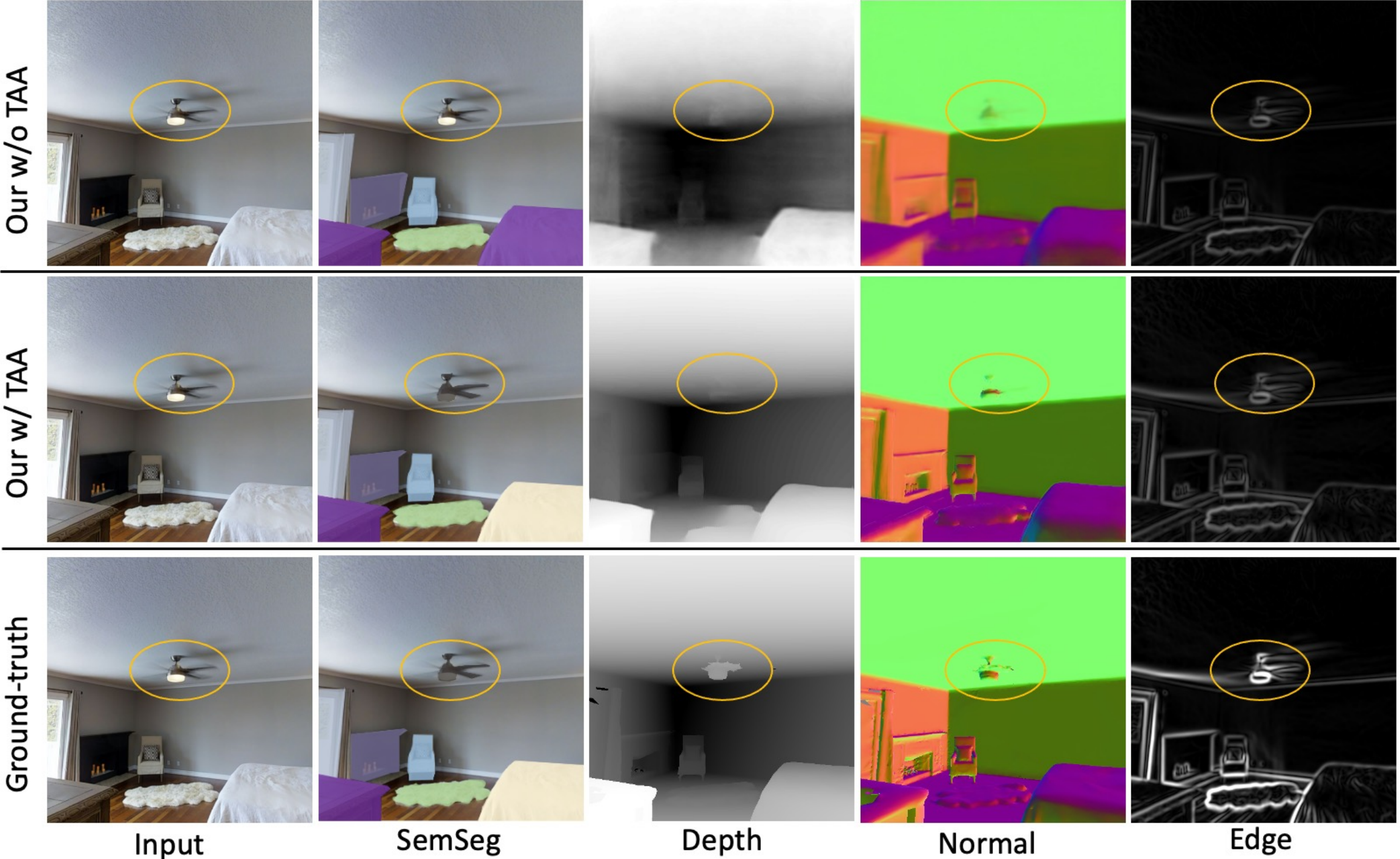}}
\setlength{\abovecaptionskip}{0pt}
\caption{\textbf{Effect of TAA on
our model.} The yellow-circled region shows how our model
with TAA improves, for instance, the semantic segmentation performance, where the fan mask is correctly classified in our predictions. However, our model without TAA fails to segment the fan. Best viewed on screen and when zoomed in. 
}\label{fig:effect-of-TAA}%\vspace{-10pt}
\end{figure}

We visualize the task-adapted attention for each tasks in the 'S-D-N-E' setting and show that it differs from the existing self-attention mechanism in Figure~\ref{fig:TAA-visualization}. TAA is more task-specific compared with self-attention, thanks to its task conditioning from TROA. 
In Figure~\ref{fig:effect-of-TAA}, we demonstrate the effect of TAA that learns the task affinities and improves the prediction for each task. For instance, TAA improves the semantic segmentation performance where the bed mask is correctly classified in our predictions as in the ground truth. Without TAA the bed is segmented as a table.
\vspace{-8pt}
%We now explain the TROA mechanism in detail.
\begin{figure*}[ht!]
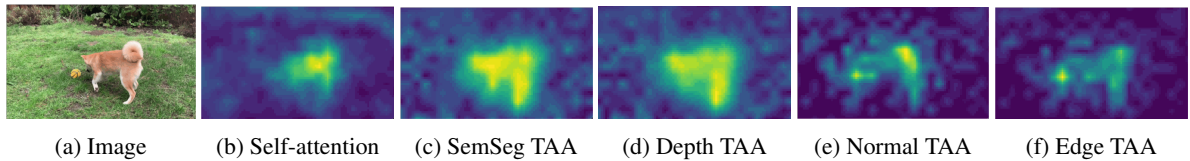

    \centering
    \subcaptionbox{Image}
    {\includegraphics[width=0.144\linewidth]{main/images/input-TAA.png}}
       \subcaptionbox{Self-attention}
    {\includegraphics[width=0.146\linewidth]{main/images/self-attention.png}}
         \subcaptionbox{SemSeg TAA}
    {\includegraphics[width=0.146\linewidth]{main/images/semseg-TAA.png}}
      \subcaptionbox{Depth TAA}
    {\includegraphics[width=0.146\linewidth]{main/images/depth-TAA.png}}
      \subcaptionbox{Normal TAA}
    {\includegraphics[width=0.146\linewidth]{main/images/normal-taa.png}}
       \subcaptionbox{Edge TAA}
    {\includegraphics[width=0.146\linewidth]{main/images/edge-TAA.png}}
    \caption[Visualizing TAA versus the self-attention]{\textbf{Visualizing TAA versus the self-attention} of the Swin-B V2 encoder layer T18. We show that TAA has more task-specific attention compared to self-attention in the encoder. Here, our model that is used for visualization is trained on MS-Coco~\cite{mscoco} with depth, surface normal, and edge labels from~\cite{midas, pseudo-surface-normal, rindnet}, respectively.}
    \label{fig:TAA-visualization}
\end{figure*}

\section{Ablation Study}
\label{sec:ablation}
\subsection{Effect of Different Modules of Our Network}
\begin{table}[ht]
\setlength\tabcolsep{3pt}
\centering
\scalebox{0.75}{
\begin{tabular}{cccccc}
\textbf{\begin{tabular}[c]{@{}c@{}}Model \\ Changes\end{tabular}}                           & \cellcolor[HTML]{FFCCC9}\begin{tabular}[c]{@{}l@{}}SemSeg\\ mIoU\%$\uparrow$\end{tabular} & \cellcolor[HTML]{DAE8FC}\begin{tabular}[c]{@{}l@{}}Depth\\ RMSE$\downarrow$\end{tabular} & \cellcolor[HTML]{FFFFC7}\begin{tabular}[c]{@{}l@{}}Normal\\ mErr. $\downarrow$\end{tabular} & \cellcolor[HTML]{C8E685}\begin{tabular}[c]{@{}l@{}}Edges\\ F1\%$\uparrow$\end{tabular} & \cellcolor[HTML]{C0C0C0}\textbf{\begin{tabular}[c]{@{}c@{}}\#Parameters\\ (Millions)\end{tabular}} \\ 
\cmidrule(lr){1-1}\cmidrule(lr){2-5}\cmidrule(lr){6-6}
{\color[HTML]{333333} \begin{tabular}[c]{@{}c@{}}Vanilla MTL Swin~\cite{swin}\end{tabular}} & {\color[HTML]{333333} 48.13}            & {\color[HTML]{333333} 0.4956}          & {\color[HTML]{333333} 24.53}             & {\color[HTML]{333333} 54.88}           & {\color[HTML]{333333} 348.0}                                                                         \\
\rowcolor[HTML]{FFFFFF} 
\begin{tabular}[c]{@{}c@{}}+ TAA \end{tabular}                  & 59.42                                   & 0.4111                                 & 18.55                                    & 69.91                                  & 408.0                                                                                               \\
\rowcolor[HTML]{FFFFFF} 
+ bottleneck                                                                                & 59.93                                   & 0.4066                                 & 18.08                                    & 70.32                                  & 104.0                                                                                                \\
\rowcolor[HTML]{EFEFEF}
\begin{tabular}[c]{@{}c@{}}+ TSN \textbf{(Our)}\end{tabular}                       & \textbf{60.80}                                  & \textbf{0.3903}                                 & \textbf{17.13}                                    & \textbf{71.09}                                  & \textbf{105.7}                                                                                          \\   
\end{tabular}}
\setlength{\abovecaptionskip}{1mm}
\caption{\textbf{Ablation study of the different components of our network} on the Taskonomy benchmark~\cite{taskonomy2018}.
We show from left to right, the performances of each added module on multiple tasks.
%on the tasks of semantic segmentation, 2D depth, surface normal estimation, edge estimation, respectively. %We add each component, one-by-one, starting from an end-to-end trained Swin backbone and then, freezing the Swin encoder to study the effect of our vision adapter conditioned on TAA. 
Our TAA and TSN components improves the performance consistently across all the tasks while the bottleneck reduces the number of parameters.}
   \label{tb:ablation}%
%\vspace{-10pt}
\end{table}

In Table~\ref{tb:ablation}, we present the results of an ablation study to determine which component of our method has the largest positive gain on the different task predictions. Starting from a Swin baseline that employs the Swin encoder and task-specific decoders as is --- initialized with the pre-trained ImageNet 22k weights ---  and trained using random task sampling, we find that the task learning interferes with each other in the absence of task-adapted attention (TAA). Note that in this setup, the trainable encoder layers and decoder layers are jointly trained with just the Vanilla Swin self-attention (SA) as in~\cite{swin}, therefore lacking in task-adapted attention (TAA). We then add our model's components, one by one, starting with TAA conditioned on the task affinity weights from TROA. However, in this part, we do not add the adapter bottleneck, i.e., FFup and FFdown as seen in Figure~\ref{fig:vision-adapter} of the main paper. We then add the bottleneck and finally, add the Task-Scaled Norm (TSN).  We report both the performances and parameters required for each added component. Not only does each module lift the task performances but the introduction of the adapter bottleneck significantly reduces the number of parameters. 

Note that {TAA variants} with operations like Matmul or concatenation between the $A'(.)$ matrix and $q.k^T$ matrix are extremely computationally expensive, scaling non-linearly with an increase in the number of tasks. Hence, we do not report them. 
Also with a TROA variant where 'w= constant' for all tasks, the model fails to leverage the task inter-dependencies~\cite{MulT, standley2019} and defaults to self-attention that is shifted by a constant. Failing to account for the task relationships, is \textit{not} a typical multitask setting~\cite{MulT, standley2019, zamir2020consistency}. We, therefore, do not report this TROA variant.
\begin{table*}[ht]
\setlength\tabcolsep{3pt}
\centering
\scalebox{0.80}{
\begin{tabular}{llcclllll}
\hline
\multicolumn{1}{c}{\begin{tabular}[c]{@{}c@{}}\textbf{Backbone} \\ \textbf{Size}\end{tabular}} & \multicolumn{1}{c}{\begin{tabular}[c]{@{}c@{}}\textbf{Pre-trained}\\ \textbf{Initialization}\end{tabular}} & \begin{tabular}[c]{@{}c@{}}\textbf{FFN} \\ \textbf{Dimension}\end{tabular} & \begin{tabular}[c]{@{}c@{}}\textbf{Bottleneck}\\ \textbf{Size}\end{tabular} & \multicolumn{1}{c}{\cellcolor[HTML]{FFCCC9}\begin{tabular}[c]{@{}l@{}}SemSeg\\ mIoU\%$\uparrow$\end{tabular}} & \multicolumn{1}{c}{\cellcolor[HTML]{DAE8FC}\begin{tabular}[c]{@{}l@{}}Depth\\ RMSE$\downarrow$\end{tabular}} & \cellcolor[HTML]{FFFFC7}\begin{tabular}[c]{@{}l@{}}Normal\\ mErr. $\downarrow$\end{tabular} & \multicolumn{1}{c}{\cellcolor[HTML]{C8E685}\begin{tabular}[c]{@{}l@{}}Edges\\ F1\%$\uparrow$\end{tabular} } & \begin{tabular}[c]{@{}l@{}}Parameter\\ (in millions)\end{tabular} \\ \cmidrule(lr){1-1}\cmidrule(lr){2-2}\cmidrule (lr){3-3}\cmidrule (lr){4-4}\cmidrule (lr){5-8}\cmidrule (lr){9-9}
                                       \multirow{12}{*}{\cellcolor{cosmiclatte}\begin{tabular}[c]{@{}l@{}}Swin-B v2\end{tabular}}  & \multirow{6}{*}{ImageNet 1K}                                                              & \multirow{3}{*}{48}                                      & 8                                                         &50.52                                                                             & 0.4511                                                                           & 22.13                                                                  & 62.16                                                                         &                                                                  103.0\\
                                                                {\cellcolor{cosmiclatte}\begin{tabular}[c]{@{}l@{}}\end{tabular}}        &                                                                                           &                                                          & 12                                                        & 56.55                                                                            & 0.4230                                                                           & 20.39                                                                   & 67.03                                                                         &  103.6                                                                \\
{\cellcolor{cosmiclatte}\begin{tabular}[c]{@{}l@{}}\end{tabular}}                                                                            &                                                                                           &                                                          & 24                                                        & 56.85                                                                            &  0.4191                                                                          & 20.22                                                                  &  67.66                                                                        & 104.4                                                                  \\ \cline{3-3} \cline{4-9} 
                                                        {\cellcolor{cosmiclatte}\begin{tabular}[c]{@{}l@{}}\end{tabular}}                     &                                                                                           & \multirow{3}{*}{96}                                      & 8                                                         &    52.18                                                                         & 0.4410                                                                           & 22.04                                                                  &  63.60                                                                        &  105.2                                                                 \\
                                                {\cellcolor{cosmiclatte}\begin{tabular}[c]{@{}l@{}}\end{tabular}}                             &                                                                                           &                                                          & 12                                                        &  58.10                                                                           &  0.4105                                                                          & 19.15                                                                  &  69.07                                                                        &   105.7                                                                \\
                                     {\cellcolor{cosmiclatte}\begin{tabular}[c]{@{}l@{}} Swin-B v2 transformer\end{tabular}}                                        &                                                                                           &                                                          & 24                                                        & 58.81                                                                            & 0.4095                                                                           & 19.08                                                                  &  69.19                                                                        &    109.4                                                               \\ \cline{2-9} 
                                                    {\cellcolor{cosmiclatte}\begin{tabular}[c]{@{}l@{}}\end{tabular}}        & \multirow{6}{*}{ImageNet 22K}                                                             & \multirow{3}{*}{48}                                      & 8                                                         & 52.11                                                                            &  0.4392                                                                          & 20.88                                                                  & 64.29                                                                         &   103.0                                                                \\
                                                            {\cellcolor{cosmiclatte}\begin{tabular}[c]{@{}l@{}}\end{tabular}}              &                                                                                         &                                                          & 12                                                        &  58.72                                                                             & 0.4051                                                                           & 18.62                                                                  & 69.10                                                                         &   103.6                                                                \\
                                                   {\cellcolor{cosmiclatte}\begin{tabular}[c]{@{}l@{}}\end{tabular}}             &                                                                                           &                                                         & 24                                                        &58.87                                                                            &0.3998                                                                           &18.20                                                                  &69.90                                                                         & 104.4                                                                  \\ \cline{3-9} 
                                                           {\cellcolor{cosmiclatte}\begin{tabular}[c]{@{}l@{}}\end{tabular}}              &                                                                                           & \multirow{3}{*}{96}                                      & 8                                                         &54.19                                                                             &0.4220                                                                           &20.03                                                                   &65.65                                                                          & 105.2                                                                  \\
                                        {\cellcolor{cosmiclatte}\begin{tabular}[c]{@{}l@{}}\end{tabular}}                 &                                                                                           &                                                          & \cellcolor[HTML]{EFEFEF}12                                                        &        \cellcolor[HTML]{EFEFEF}60.80                                                                   &    \cellcolor[HTML]{EFEFEF}0.3903                                                                       &   \cellcolor[HTML]{EFEFEF}17.13                                                               &   \cellcolor[HTML]{EFEFEF}71.09                                                                     &  \cellcolor[HTML]{EFEFEF}105.7                                                              \\
                                    {\cellcolor{cosmiclatte}\begin{tabular}[c]{@{}l@{}}\end{tabular}}                  &                                                                                           &                                                          & 24                                                       &  60.83                                                                           & 0.3892                                                                           & 17.09                                                                  &  71.13                                                                        & 109.4                                                                                                        \\ \midrule
                                                                    \multirow{6}{*}{{\cellcolor{cosmiclatte}\begin{tabular}[c]{@{}l@{}}Swin-L V2 \end{tabular}} }                                                   & \multirow{6}{*}{ImageNet 22K}                                                              & \multirow{3}{*}{48}                                      & 8                                                         &52.20                                                                             & 0.4385                                                                           & 20.81                                                                  & 64.38                                                                         &   360.0                                                                \\
                                                   {\cellcolor{cosmiclatte}\begin{tabular}[c]{@{}l@{}}\end{tabular}}                            &                                                                                           &                                                          & 12                                                        & 59.01                                                                            &  0.4042                                                                          & 18.50                                                                  &  69.22                                                                        &    364.2                                                               \\
                                        {\cellcolor{cosmiclatte}\begin{tabular}[c]{@{}l@{}} Swin-L v2 transformer\end{tabular}}                                       &                                                                                           &                                                          & 24                                                        & 58.92                                                                            &0.3990                                                                            & 18.11                                                                  &  69.98                                                                        &    367.8                                                               \\ \cline{3-9} 
                                   {\cellcolor{cosmiclatte}\begin{tabular}[c]{@{}l@{}}\end{tabular}}                                            &                                                                                           & \multirow{3}{*}{96}                                      & 8                                                         & 54.31                                                                            &  0.4150                                                                          & 19.91                                                                  &   65.88                                                                       &    380.8                                                               \\
                                 {\cellcolor{cosmiclatte}\begin{tabular}[c]{@{}l@{}}\end{tabular}}                                              &                                                                                           &                                                          & 12                                                        & 60.89                                                                            &  0.3892                                                                          &17.02                                                                   &    71.15                                                                      &   383.4                                                                \\
                                            {\cellcolor{cosmiclatte}\begin{tabular}[c]{@{}l@{}}\end{tabular}}                                   &                                                                                           &                                                          & 24                                                        &  60.95                                                                           & 0.3880                                                                           & 16.90                                                                  &  71.33                                                                        &   388.0                                                                                                                    \\ \hline
\end{tabular}}
\caption[Ablation study for the network sizes]{\textbf{Ablation study for the network sizes} on the Taskonomy~\cite{taskonomy2018} benchmark for $\textit{'S-D-N-E'}$ task set. We study the effect of different Swin backbone network sizes, different pre-trained initializations, 2 different feed-forward network (FFN) sizes, and 3 different bottleneck sizes on our model, respectively. As shown, in the grey row, we select the model which outperforms the baselines while being parameter efficient. Note Swin-L does not have pre-trained models with ImageNet 1K.}
   \label{tb:ablation-network}%
%\end{table*}
%\begin{table*}[ht]
\vspace{10pt}
\setlength\tabcolsep{3pt}
\centering
\scalebox{0.8}{
\begin{tabular}{lllllllll}
\hline
\multicolumn{4}{c}{\textbf{Adapter Placement}} & \multicolumn{5}{c}{\textbf{'S-D-N-E'}}                                                                                                                                                                                                                 \\
\textbf{Stage 1}  & \textbf{Stage 2} & \textbf{Stage 3} & \textbf{Stage 4} & \multicolumn{1}{c}{\cellcolor[HTML]{FFCCC9}\begin{tabular}[c]{@{}l@{}}SemSeg\\ mIoU\%$\uparrow$\end{tabular}} & \multicolumn{1}{c}{\cellcolor[HTML]{DAE8FC}\begin{tabular}[c]{@{}l@{}}Depth\\ RMSE$\downarrow$\end{tabular}} & \cellcolor[HTML]{FFFFC7}\begin{tabular}[c]{@{}l@{}}Normals\\ mErr. $\downarrow$\end{tabular} & \multicolumn{1}{c}{\cellcolor[HTML]{C8E685}\begin{tabular}[c]{@{}l@{}}Edges\\ F1\%$\uparrow$\end{tabular}} & \begin{tabular}[c]{@{}l@{}}Parameter\\ (in millions)\end{tabular} \\ \hline
        \greencheck &         &         &         & 51.05     & 0.4913                                                     & 28.11                                                        & 50.34      &97.20\\
         &    \greencheck      &         &         & 51.11    & 0.4899                                                     &  27.02                                                       & 54.99          &94.00\\
         &         &      \greencheck    &         & 58.85                                                    &0.4001                                                      & 18.23                                                        & 69.88      &159.4\\
         &         &         &    \greencheck      & 54.07                                                      &0.4412                                                      & 20.95                                                        & 65.77  &92.60 \\ \hline
         &    \greencheck     &         &   \greencheck      &57.81                                                       &  0.4121                                                    & 20.03                                                        & 66.13          &119.0\\
    \rowcolor[HTML]{C0C0C0}
         &         &     \greencheck     &    \greencheck      & {60.85}                                                     &     {0.3888}                                                &   {17.07}                                                     &   {71.21}       & 163.0\\
         &     \greencheck     &    \greencheck     &    \greencheck      &    60.89                                                   &  0.3885                                                    & 17.01                                                        & 71.28      &188.0\\ 
   \greencheck        &     \greencheck     &    \greencheck     &    \greencheck      & {60.92}                                                      &  {0.3884}                                                 & {16.98}                                                       & {71.31}      &227.0\\
         \hline
\end{tabular}}
\setlength{\abovecaptionskip}{1mm}
\caption{\textbf{Ablation study for varying the placement of our adapters} as well as varying the overall number of adapters across the different Swin encoder stages. In this setting, our adapters are placed at \emph{every} transformer layer for a given stage if that stage is marked with a (\greencheck). The \textcolor{darkgrey}{\textbf{grey}} row is \emph{not} our model.  The upper part shows our vision transformer adapters perform better at later stages in the encoder (i.e. stages 3 and 4). Applying adapters to more Swin encoder stages leads to a small boost at the cost of more parameters.}
   \label{tb:ablation-adapter}%
\vspace{10pt}
%\end{table*}
%\begin{table*}[ht]
\setlength\tabcolsep{3pt}
\centering
\scalebox{0.8}{
\begin{tabular}{lllllll}
\hline
\multicolumn{2}{c}{\textbf{Adapter Placement}} & \multicolumn{5}{c}{\textbf{'S-D-N-E'}}                                                                                                                                                                                                                 \\
 \textbf{Stage 3} & \textbf{Stage 4} & \multicolumn{1}{c}{\cellcolor[HTML]{FFCCC9}\begin{tabular}[c]{@{}l@{}}SemSeg\\ mIoU\%$\uparrow$\end{tabular}} & \multicolumn{1}{c}{\cellcolor[HTML]{DAE8FC}\begin{tabular}[c]{@{}l@{}}Depth\\ RMSE$\downarrow$\end{tabular}} & \cellcolor[HTML]{FFFFC7}\begin{tabular}[c]{@{}l@{}}Normals\\ mErr. $\downarrow$\end{tabular} & \multicolumn{1}{c}{\cellcolor[HTML]{C8E685}\begin{tabular}[c]{@{}l@{}}Edges\\ F1\%$\uparrow$\end{tabular}} & \begin{tabular}[c]{@{}l@{}}Parameter\\ (in millions)\end{tabular} \\ \hline
        \rowcolor[HTML]{EFEFEF}    Layers 15-18     &   Layers 1-2 (all)      &    ~\underline{60.80}                                                   & \underline{0.3903}                                                  &  \underline{17.13}                                                       &  \underline{71.09}    & \textbf{105.7}\\
         Layers 1-18 (all)    &    Layers 1-2 (all)      &\textbf{ 60.85}                                                     &      \textbf{0.3888}                                                &    \textbf{17.07}                                                     &   \textbf{71.21}        & \underline{163.0}\\
        
         \hline
\end{tabular}}
\setlength{\abovecaptionskip}{1mm}
\caption{\textbf{Ablation study for the Swin encoder stages that applies our vision adapters.} We study the effect of appending adapters to the later Swin encoder layers as opposed to all the Swin encoder layers in Stage 3. Following this setting, we report the performances and number of parameters for the \textit{'S-D-N-E'} setting on the Taskonomy~\cite{taskonomy2018} benchmark. The upper row \textbf{(our model)} shows our vision transformer adapters are more parameter efficient when located in the later transformer layers while giving a comparable performance with those reported in the bottom row. Bold and underlined values show the best and second-best results, respectively.}
   \label{tb:ablation-adapter-layers}%
\vspace{-10pt}
\end{table*}

Furthermore, in Table~\ref{tb:ablation-network}, we study the effect of different Swin V2~\cite{swin} backbones such as Swin-B and Swin-L, different pre-trained initializations, i.e. Imagenet-1K and Imagenet-22K, various hidden feed-forward network (FFN) dimensions with 48 or 96 hidden dimensions, and different bottleneck sizes for the FF down and FF up in our vision transformer adapters. We observe that the configuration of Swin-B backbone initialized with ImageNet-22K, an FFN with a hidden dimension of 96 and a bottleneck dimension of 12 achieves an improved performance across all tasks. Other configurations with a larger Swin network, a larger FFN dimension, and a larger bottleneck size give slight performance gains but they are parametrically costly.
\paragraph{Effect of Adapter Placement and Number of Adapters.} In Table~\ref{tb:ablation-adapter}, we study the effect of varying the placement of the vision adapters, as well as varying the overall number of adapters in the different stages. We append adapters to \emph{every} transformer layer for a given stage. We show that our  adapters are more efficient when located later in the encoder stages (i.e. stage 3 or 4), thereby leveraging the richer semantics. Drawing motivation from Table~\ref{tb:ablation-adapter}, we study the effect of applying the vision adapters to fewer transformer layers as opposed to all of them, in Stage 3. In Table~\ref{tb:ablation-adapter-layers}, we apply the adapters to the later layers of stage 3, guided by the principle of extracting richer semantics. This configuration achieves comparable performance on all four tasks while significantly reducing the number of parameters. Therefore, we use this layout as our model, where the adapters are applied to transformer layers 15-18 in Stage 3 and layers 1-2 in Stage 4, respectively.%\balance

%\section{ Effect of the training vs. testing data ratio}
\section{Additional Qualitative Results}
\label{sec:addtional-qualitative}
\begin{figure*}[ht]
\centering
{\includegraphics[ width=1.0\linewidth ]{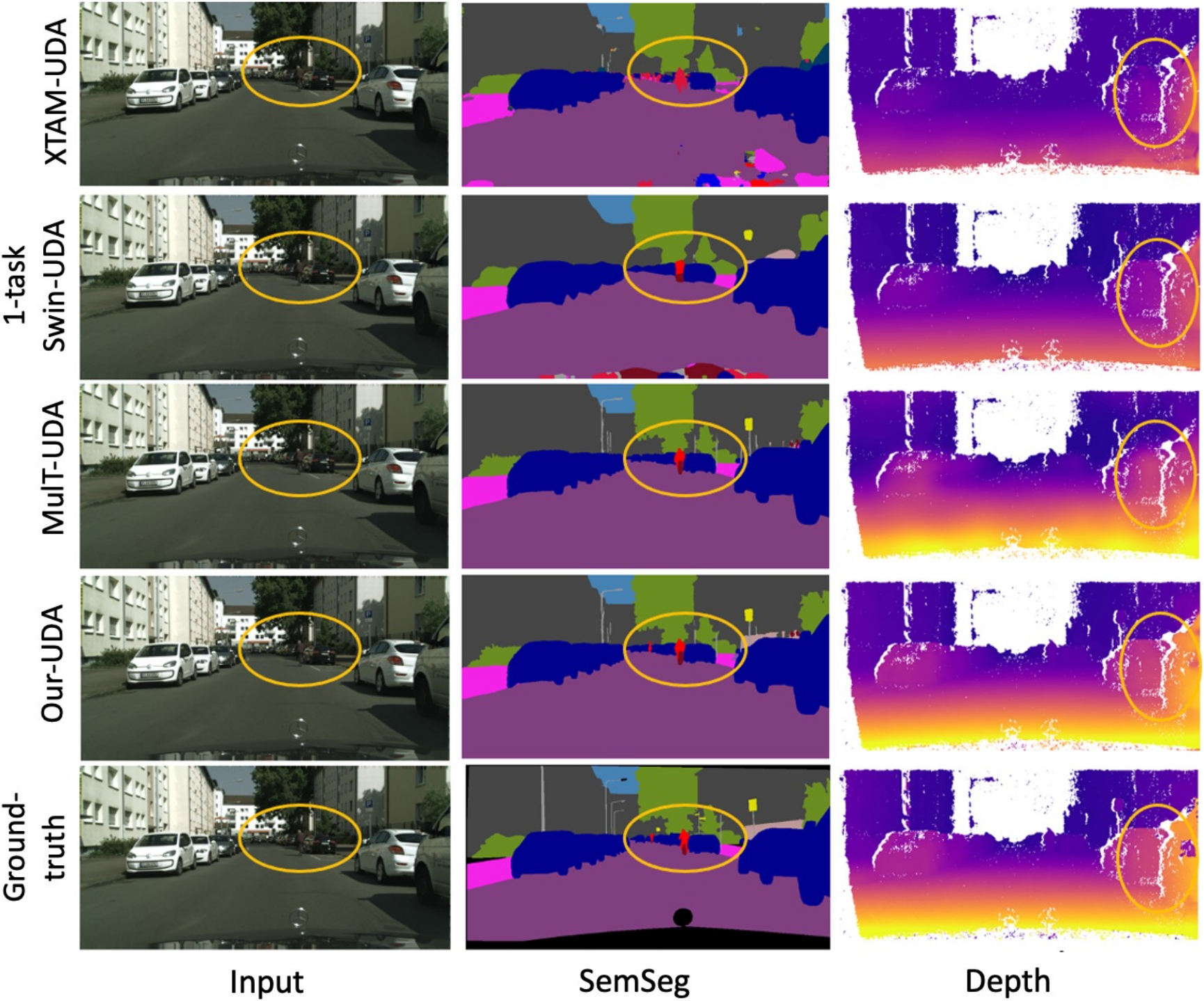}}
\setlength{\abovecaptionskip}{0pt}
\caption[UDA results in the real domain]{\textbf{Unsupervised Domain Adaptation (UDA)} results of the best-performing methods in Table~\ref{tb:uda-syn2cityscapes} on Synthia~\cite{synthia}$\rightarrow$Cityscapes~\cite{Cordts2016Cityscapes}.  Our model outperforms the CNN-based baseline (XTAM-UDA~\cite{xtam}) and the Swin-B V2-based baselines (1-task Swin-UDA~\cite{swin}, MulT-UDA~\cite{MulT}), respectively. For instance, our method can predict the depth of the car tail light, unlike the baselines. Best seen on screen and zoomed
within the yellow circled region.}\label{fig:uda}
\end{figure*}

\begin{figure*}[h]
\centering
{\includegraphics[ width=1.0\linewidth ]{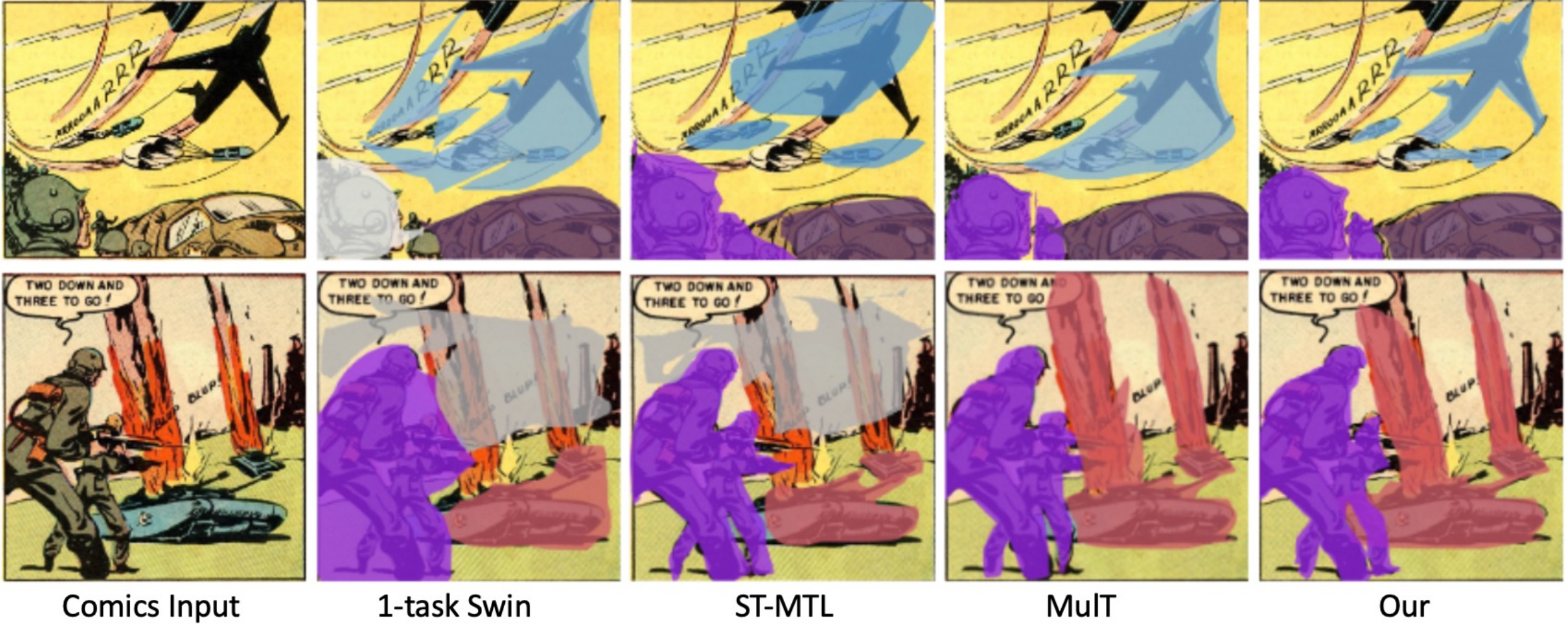}}
\setlength{\abovecaptionskip}{0pt}
\caption{\textbf{Generalization of our model trained on MS-Coco~\cite{mscoco} and applied to DCM comics~\cite{dcm} for segmentation.} Our method outperforms both the 1-task Swin and the MTL models~\cite{spatiotemporalMTL, MulT}, respectively. For instance, the airplane is more accurately segmented than the one in the baselines. All the methods are based on the same Swin-B V2 backbone. We show the best-performing methods in Table~\ref{tb:uda-dcm}. Best viewed on screen and when zoomed in.
}\label{fig:seg-dcm-generalization}
%\end{figure*}
%\begin{figure*} [h]
\vspace{5pt}
\centering
{\includegraphics[ width=1.0\linewidth ]{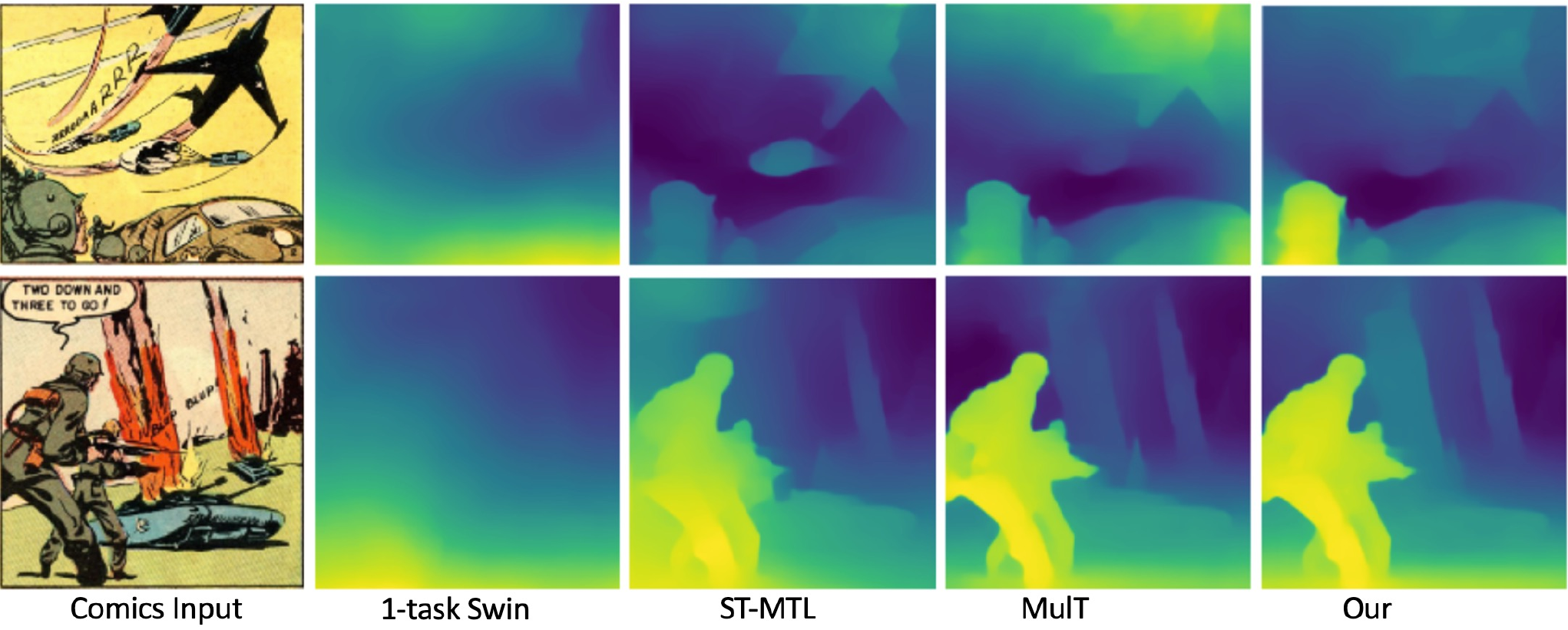}}
\setlength{\abovecaptionskip}{0pt}
\caption{\textbf{Generalization of our model trained on MS-Coco~\cite{mscoco} and applied to DCM comics~\cite{dcm} for depth.} Our method outperforms both the 1-task Swin~\cite{swin} and the MTL baselines~\cite{spatiotemporalMTL, MulT}, respectively. For instance, our method correctly separates the foreground depth plane from the background, unlike the baselines. All the methods are based on the same Swin-B V2 backbone. We show the best-performing methods in Table~\ref{tb:uda-dcm}. Best viewed on screen and when zoomed in.
}\label{fig:depth-dcm-generalization}
\end{figure*}

We compare the best-performing methods in the UDA setting with the source domain as Synthia~\cite{synthia} and the target domain as Cityscapes~\cite{Cordts2016Cityscapes}. Our method outperforms all the baselines as shown in Figure~\ref{fig:uda}.
We qualitatively compare the best-generalizing methods to a novel domain of comics for segmentation in Figure~\ref{fig:seg-dcm-generalization} and depth in Figure~\ref{fig:depth-dcm-generalization}, respectively.

\begin{figure*}[h]
\centering
{\includegraphics[ width=1.0\linewidth ]{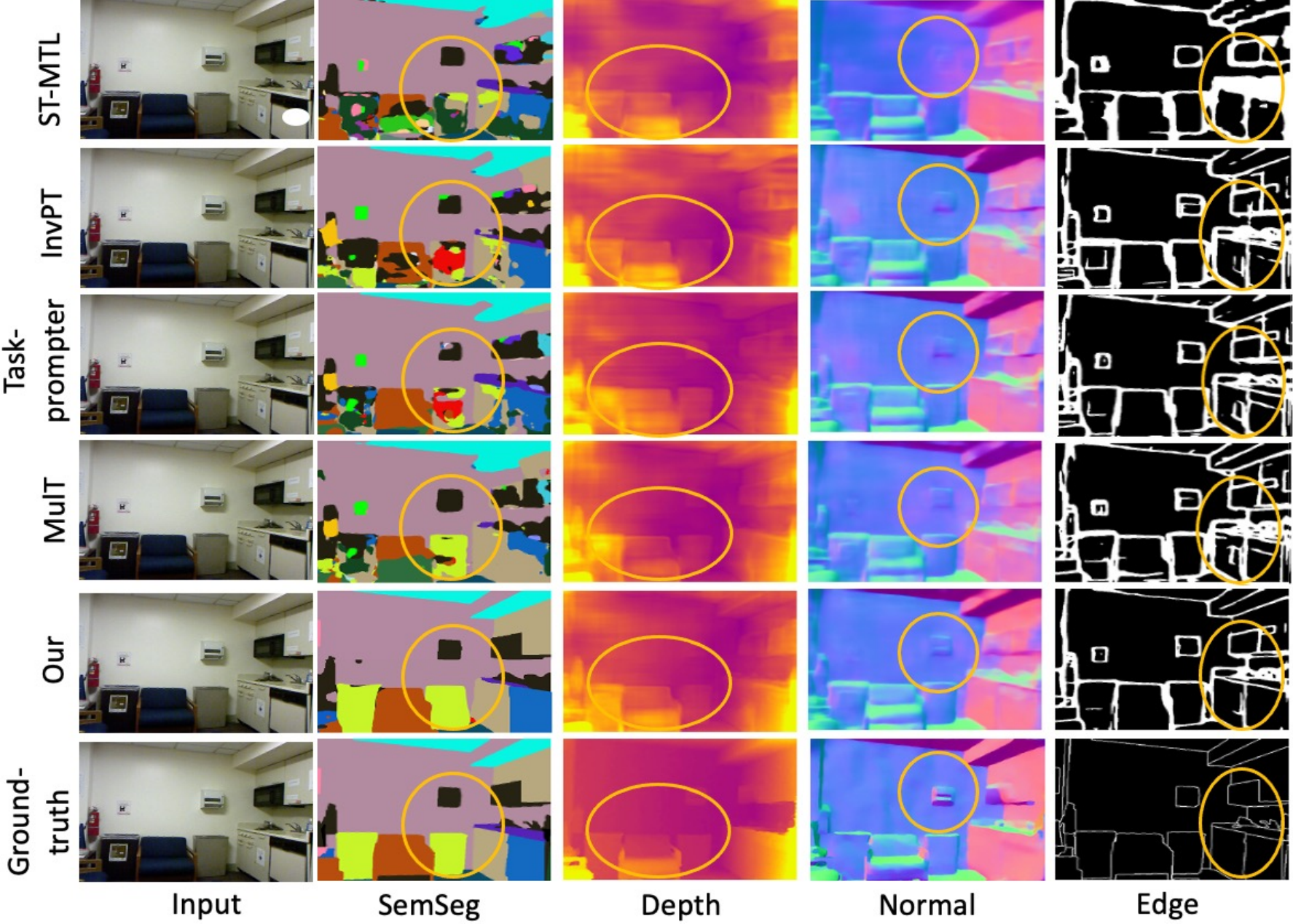}}
\setlength{\abovecaptionskip}{0pt}
\caption{{\textbf{Multitask Learning comparison on NYUDv2}~\cite{NYU} benchmark in the \textit{'S-D-N-E'} setting. Our model outperforms all the multitask baselines, i.e. ST-MTL~\cite{spatiotemporalMTL}, InvPT~\cite{invpt2022}, Taskprompter~\cite{taskprompter2023}, and MulT~\cite{MulT}, respectively. For instance, our model correctly segments and predicts the surface normal of the elements within the yellow-circled region, unlike the baseline. All the methods are based on the same Swin-B V2 backbone. We show the best-performing methods in Table~\ref{tb:AVTaR-nyudv2-results}. Best seen on screen and zoomed in.} 
}\label{fig:nyu-qr-sdne}
\end{figure*}

\begin{figure*}[h]
\centering
{\includegraphics[ width=1.0\linewidth ]{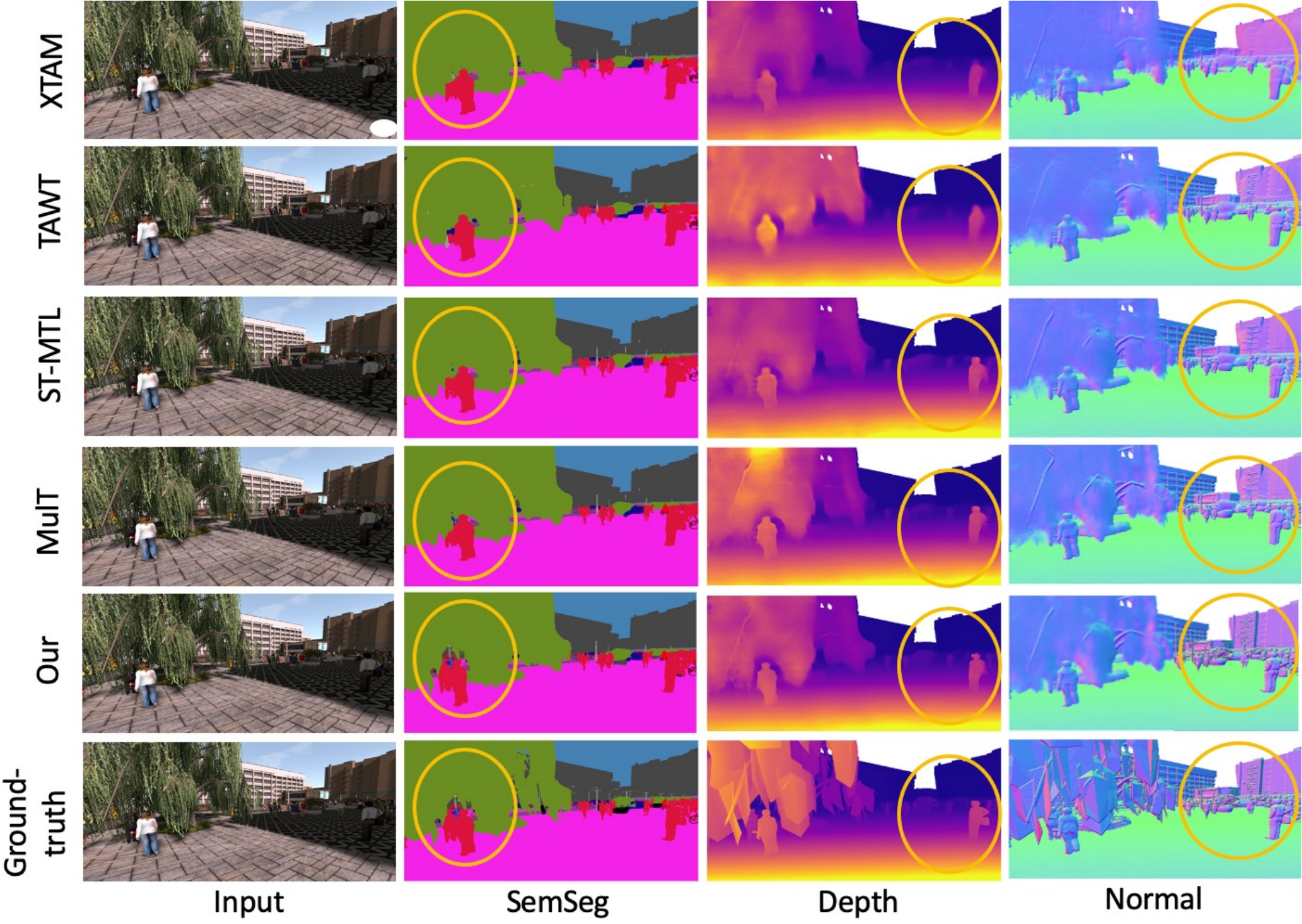}}
\setlength{\abovecaptionskip}{0pt}
\caption{{\textbf{Multitask Learning comparison on Synthia}~\cite{synthia} benchmark in the \textit{'S-D-N'} setting. Our model outperforms all the multitask baselines. For instance, our method correctly segments the people, unlike the baselines. All the methods are based on the same Swin-B V2 backbone. We show the best-performing methods in Table~\ref{tb:AVTaR-synthia-vkitti2-results}, i.e. XTAM~\cite{xtam}, TAWT~\cite{tawt}, ST-MTL~\cite{spatiotemporalMTL}, and MulT~\cite{MulT}, respectively. Best seen on screen and zoomed within the yellow circled regions. } 
}\label{fig:synthia-qr-sdn}\vspace{2pt}
\end{figure*}

\begin{figure*}[h]
\centering
{\includegraphics[ width=1.0\linewidth ]{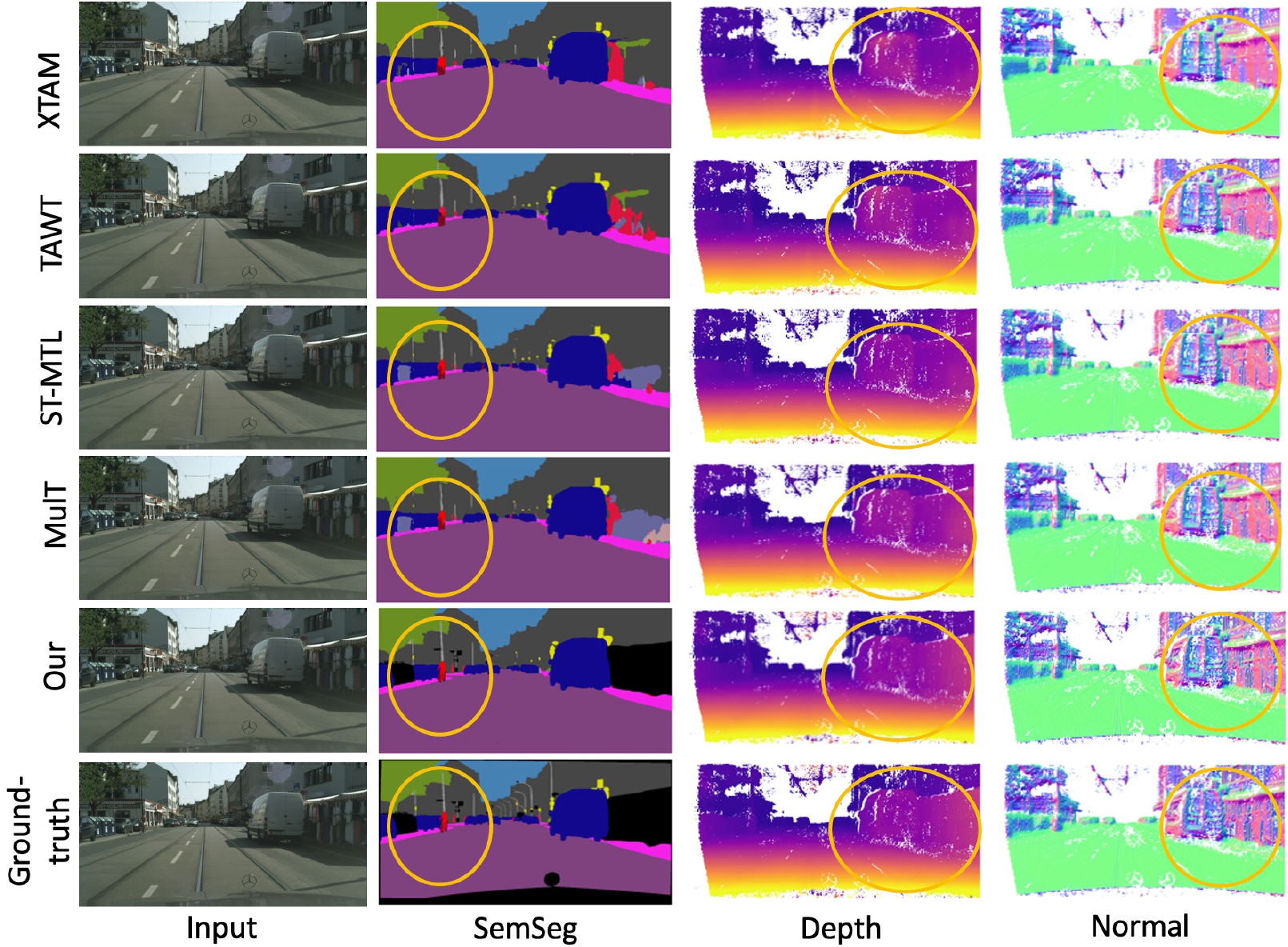}}
\setlength{\abovecaptionskip}{0pt}
\caption{{\textbf{Multitask Learning comparison on Cityscapes}~\cite{Cordts2016Cityscapes} benchmark in the \textit{'S-D-N'} setting. Our model outperforms all the multitask baselines. For instance, our method correctly segments the elements within the yellow-circled region, unlike the baselines. We show the best-performing methods in Table~\ref{tb:AVTaR-taxonomy-nyudv2-results} of the main paper, i.e. XTAM~\cite{xtam}, TAWT~\cite{tawt}, ST-MTL~\cite{spatiotemporalMTL}, and MulT~\cite{MulT}, respectively. Best seen on screen and zoomed in.} 
}\label{fig:cs-qr-sdn}\vspace{-10pt}
\end{figure*}

Additionally, we qualitatively compare our model for the multi-task learning setting with the best-performing baselines that utilize the same Swin-B V2 backbone. The results in Figure~\ref{fig:nyu-qr-sdne}, Figure~\ref{fig:synthia-qr-sdn}, and Figure~\ref{fig:cs-qr-sdn} show the performance of the different networks across multiple vision tasks on the NYUDv2~\cite{NYU}, Synthia~\cite{synthia}, and Cityscapes~\cite{Cordts2016Cityscapes}, respectively. %(c.f. L646- 647 main paper).
%All the multi-tasking models are jointly trained on the relevant task sets as per the datasets used and the single-task models are trained separately on each task. 
Our model yields higher-quality predictions than all the multitask baselines. 

\paragraph{Acknowledgement.} {This work was supported in part by the Swiss National Science Foundation via the Sinergia grant CRSII5$-$180359.} 
\balance

\clearpage
%%%%%%%%% REFERENCES
%{\small
%\bibliographystyle{ieee_fullname}
%\bibliography{egbib}
%}

\end{document}